\documentclass[review]{elsarticle}

\journal{Artificial Life}

\usepackage{graphicx}
\usepackage[cmex10]{amsmath}
\usepackage{array}
\usepackage{booktabs}
\usepackage{multirow}
\usepackage{graphicx,epstopdf,epsfig}
\graphicspath{{figures/}}
\DeclareGraphicsExtensions{.eps}
\usepackage{url}
\usepackage{float}
\usepackage[caption=false,font=footnotesize,labelfont=sf,textfont=sf]{subfig}
\usepackage{enumerate}
\usepackage{hyperref}

\begin{document}
	
\begin{frontmatter}
		
	\title{
		Towards Computational Models and Applications of Insect Visual Systems for Motion Perception: A Review
	}
		
	\author[add1,add2,add3,add4]{Qinbing Fu\fnref{fn1}}
	\ead{qifu@lincoln.ac.uk}
	\author[add3,add4]{Hongxin Wang}
	\ead{howang@lincoln.ac.uk}
	\author[add1,add2,add3,add4]{Cheng Hu}
	\ead{chu@lincoln.ac.uk}
	\author[add1,add2,add3,add4]{Shigang Yue\corref{cor1}}
	\ead{syue@lincoln.ac.uk}
	\fntext[fn1]{First author}
	\cortext[cor1]{Corresponding author}
	\address[add1]{School of Mechanical and Electrical Engineering, Guangzhou University, China}
	\address[add2]{Machine Life and Intelligence Research Centre, Guangzhou University, China}
	\address[add3]{Computational Intelligence Lab, School of Computer Science, University of Lincoln, UK}
	\address[add4]{Lincoln Centre for Autonomous Systems, University of Lincoln, UK}
		
	\begin{abstract}
			Motion perception is a critical capability determining a variety of aspects of insects' life, including avoiding predators, foraging and so forth.
			A good number of motion detectors have been identified in the insects' visual pathways. 
			Computational modelling of these motion detectors has not only been providing effective solutions to artificial intelligence, but also benefiting the understanding of complicated biological visual systems. 
			These biological mechanisms through millions of years of evolutionary development will have formed solid modules for constructing dynamic vision systems for future intelligent machines.
			This article reviews the computational motion perception models originating from biological research of insects' visual systems in the literature.
			These motion perception models or neural networks comprise the looming sensitive neuronal models of lobula giant movement detectors (LGMDs) in locusts, the translation sensitive neural systems of direction selective neurons (DSNs) in fruit flies, bees and locusts, as well as the small target motion detectors (STMDs) in dragonflies and hover flies. 
			We also review the applications of these models to robots and vehicles.	
			Through these modelling studies, we summarise the methodologies that generate different direction and size selectivity in motion perception.
			At last, we discuss about multiple systems integration and hardware realisation of these bio-inspired motion perception models. 
	\end{abstract}
		
\end{frontmatter}
	

\section{Introduction}
\label{Sec: Introduction}

Motion perception is critically important to serve a variety of daily tasks for animals and humans.
Insects, in particular, are `experts' in motion perception, even though they have tiny brains and much smaller number of visual neurons compared to vertebrates.
Much evidence has demonstrated their amazing ability to deal with visual motion and interacting with dynamic and cluttered scenes corresponding to quick and flexible reactions like collision avoidance and target tracking and following, and even on some aspects, performing better than vertebrates and humans \cite{Borst-2010(review-fly-vision), Borst2011(review-motion),Borst-2014(review-fly),Borst2015(common-circuit-motion),Fu-TAROS(review),Nicolas-2014(Review-Fly-Robot),Serres2017(review-optic-flow)}.

Insects have compact visual systems that can extract meaningful motion cues and distinguish different motion patterns for proper behavioural response.
For example, locusts can fly for hundreds of miles in dense swarms free of collision; 
honeybees show centre response when crossing a tunnel; 
preying mantises can monitor small moving prey in visual clutter. 
Such appealing talent draws attention from not only biologists but also computational modellers and engineers.
In terms of biology, the underlying circuits and mechanisms of insects' visual processing systems remain largely unknown to date \cite{Serres2017(review-optic-flow),Nicolas-2014(Review-Fly-Robot),Fu-TAROS(review)}. 
While the biological substrates are elusive, the computational modelling and applications to machine vision are of particular usefulness to help understand the neural characteristics and demonstrate the functionality of visual circuits or pathways \cite{Serres2017(review-optic-flow),Borst2015(common-circuit-motion),Nicolas-2014(Review-Fly-Robot)}.
In addition, these models can be ideal modules for designing dynamic vision systems or sensors for future intelligent machines like robots and vehicles for motion perception in a low-energy, fast and reliable manner.

\subsection{Related survey of biological visual systems research}
\label{Sec: Introduction: bio-reviews}

The past several decades have witnessed much progress in our understanding of cellular and sub-cellular mechanisms of mysterious biological visual systems. 
There have been some specific visual neurons or pathways identified in insects like various kinds of flies \cite{Fly-1998(dendritic-optic-flow),Fly-2004(Tammero-spatial-visuomotor),Fly-2006(feedback-control-photoreceptor),Fly-2013(motion-blind-tracking),Fly-2013(motion-circuit-connectomics),Fly-2013(neuronal-variability),Fly-2014(processing-properties-motion),Fly-2015(model-collision-reichardt),Fly-DS-2016(preferred-null),Fly-DS-2017(emergence-direction-selectivity),Fly-Motion-2017(cellular-hybrid-detector),Fly1995(network-tracking-fly)}, locusts \cite{Locusts-2012(predator-prey-behaviours),LGMD-2016(Gray-background-motion),LGMD-2009(presynaptic-looming-selectivity),LGMD1-Escapes2010(visual-startle-locusts),LGMD1-1999(Rind-seeing-collision),LGMDs-2016,LGMD1-synaptic2015(first-stage-locust),Sztarker2014(LGMD2-development)}, bees \cite{Bee-2009(visual-processing-brain),Baird2010(viewing-angle-bumblebees),Landing-2013(visually-guided-strategy)}, as well as ants \cite{Zeil2012(visual-homing-insect)} and mantises \cite{Mantis-2000(behavioural-model-robot),Yamawaki2011(mantis-looming-defence)}.
Two researches have reviewed fundamental mechanisms in insect visual motion detection; these comprise classic models and functions \cite{MotionReview2002(models-cells-functions),Borst-2002(review-networks-fly)}.
At early stages, fly visual systems are seminal models to study animals' motion-detecting strategies \cite{EMD-1989(principles-review)}.
Borst et al. have reviewed thoroughly the step-by-step physiological findings on the fly visual systems and summarised the visual course control; these include behaviours, algorithms and circuits \cite{Borst-2002(review-networks-fly),Borst-2010(review-fly-vision),Borst-2014(review-fly),Borst2011(review-motion),Egelhaaf1993(fly-algorithms-neurons)}.
Importantly, they have also pointed out the commonality in design of fly and mammalian motion vision circuits \cite{Borst2015(common-circuit-motion)}. 
By contrast to the correlated elementary motion which is velocity-dependent, Aptekar briefly reviewed the higher-order figure detection with non-Fourier or statistical features in flies that correlates with human vision  \cite{Motion-2013(higher-order-discrimination)}.
In addition, Rind et al. devoted to research the underlying structures and mechanisms of locust visual systems to learn looming perception and collision avoidance schemes \cite{Rind1998(local-circuit-locust),Rind2000(bio-evidence),Rind_2008(avoidance-flying-locust),Simmons-1997(LGMD2-neuron-locusts),LGMD1-1996(Rind-intracellular-neurons),LGMD1-1996(Rind-neural-network),LGMD1-1999(Rind-seeing-collision),LGMDs-2016,Sztarker2014(LGMD2-development),LGMD1-synaptic2015(first-stage-locust),DCMD-1997(collision-trajectories),DCMD-1992(selective-response-approaching),DCMD-2013}.
On the behavioural level, a research reviewed escape behaviours in insects caused by visual stimuli, and moreover, demonstrated the complexity of both visual and escape circuits \cite{Escape-2012(insects-behaviours)}.

\subsection{Related survey of bio-inspired models and applications}
\label{Sec: Introduction: model-reviews}

These naturally evolved vision systems have been providing us with a rich source of inspiration for developing artificial visual systems for motion perception.
As hardware has swiftly developed, these bio-inspired models become applicable to robotics.
A good number of surveys emerged to demonstrate how machine vision benefits from computational modelling of insects vision.
Iida reviewed the models motivated by flying insects and the applications to robotics \cite{Iida-2012(book-review)}.
Floreano et al. proposed an overview of applying bio-inspired control methodology for vision-based wheeled and flying robots \cite{Floreano2005(wheel-wing-spiking)}.
Srinivasan et al. studied rigorously the models and control methods inspired by visual systems in flying insects like honeybees for visually guided flight control and navigation \cite{Navigation-1999(robot-insect-vision),Srinivasan2011(visual-insects-robotics),Honeybee-2011(flight-navigation-robotics)}.
Huber presented visuomotor control in flies and visually behaviour-based models, control and design for robotics \cite{Huber-2003(visuomotor-book-review)}.
Franceschini devoted to survey biological research and computational models on the basis of a scheme of fly elementary motion detectors (EMDs) and relevant bio-robotics applications, systematically \cite{Nicolas-2014(Review-Fly-Robot)}.
Recently, Serres and Ruffier reviewed the applications of fly optic flow-based strategies to UAVs and MAVs for multiple visually guided behaviours, like collision avoidance, terrain following, landing and tunnel crossing \cite{Serres2017(review-optic-flow)}.
More generically, Desouza and Kak surveyed vision techniques, that varied from traditional computer vision methods to insects optic flow strategies, for mobile robots navigation \cite{DeSouza_2002(survey-vision-robot)}.
On the other hand, Webb reflected with the influence of robot-based research including bio-inspired vision on biological behaviour of animals \cite{Webb2000(robotics-animal-behaviour),Webb2001(robots-models-behaviour)};
importantly, these seminal works revealed that the bio-robotic studies could be good paradigms for studying biological behaviours.

\subsection{Taxonomy of the proposed review}
\label{Sec: Introduction: our-review}

There are a lot of publications in computational modelling of motion perception neural systems originating from animals' visual systems research in the past decades, shedding light on significant breakthrough in bio-inspired artificial visual systems for future robotics and autonomous vehicles. 
These publications cover computational modelling of different motion patterns, such as looming, translation, small target motion, and rotation and etc., as well as various applications.
However, there is no systematic review on this promising research field though some of them have been casually touched upon in different papers.

\begin{figure}[H]
	\centering
	\includegraphics[width=0.9\textwidth]{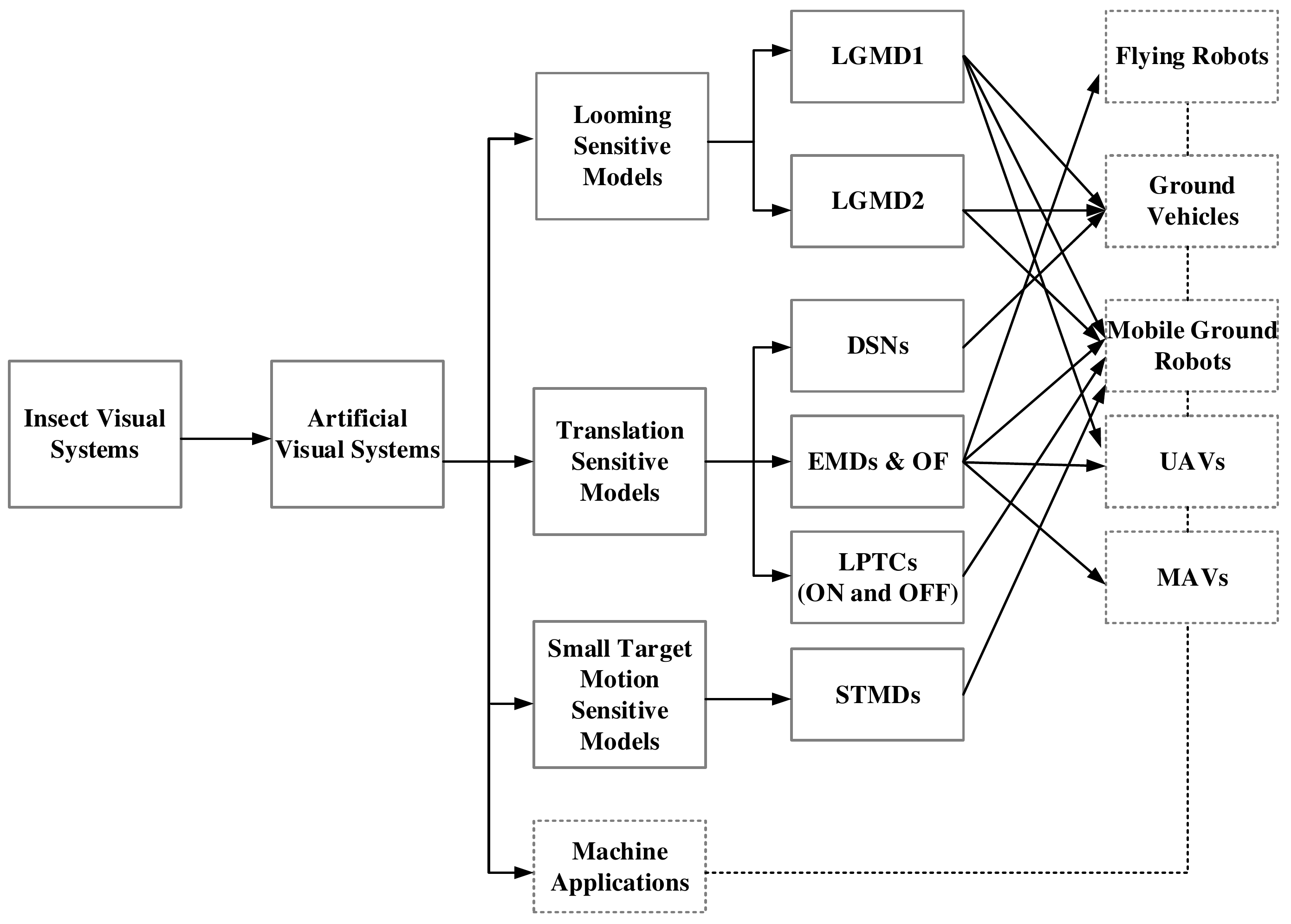}
	\caption{The taxonomy proposed in this review.}
	\label{Fig: taxonomy}
\end{figure}

To the best of our knowledge, the proposed survey for the first time covers computational models sensitive to different motion patterns including looming, translation, small target motion, that originate from several kinds of insects' visual system research. 
These insects include locusts, fruit flies (or drosophila), dragonflies, hover flies, and bees (bumblebees and honeybees).
The vast majority of biological and computational studies have been focusing on the various kinds of flies and locusts.
Although there have been a few reviews on biological and computational models, as well as applications of fly visual systems, e.g. \cite{Serres2017(review-optic-flow),Nicolas-2014(Review-Fly-Robot),Borst-2014(review-fly)}; no survey has been provided to involve the looming sensitive neuron models inspired by locust visual systems and the small target motion sensitive neuron models inspired by dragonflies and hover flies and corresponding applications, systematically.

\begin{table}[t]
	\centering
	\caption{Nomenclature of this review}
	\begin{tabular}{l|l}
		\toprule
		\cmidrule{1-2}
		LGMD(s)	&lobula giant movement detector(s)\\
		DCMD	&descending contra-lateral movement detector\\
		DSN(s)	&direction selective neuron(s)\\
		STMD(s)	&small target motion detector(s)\\
		SFS		&small field system\\
		EMD(s) 	&elementary motion detector(s)\\
		LPTC(s) &lobula plate tangential cell(s)\\
		LGN		&lobula giant neuron\\
		OF 		&optic flow\\
		HR(C)	&Hassenstein-Reichardt (correlator)\\
		FDN(s)	&figure detection neuron(s)\\
		PD(s)	&preferred direction(s)\\
		ND(s)	&null or non-preferred direction(s)\\
		DS		&direction selectivity\\
		VLSI	&very large scale integration\\
		FPGA	&field-programmable gate array\\
		UAV		&unmanned aerial vehicle\\
		MAV		&micro air vehicle\\
		\bottomrule
	\end{tabular}
	\label{Tab: Nomenclature}
\end{table}

In the real world, the diversity of motion patterns can be categorised into a few types that involve expansion and contraction of objects, translation, rotational or spiral motion and etc.
Depending on the distance between moving objects and the observer, it can be also defined the other specific motion pattern of small target movements.
Perceiving and recognising these different motion patterns in a visually cluttered and dynamic environment is critically important for the survival of insects.
With regard to the diversity of visual neurons that possess specific sensitivity to different motion patterns, this article reviews bio-inspired motion perception models and applications in the literature according to different direction and size selectivity, as illustrated in the Fig. \ref{Fig: taxonomy}.
These models represent the distinct direction selectivity (DS) to looming and translation in visual neurons of locusts, fruit flies and bees, as well as the specific size selectivity to small target motion in visual neurons of dragonflies and hover flies.

Another significance of this review is that it summarises the similarities in computational modelling of different visual neurons. 
Moreover, this review demonstrates the key methods for generation of both the direction and the size selectivity in motion perception models. 
This review also suggests that the computational modelling of similar motion sensitive neurons in other insects like mantis and arthropods like crabs may learn from the existing models. 
Furthermore, we discuss about the integration of multiple neural systems to handle more complex visual tasks, and point out the hardware realisation of these models.

The taxonomy proposed in this review is given in the Fig. \ref{Fig: taxonomy}.
The Table \ref{Tab: Nomenclature} illustrates the nomenclature of this review.
The rest of this article is organised as follows:
First, the looming sensitive neuronal models and applications of two locust lobula giant movement detectors (LGMDs) -- LGMD1 and LGMD2 will be reviewed in the Section \ref{Sec: Looming-Motion}. 
In the Section \ref{Sec: Translational-Motion}, we will introduce translation sensitive neural systems and their wide applications in flying robots including UAVs and MAVs; we will also present the cutting-edge research of ON and OFF pathways underlying motion perception in biological visual systems. 
In the Section \ref{Sec: Small-Target-Motion}, we will survey a specific group of visual neural networks for sensing small target motion. 
In the Section \ref{Sec: Discussion}, we will summarise the commonality in computational modelling of different insect visual systems, as well as demonstrating the generation of different direction and size selectivity in these models. 
We will also have further discussion about multiple systems integration, potential hardware implementations of these motion perception models. 
Finally, we will summarise this review paper in the Section \ref{Sec: Conclusion}.

\section{Looming perception neuron models}
\label{Sec: Looming-Motion}

This section reviews looming sensitive neuronal models as collision-detecting systems and applications inspired by the locust visual systems.
These include two neuronal models of the LGMD1 and the LGMD2. 
In addition, this section introduces different methods to shape the looming selectivity in computational structures.
This section covers also existing applications of these looming detectors in mobile ground robots, UAVs and ground vehicles.

The looming stimuli indicate movements in depth of objects that approach representing collision, which is very frequent visual challenges to animals.
Recognising looming objects, timely and accurately, is certainly crucial for animals' survival deciding a variety of visually guided behaviours like avoiding predators. 
A few looming sensitive neurons have been explored in insects like locusts \cite{LGMD-1974} and flies \cite{FlyLooming-2012}, and arthropods like crabs \cite{Crab2014(computation-approach-neurons)}.
Amongst these animals, the locusts have been researched with a good number of studies demonstrating looming perception schemes, e.g. \cite{LGMD1-1999(Rind-seeing-collision),LGMDs-2016,Locusts-2012(predator-prey-behaviours)}, which have been adopted to design artificial collision sensitive models and sensors, e.g. \cite{LGMD1-2007(obstacle-avoidance-robot),LGMD1-2001(sensory-coding-robot),LGMD1-Yue2009(near-range-navigation),Fu-2016(LGMD2-BMVC),Badia-2010(LGMD1-nonlinear-model)}. 
As the results of millions of years of evolutionary development, locusts are `experts' in collision detection and avoidance, which can fly in dense swarms for hundreds of miles free of collision. 
Realising this robust ability is required for future intelligent machines like autonomous robots and vehicles interacting with dynamic and complex environments.

\subsection{LGMD1-based neuron models and applications}
\label{section: literature: looming: lgmd1}

\subsubsection{Biological research background}

As early in the 1970s, biologists had explored, anatomically, a group of large inter-neurons in the lobula neuropile layer of the locust's visual brain. 
These neurons were named as lobula giant movement detectors (LGMDs) \cite{LGMD-1974,LGMD-1976(ON-OFF)}.
The LGMD1 was first identified as a movement detector and gradually recognised as a looming objects detector, e.g. \cite{LGMD-1974,DCMD-1992(selective-response-approaching),LGMD1-1996(Rind-intracellular-neurons)}. 
In the same place, the LGMD2 was also identified as a looming detector but with unique characteristics that are different to the LGMD1 \cite{Simmons-1997(LGMD2-neuron-locusts)}. 
Both the LGMD1 and the LGMD2 respond most strongly to objects that approach over other kinds of movements like recession and translation \cite{LGMD1-1996(Rind-neural-network),LGMD1-1999(Rind-seeing-collision),LGMDs-2016}.

\begin{figure}[H]
	\centering
	\subfloat[]{\includegraphics[width=0.45\textwidth]{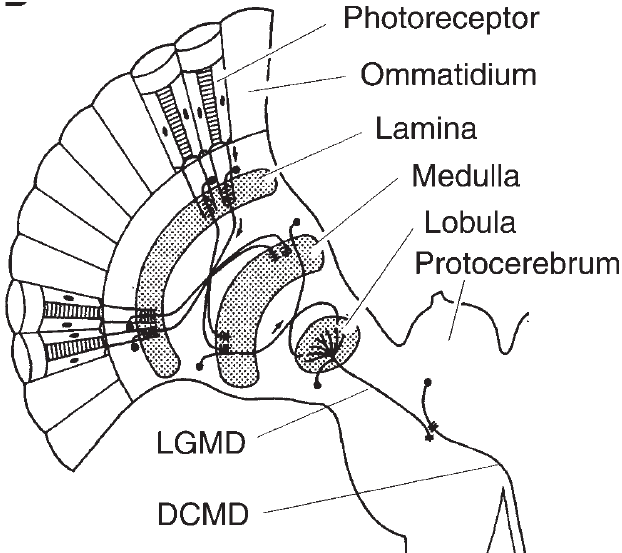}
		\label{LGMD-neuromorphology-Rind}}
	\hfil
	\subfloat[]{\includegraphics[width=0.4\textwidth]{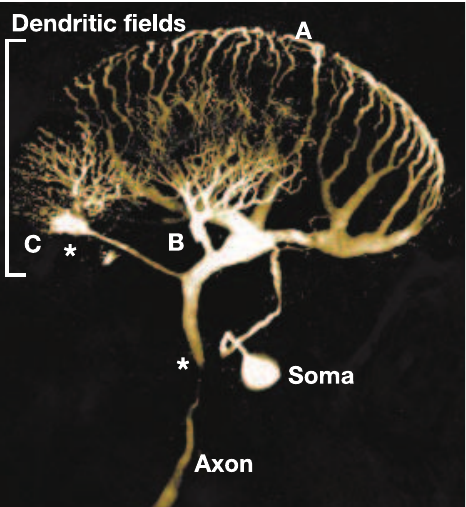}
		\label{LGMD-neuromorphology-Gabbiani}}
	\caption{Illustrations of the LGMD1 morphology: (a) illustration of the pre-synaptic neuropile layers to the LGMD1 neuron and the 					post-synaptic one-to-one target DCMD neuron, adapted from \cite{LGMD1-1999(Rind-seeing-collision)}, 
		(b) illustration of the LGMD1's large dendritic fan (A) and two additional dendritic fields (B, C) that receive distinct synaptic inputs, adapted from \cite{Gabbiani2002(multiplicative-computation-LGMD)}.}
	\label{Fig: LGMD1-neuromorphology}
\end{figure}

\begin{figure}[H]
	\centering
	\subfloat[]{\includegraphics[width=0.35\textwidth]{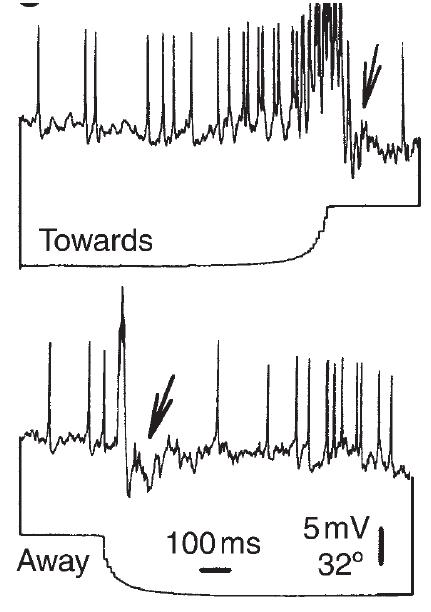}
		\label{LGMD1-neural-response-Rind}}
	\hfil
	\subfloat[]{\includegraphics[width=0.5\textwidth]{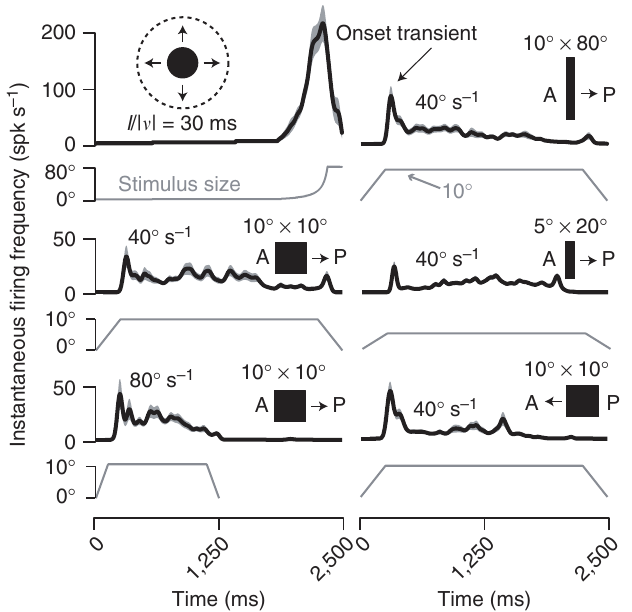}
		\label{LGMD1-neural-response-Gabbiani}}
	\caption{LGMD1 neural response to approaching and receding stimuli (a), adapted from \cite{LGMD1-1999(Rind-seeing-collision)}: arrows indicate a hyper-polarisation response of strong inhibition after activation.
		(b) LGMD1 neural response to approaching and translating stimuli by a variety size and speed of moving objects, adapted from \cite{SFA-2009}.}
	\label{Fig: LGMD1-neural-response}
\end{figure}

The vast majority of researches have been concentrating on the LGMD1.
This neuron has been demonstrated to play dominant roles in locusts capable of flying \cite{LGMD-2016(Gray-background-motion),LGMD1-1999(Rind-seeing-collision),Rind2000(bio-evidence),Sztarker2014(LGMD2-development),LGMD1-synaptic2015(first-stage-locust),LGMD1-1996(Rind-neural-network)}. 
In terms of neuromorphology, the Fig. \ref{Fig: LGMD1-neuromorphology} illustrates an LGMD1 neurone and both its pre-synaptic and post-synaptic neural structures. 
Generally speaking, the LGMD1 integrates visual signals from different dendritic areas; these generate two kinds of flows, i.e. the excitations and the inhibitions. 
The neural processing within the circuit is a competition between these two types of flows \cite{LGMD1-1996(Rind-neural-network),Gabbiani2004(invariance-LGMD)}.
More precisely, the activation of LGMD1 requires the `winner' of competition to be the excitatory flow. 
In addition, the descending contra-lateral movement detector (DCMD) is a one-to-one connection of post-synaptic target neuron to the LGMD1 \cite{DCMD-1992(selective-response-approaching),DCMD-1997(collision-trajectories),DCMD-2013,Gray2001(DCMD-headon-stimuli)}; this neuron conveys the generated spikes by the LGMD1 to following motion control neural systems corresponding to avoidance behaviours \cite{DCMD-2013}.

\begin{figure}[H]
	\centering
	\includegraphics[width=0.86\textwidth]{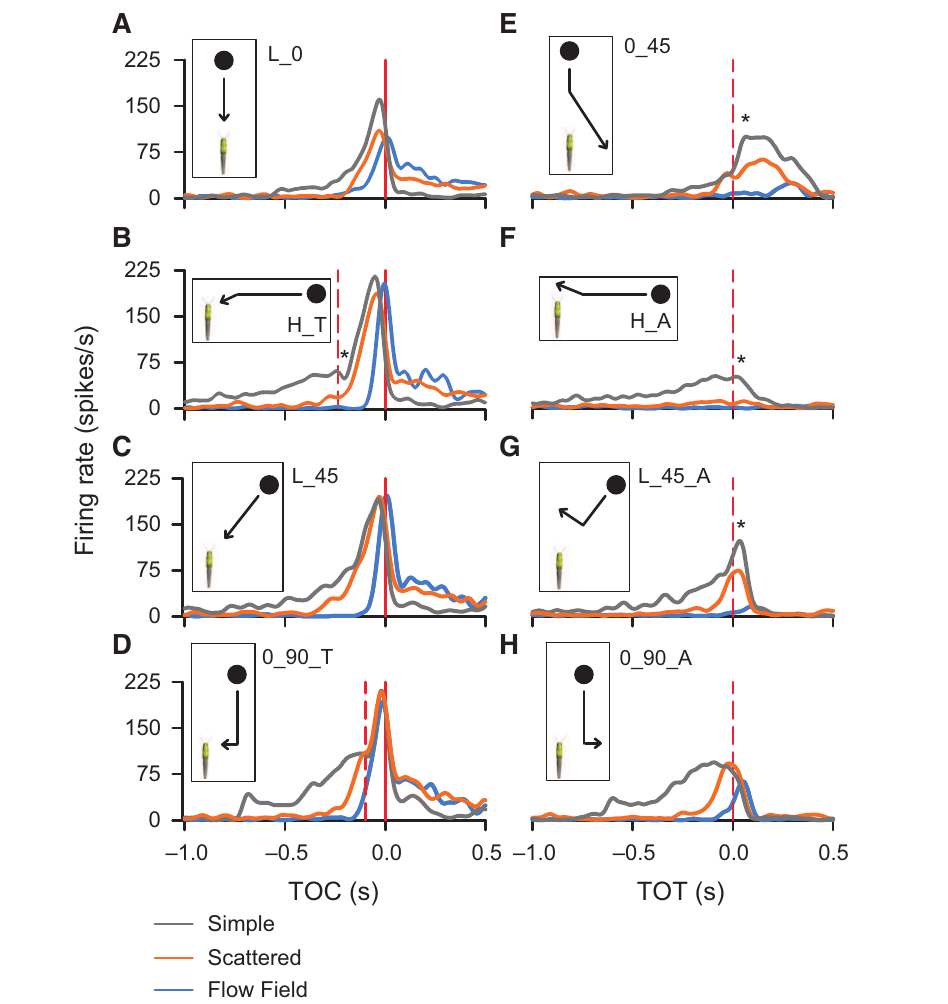}
	\caption{The LGMD1-DCMD pathway in locusts responds to a variety of courses of collision and deviations from collision, against a simple background or dynamic clutter, adapted from \cite{LGMD-2016(Gray-background-motion)}.
		Red solid vertical lines indicate time of collision. 
		Red dashed vertical lines represent time of transition.
		Asterisks indicate the time of a local valley or peak in response to a transition.}
	\label{Fig: LGMD1-background}
\end{figure}

So, what does the LGMD1's neural response look like?
The Fig. \ref{Fig: LGMD1-neural-response} demonstrates the responses to different visual stimuli; these comprise objects approaching and receding (Fig. \ref{LGMD1-neural-response-Rind}), as well as translating stimuli by varied sizes of dark objects and speeds of translations (Fig. \ref{LGMD1-neural-response-Gabbiani}).
It can be clearly seen from the Fig. \ref{Fig: LGMD1-neural-response} that the LGMD1 neuron responds most strongly to looming objects that approach, representing the highest firing rates.
It is only briefly excited by the object moving away.
The translating stimuli bring about short-term and weak responses of the LGMD1 regardless of the size, direction and speed of objects.
More recently, Yakubowski et al. demonstrated the neural response of LGMD1 in locusts against a visually cluttered or dynamic background and more abundant visual stimuli including objects deviate from a collision course \cite{LGMD-2016(Gray-background-motion)}.
It can be seen from the Fig. \ref{Fig: LGMD1-background} that the LGMD1 responds vigorously to a variety of oncoming threats; it can well discriminate collision from movements that objects transit to recession; this response is also affected by the degree of complexity of background motion like dynamic visual clutter.

\begin{figure}[H]
	\centering
	\includegraphics[width=0.85\textwidth]{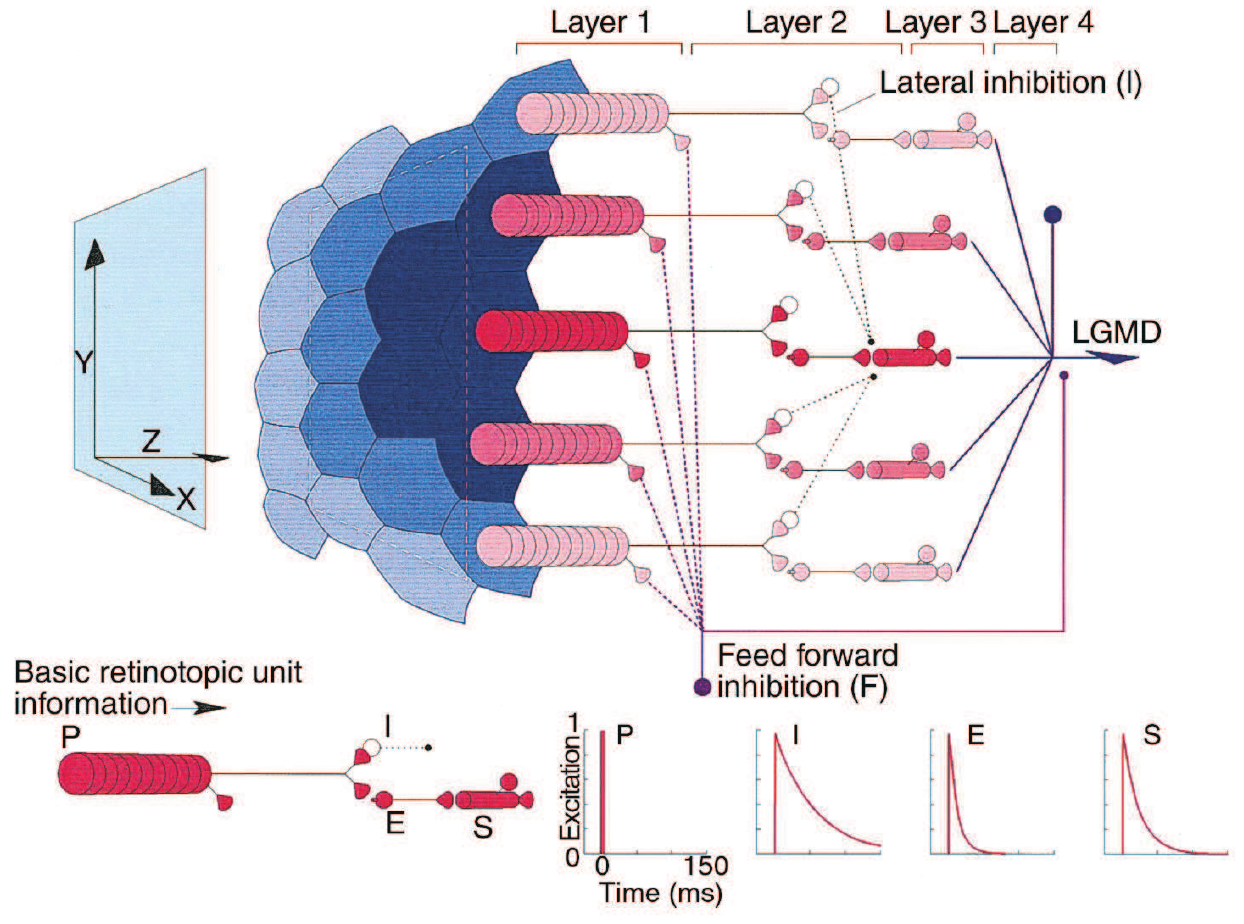}
	\caption{Schematic of LGMD1 visual neural network proposed by Rind, adapted from \cite{LGMD1-1999(Rind-seeing-collision)}: this network consists of four layers of photoreceptors (P), excitations (E), lateral inhibitions (I) and summation cells (S), as well as two cells of LGMD1 and feed forward inhibition (F).}
	\label{Fig: LGMD1-network-Rind-1996}
\end{figure}

\begin{figure}[H]
	\centering
	\subfloat[]{\includegraphics[width=0.3\textwidth]{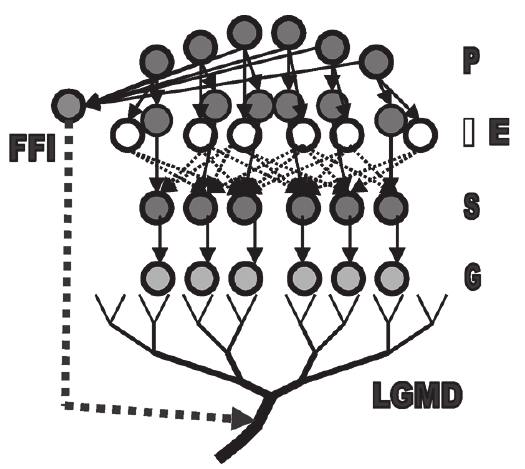}}
	\hfil
	\subfloat[]{\includegraphics[width=0.33\textwidth]{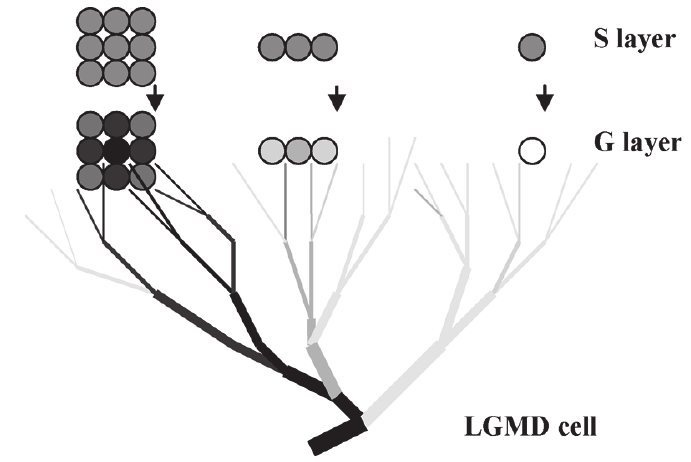}}
	\hfil
	\subfloat[]{\includegraphics[width=0.33\textwidth]{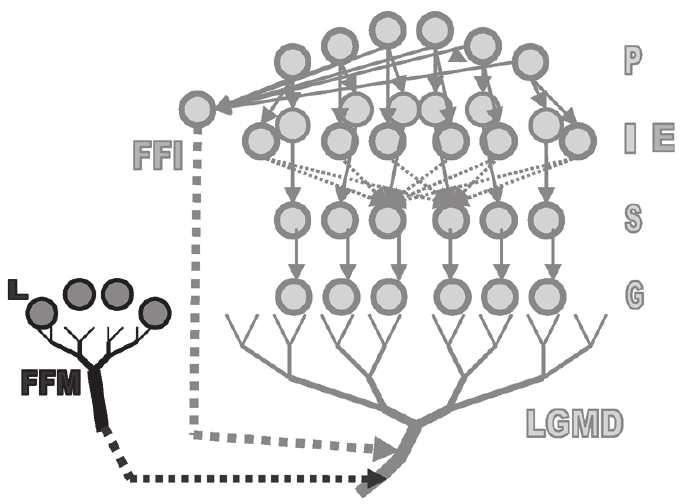}}
	\caption{Schematic of an LGMD1-based visual neural network taken only $6$ cells as an instance (a), with a G (grouping) layer (b) and a FFM (feed forward mediation) mechanism (c), adapted from \cite{LGMD1-Glayer(feature-enhancement)}.}
	\label{Fig: LGMD1-Glayer}
\end{figure}

\subsubsection{Computational models and applications}

Computational modelling of such a fascinating looming sensitive neuron has emerged since the 1990s.
A seminal work was proposed by Rind and Bramwell to model an LGMD1-based neural network \cite{LGMD1-1996(Rind-neural-network)}, as illustrated in the Fig. \ref{Fig: LGMD1-network-Rind-1996}. 
In this research, they looked deeper into the pre-synaptic signal processing mechanism in the looming sensitive neural network and proposed a way explaining how the lateral inhibitions can play crucial roles to cut down the motion-dependent excitations in a both spatial and temporal mode; importantly, this mechanism effectively shapes the LGMD1's looming selectivity to respond most powerfully to approaching objects.
Importantly, this research highlighted that the visual information sensed by the first layer of photoreceptors is divided into two kinds of signals within the pre-synaptic structure, that is, the excitations and the lateral inhibitions. 
In addition, the lateral inhibitions are temporally delayed relative to the excitations and spread out to neighbouring cells, symmetrically in space and decaying in temporal. 
The interaction between such two types of signals determines the specific looming selectivity of the LGMD1 to approaching rather than receding and translating movements.
In addition, a feed forward inhibition can suppress the LGMD1 neuron, directly. 
It can also mediate the LGMD1's response at some critical moments like the end of approach and the start of recession.

\begin{figure}[H]
	\centering
	\subfloat[]{\includegraphics[width=0.33\textwidth]{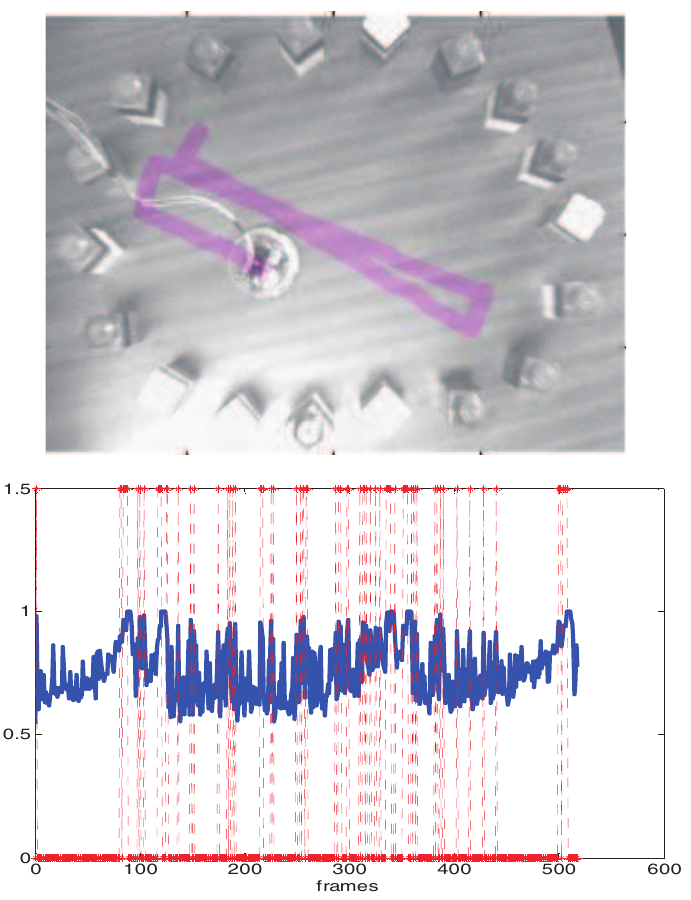}}
	\hfil
	\subfloat[]{\includegraphics[width=0.33\textwidth]{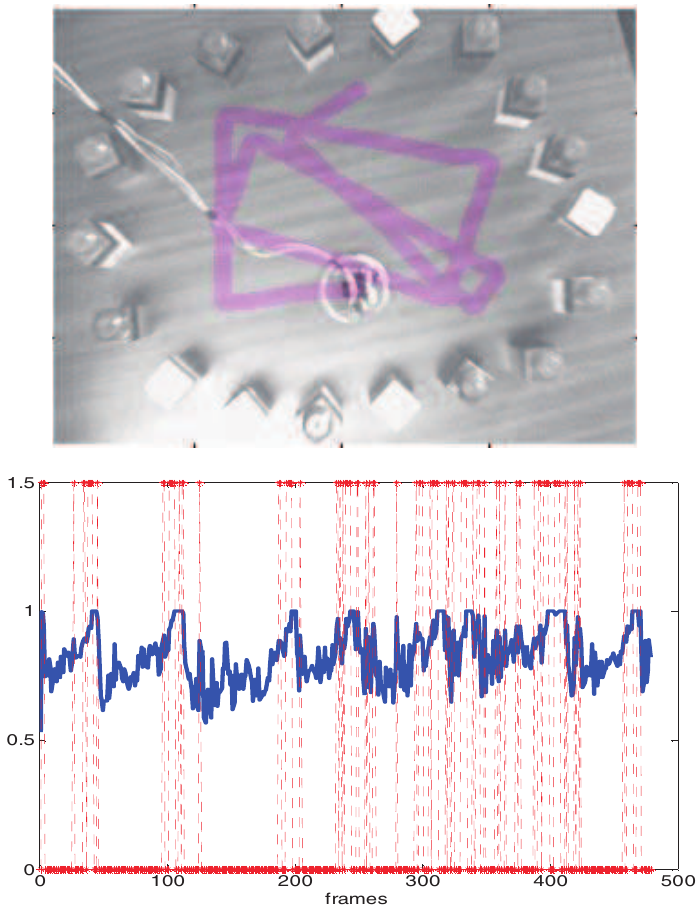}}
	\hfil
	\subfloat[]{\includegraphics[width=0.33\textwidth]{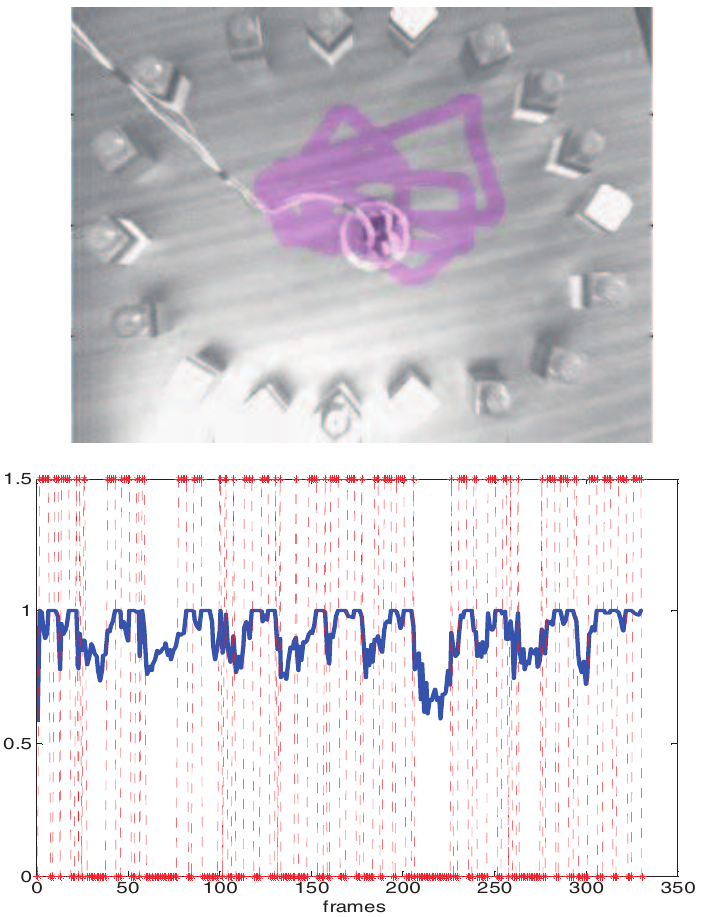}}
	\vfil
	\vspace{-0.1in}
	\subfloat[]{\includegraphics[width=0.33\textwidth]{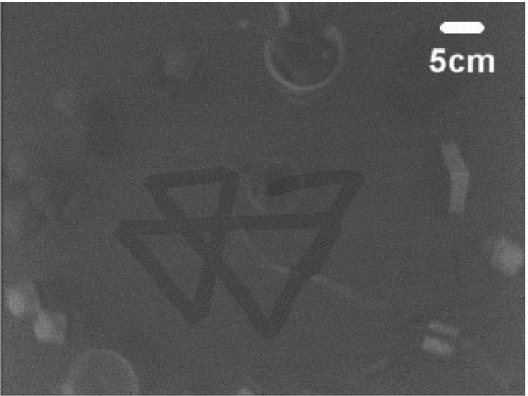}}
	\hfil
	\subfloat[]{\includegraphics[width=0.33\textwidth]{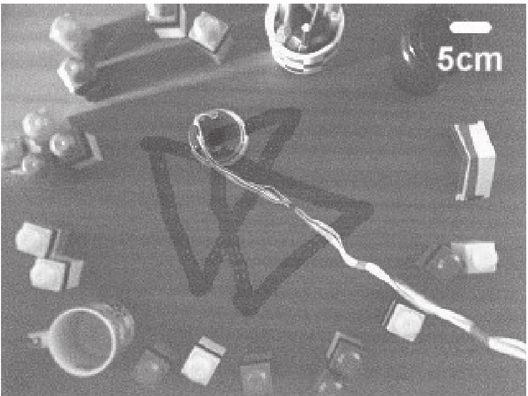}}
	\hfil
	\subfloat[]{\includegraphics[width=0.33\textwidth]{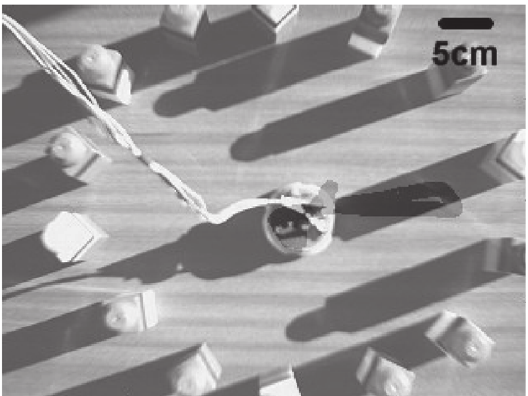}}
	\caption{The results of robot collision detection with normalised neural responses (blue lines) and burst of spikes (red-dashed lines), tested by different speeds from slow (a) to fast (c) as well as light conditions from dim (d) to extremely bright (f): the overtime trajectory is shown for each result.
		(a)--(c) are adapted from \cite{Yue-2005(LGMD1-ICRA)} and (d)--(f) are adapted from \cite{LGMD1-Glayer(feature-enhancement)}.}
	\label{Fig: LGMD1-Glayer-robot}
\end{figure}

\begin{figure}[H]
	\centering
	\includegraphics[width=0.9\textwidth]{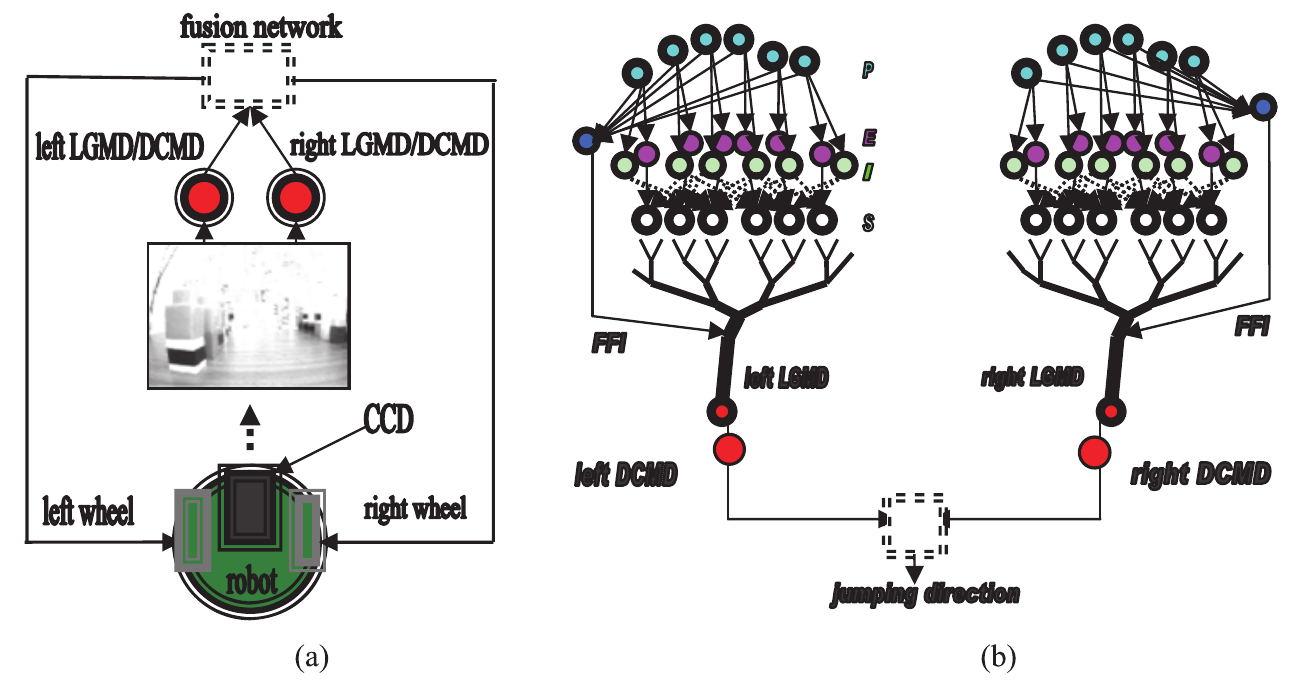}
	\caption{A reactive collision avoidance strategy via integrating a bilateral pair of LGMD1 neuronal models to control left and right wheels of the robot, respectively, adapted from \cite{Yue-2010(LGMD1-robot-bilateral)}: (a) the control strategy in robot, (b) the bilateral LGMD1-DCMD visual neural networks.}
	\label{Fig: LGMD1-Yue-robot2009}
\end{figure}

\begin{figure}[H]
	\centering
	\includegraphics[width=0.9\textwidth]{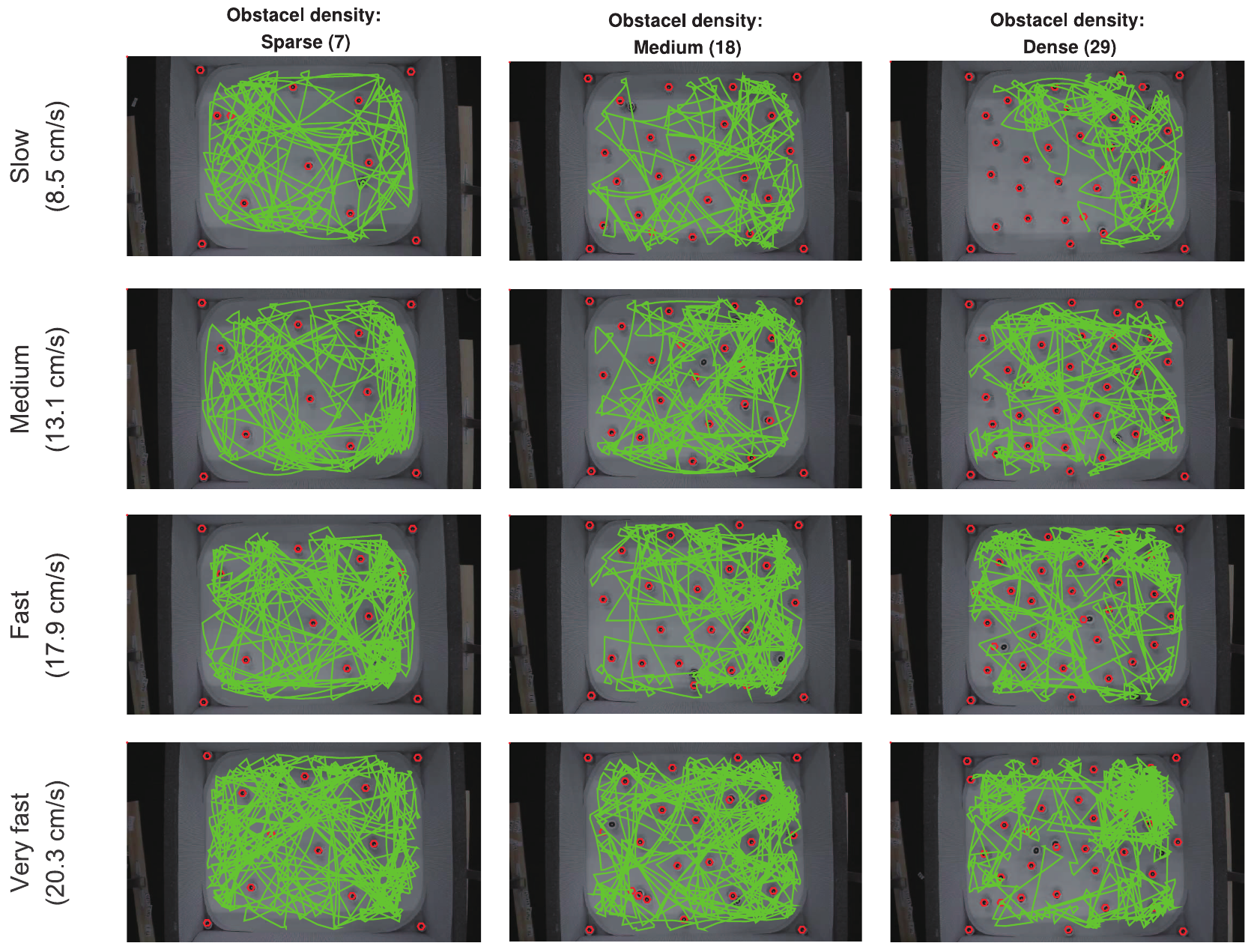}
	\caption{Micro-robot arena tests implemented with an LGMD1 model \cite{Hu-2017(Colias-LGMD1)} as the only collision-detecting sensor: the agent was tested at different speeds and with varied densities of obstacles in an arena. 
		The green line indicates robot overtime trajectory and the red circles denote the obstacles. 
		The experimental data is adapted from \cite{Hu-2017(Colias-LGMD1)}.}
	\label{Fig: LGMD1-Cheng-arena}
\end{figure}

\begin{figure}[H]
	\centering
	\subfloat[]{\frame{\includegraphics[width=0.33\textwidth]{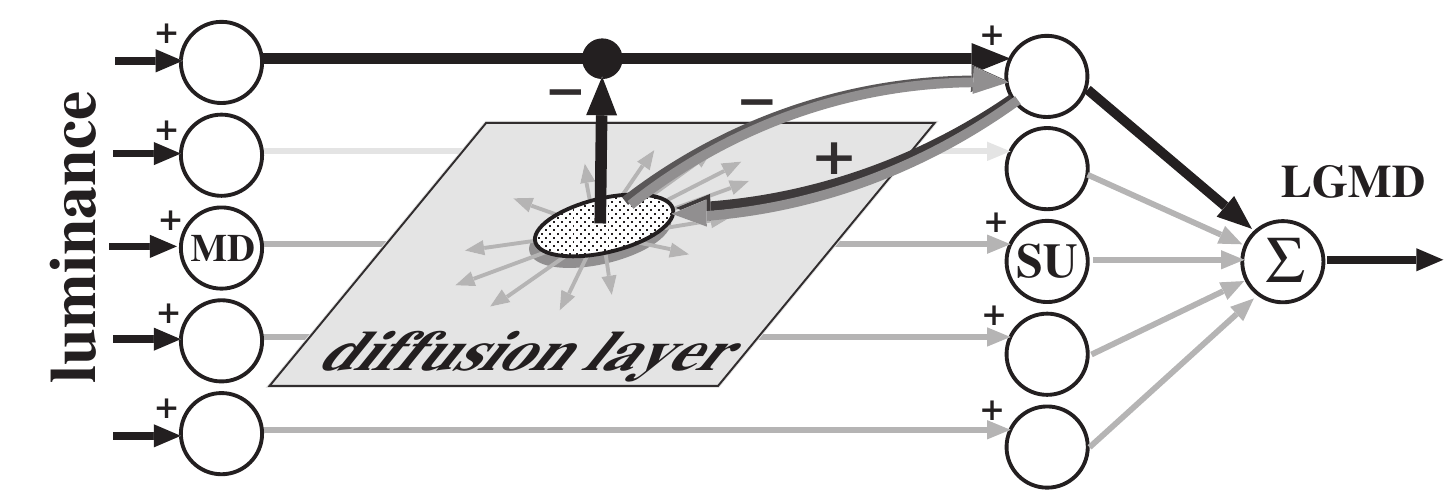}}}
	\hfil
	\subfloat[]{\frame{\includegraphics[width=0.31\textwidth]{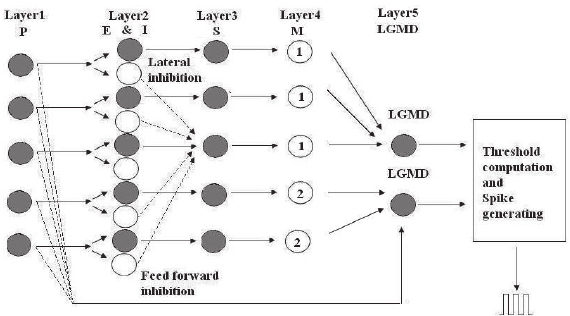}}}
	\hfil
	\subfloat[]{\frame{\includegraphics[width=0.3\textwidth]{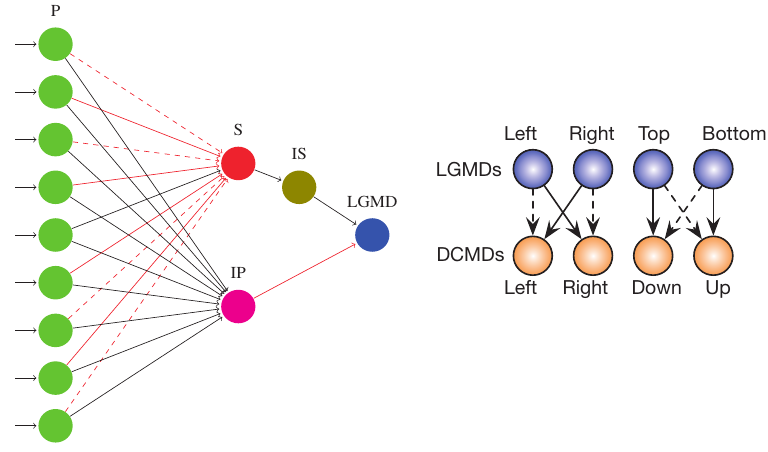}}}
	\vfil
	\vspace{-0.1in}
	\subfloat[]{\frame{\includegraphics[width=0.19\textwidth]{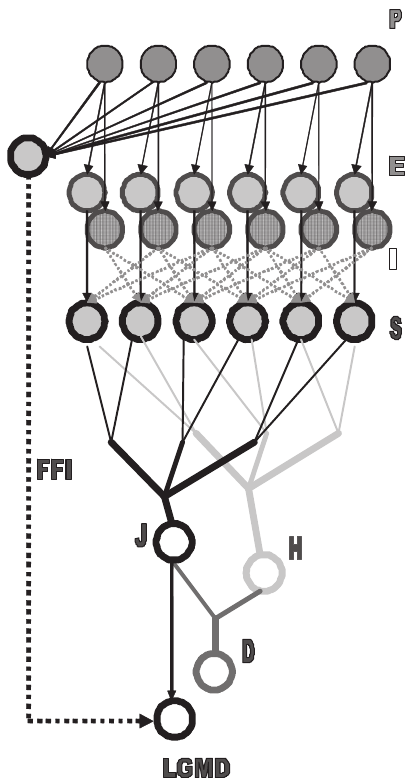}}
		\label{LGMD1-Meng2010}}
	\hfil
	\subfloat[]{\frame{\includegraphics[width=0.3\textwidth]{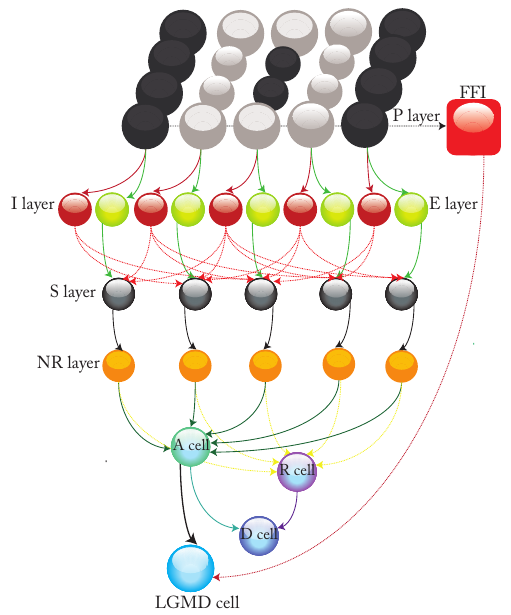}}}
	\hfil
	\subfloat[]{\frame{\includegraphics[width=0.4\textwidth]{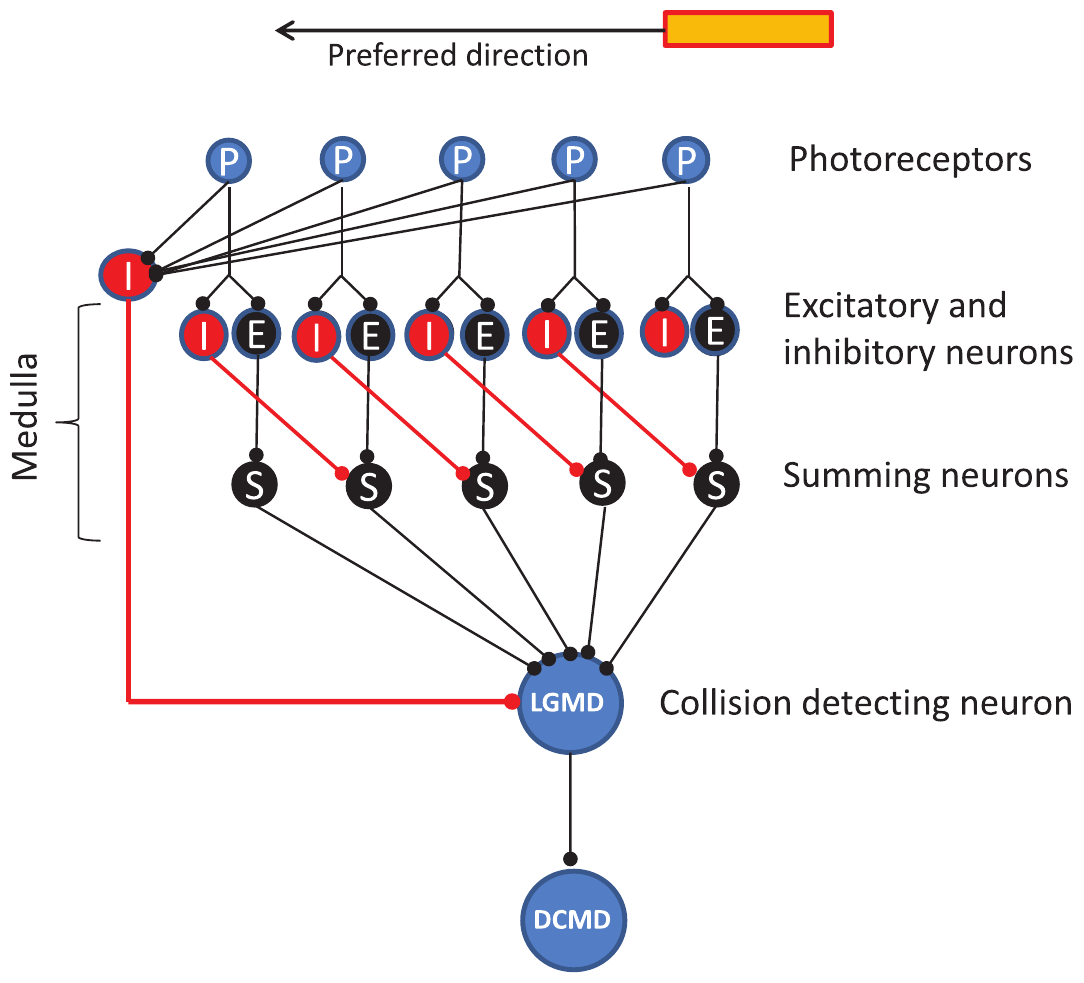}}}
	\caption{Schematics of a variety of LGMD1-based visual neural networks: (a) an LGMD1 model encoding onset and offset responses by luminance increments and decrements, adapted from \cite{LGMD1-ONOFF-2004}, (b) a modified LGMD1 model for multiple looming objects detection, adapted from \cite{LGMD1-2012(multi-object-detection)}, (c) a simplified LGMD1 model for collision avoidance of an UAV, adapted from \cite{UAV2017(LGMD1-spiking)}, (d) a modified LGMD1 model with enhancement of collision selectivity, adapted from \cite{Meng-2009(IJCNN-LGMD1),Meng-2010(LGMD1-FPGA)}, (e) a modified LGMD1 model with a new layer for noise reduction and spiking-threshold mediation, adapted from \cite{LGMD1-2013(Silva-IJCNN),LGMD1-Silva1}, (f) an LGMD1 neural network based on the modelling of elementary motion detectors for collision detection in ground vehicle scenarios, adapted from \cite{LGMD-car-2017(bionic-vehicle-collision)}.}
	\label{Fig: LGMD1-other-models}
\end{figure}

\begin{figure}[H]
	\centering
	\includegraphics[width=0.5\textwidth]{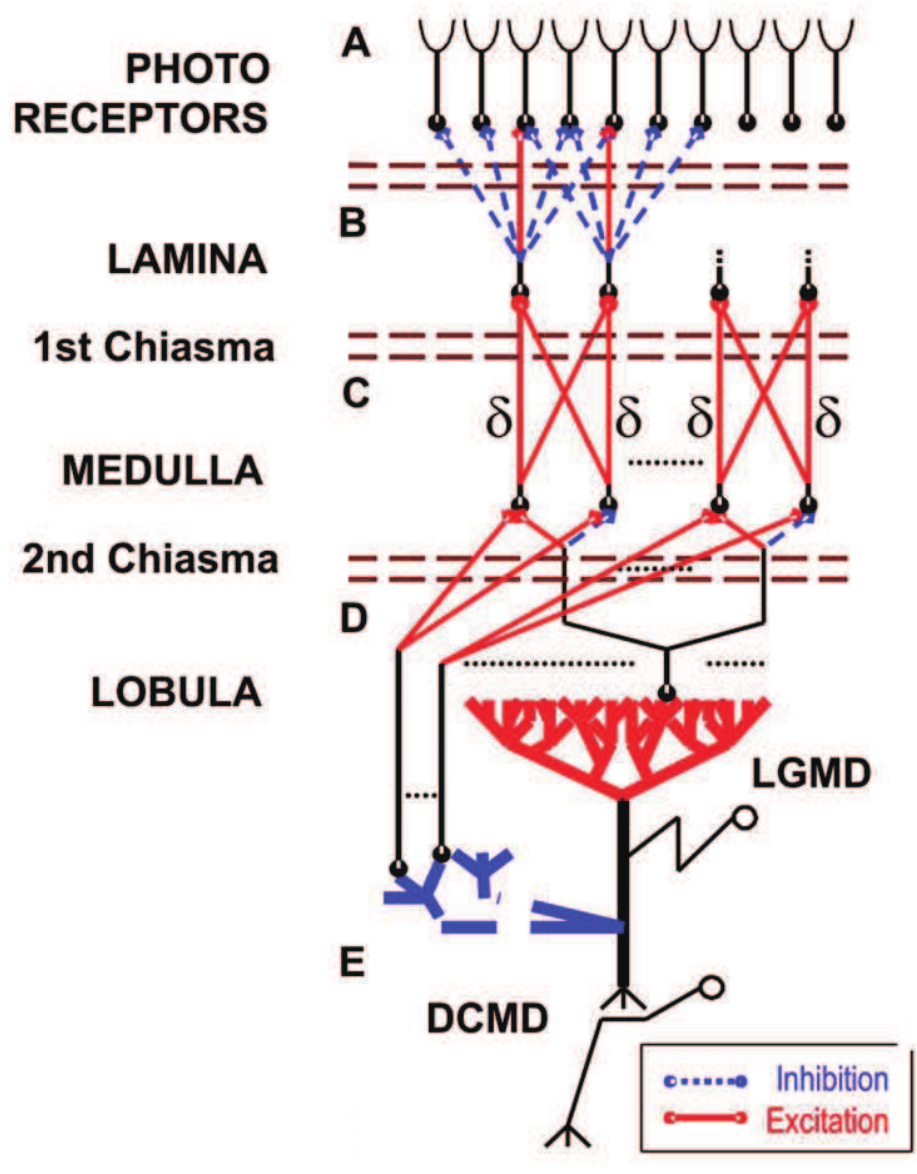}
	\caption{Schematic of a non-linear and multi-layered LGMD1 visual neural network, adapted from \cite{Badia-2010(LGMD1-nonlinear-model)}: this modelling study emphasises the non-linear interactions between the feed-forward excitation and the feed-forward inhibition.}
	\label{Fig: LGMD1-Badia2010}
\end{figure}

Based on this LGMD1 modelling theory, a good number of models have been produced during the past two decades; these works have not only been extending and consolidating the LGMD1's original functionality for looming perception, but also investigating the possible applications to mobile machines like robots and vehicles.
More specifically, the proposed LGMD1 neuronal model by Rind \cite{LGMD1-1996(Rind-neural-network)} was for the first time implemented in a ground mobile robot for collision detection in two seminal works \cite{LGMD1-2000(model),LGMD1-2001(sensory-coding-robot)}.
Rind further demonstrated the usefulness of LGMD for guiding flying robot behaviour and pointed out another group of directional selective neurons that sense ego-motion could be integrated with the LGMD for better the collision-detecting performance \cite{Rind2002(locust-biology-robot)}.
In addition, Yue et al. applied the LGMD1 model as an optimised collision-detecting solution for cars; a novel genetic algorithm was for the first time incorporated in this neuronal model for improving collision detection performance in driving scenes \cite{Yue-2006(LGMD1-car)}.
After that, Yue and Rind developed new mechanisms in the LGMD1-based visual neural network, to enhance the ability of extracting looming features from complex and dynamic environments and adapting to different illuminations \cite{LGMD1-Glayer(feature-enhancement)}, as illustrated in the Fig. \ref{Fig: LGMD1-Glayer}. 
This method was verified by a vision-based mobile ground robot \cite{Yue-2005(LGMD1-ICRA),LGMD1-Glayer(feature-enhancement)} with better performance in collision detection. 
Compared with previous bio-robotic studies, the robot agent can recognise potential collision tested by different speeds and light conditions, as shown in the Fig. \ref{Fig: LGMD1-Glayer-robot}.
With similar ideas, Yue and Rind continued exploring the potential of LGMD1 model in robotic applications like near range path navigation; these works include a development of a visually guided control with a bilateral pair of LGMD1-DCMD neurones for a reactive collision avoidance strategy \cite{Yue-2010(LGMD1-robot-bilateral),LGMD1-Yue2009(near-range-navigation)} (Fig. \ref{Fig: LGMD1-Yue-robot2009}).
Hu et al. applied a similar LGMD1 visual neural network as an embedded vision system in a vision-based autonomous micro-robot to demonstrate its computational simplicity for in-chip visual processing \cite{Hu-2014(LGMD1-ICDL),Hu-2017(Colias-LGMD1)}.
To verify its validity and reliability, the miniaturised robot with on-board LGMD1 processing was tested in an arena mixed with many obstacles, as the results shown in the Fig. \ref{Fig: LGMD1-Cheng-arena}. 
The results demonstrated very high success rate of collision detection tested by different speeds of robot and densities of obstacles.
Very recently, the similar approach has been implemented in a hexapod walking robot \cite{LGMD1-walking-robot} and a small quad-copter for collision avoidance in short-range navigation \cite{UAV-2018(LGMD1-Opticflow-PID)}.

Moreover, there are many derivatives of the proposed LGMD1-based neural network by Rind \cite{LGMD1-1996(Rind-neural-network)}, as illustrated in the Fig. \ref{Fig: LGMD1-other-models}. 
These computational models consist of new methods to enhance the collision selectivity to approaching objects \cite{Meng-2009(IJCNN-LGMD1)}, new layers to reduce environmental noise \cite{LGMD1-2013(Silva-IJCNN),LGMD1-Silva1}, and etc.
There are also researches on corresponding applications for cars \cite{LGMD1-car-2011(risk-collision-road),LGMD-car-2017(bionic-vehicle-collision)} and mobile robots \cite{LGMD1-2007(obstacle-avoidance-robot)}, as well as implementations in hardware like field-programmable gate array (FPGA) \cite{Meng-2010(LGMD1-FPGA)}.

Interestingly, Gabbiani has pointed out that there are many ways to build the looming sensitive neuronal models like the locust LGMD \cite{Gabbiani-1999(many-ways-LGMD)}.
For example, another important theory underlines the non-linearity in the modelling of looming sensitive neurons, that is, the LGMD1 would represent a highly non-linear processing or competition between the inhibitory and the excitatory flows, proposed by Gabbiani et al. \cite{Gabbiani-2002(LGMD-multiplicative-computation),Gabbiani1999(computation-LGMD),Gabbiani2002(multiplicative-computation-LGMD)}.
They have also demonstrated the correspondence between the calculations of feed forward excitation/inhibition and the angular speed/size of looming objects within the field of view. 
Here, the feed forward inhibition features a critical role to shape the collision selectivity of the LGMD1 \cite{Gabbiani2005(feed-forward-inhibition)}.
With respect to the non-linear interactions between the excitations and inhibitions, the LGMD1 neuronal model could possess a biologically plausible invariance to varied shapes, textures, grey levels and approaching angles of looming patterns \cite{Gabbiani-2001(LGMD-invariance-angular),Gabbiani2004(invariance-LGMD)}.

Based on the non-linear theory of modelling the LGMD, Keil gave an insight into the mathematical explanations for the generation of non-linearity in the LGMD neuronal model \cite{Keil-2011(LGMD-algorithm-NIPS),Keil-2015(LGMD-dendritic-noisy)}.
Badia et al. incorporated the non-linear (multiplicative) elementary motion detectors (EMDs) to construct the LGMD1 for sensing and avoiding potential collision \cite{LGMD1-2004(Badia)}.
Stafford et al. also applied similar strategies to model the LGMD1 for handling looming perception in driving scenarios \cite{LGMD1-car2007(collision-detection-cars)}.
In addition, as illustrated in the Fig. \ref{Fig: LGMD1-Badia2010}, a non-linear LGMD1 visual neural network was proposed by Badia et al. \cite{Badia-2010(LGMD1-nonlinear-model)}; the functionality of this model fits well the non-linear properties of an LGMD1 neuron given by Gabbiani \cite{Gabbiani2002(multiplicative-computation-LGMD)}, and it possesses the invariance of collision detection to looming stimuli with varied shapes, textures and approaching angles \cite{Gabbiani2004(invariance-LGMD)}.
Importantly, they also demonstrated that the LGMD1 model can encode onset and offset response depending on luminance increments and decrements that bring about different delayed information of excitations and inhibitions, after a seminal work with ON and OFF mechanisms proposed in \cite{LGMD1-ONOFF-2004}.
Moreover, this model has been successfully implemented in a mobile ground robot performing well in an arena for collision detection in near range navigation.

\subsection{LGMD2-based neuron models and applications}
\label{section: literature: looming: lgmd2}

\subsubsection{Characterisation}

\begin{figure}[H]
	\centering
	\subfloat[]{\includegraphics[width=0.38\textwidth]{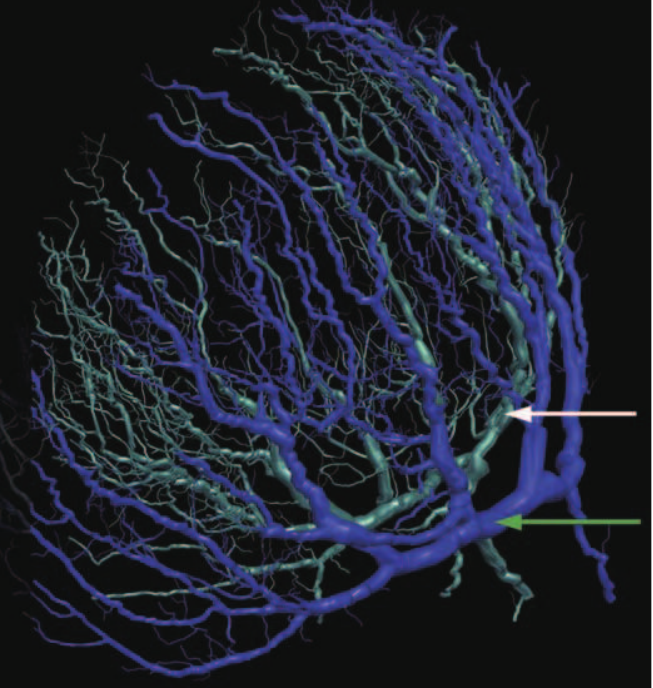}
		\label{LGMDs-neuro-trees}}
	\hfil
	\subfloat[]{\includegraphics[width=0.5\textwidth]{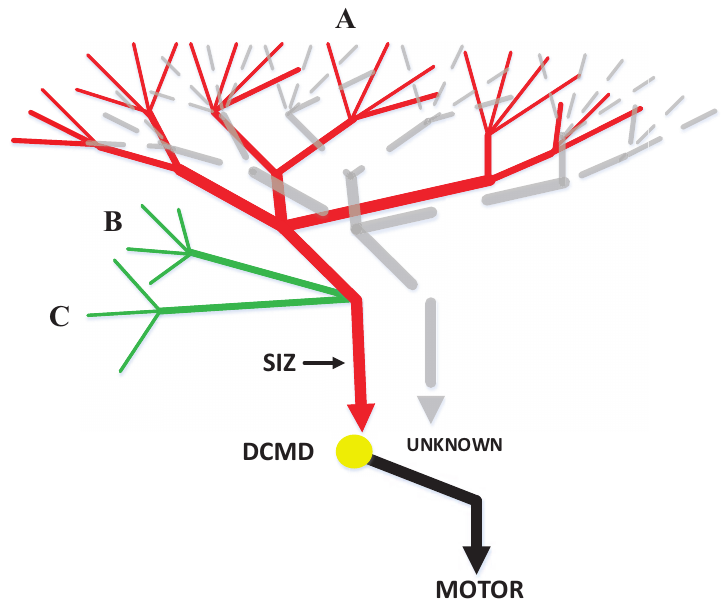}
		\label{LGMDs-neuro-schematics}}
	\caption{Neuromorphology of the LGMD1 and the LGMD2: (a) 3D reconstruction of dendritic trees of LGMD1 and LGMD2 indicated by white and green arrows, respectively, adapted from \cite{Sztarker2014(LGMD2-development)}, (b) a schematic diagram of both the pre-synaptic and the post-synaptic areas of LGMD1 (red) and LGMD2 (grey), adapted from \cite{Fu-2018(LGMD1-NN)}.}
	\label{Fig: LGMD2-neuromorphology}
\end{figure}

The LGMD2 is a neighbouring partner to the LGMD1 also as a looming detector. 
It has similar characteristics but different selectivity to the LGMD1 \cite{Simmons-1997(LGMD2-neuron-locusts),LGMDs-2016,Sztarker2014(LGMD2-development),LGMD1-1996(Rind-intracellular-neurons),DCMD-1997(collision-trajectories),Rind2000(bio-evidence),DSNs-1990(Rind-locust-DSNs)}.
On the aspect of neuromorphology, as illustrated in the Fig. \ref{Fig: LGMD2-neuromorphology}, the LGMD2 has also large fan-shaped dendrite trees within its pre-synaptic area (Fig. \ref{LGMDs-neuro-trees}); however, comparing to the LGMD1, the lateral fields (B, C in the Fig. \ref{LGMDs-neuro-schematics}) that convey feed forward inhibitions are not found to the LGMD2, and moreover the post-synaptic target neuron to the LGMD2 has not been explored so far.
Importantly, a physiological study has demonstrated the development of both neurons in locusts, from adolescence to adulthood: the LGMD2 matures earlier in juvenile locusts that lack wings and live mainly on the ground \cite{Sztarker2014(LGMD2-development)}.
As a result, the LGMD2 plays crucial roles in juvenile locusts for perceiving predators and likely leads to hiding behaviours against looming stimuli \cite{Sztarker2014(LGMD2-development),Fu-2016(LGMD2-BMVC)}.

\begin{figure}[H]
	\centering
	\subfloat[]{\frame{\includegraphics[width=0.5\textwidth]{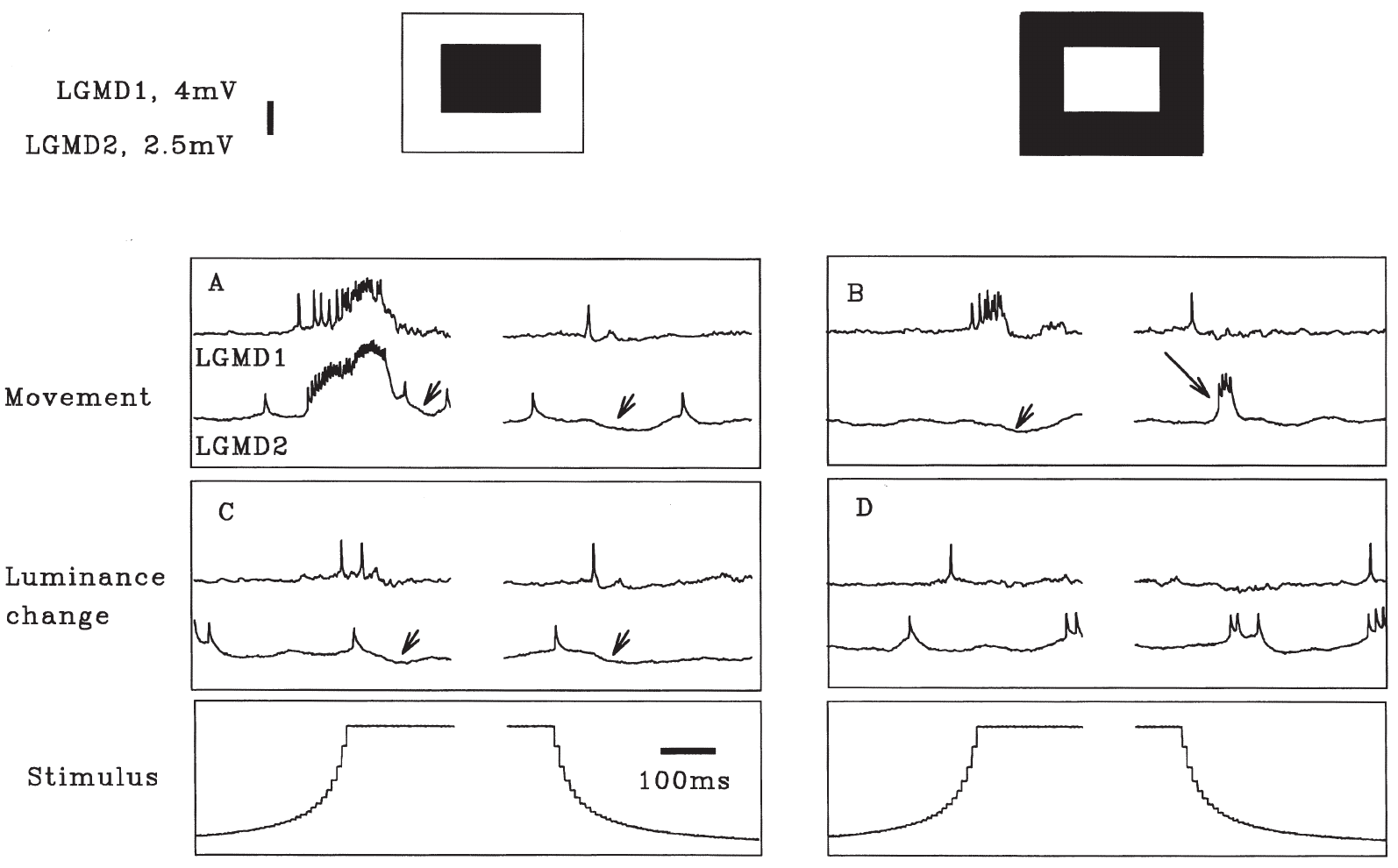}}
		\label{Fig: exp: bio-data: looming}}
	\hfil
	\subfloat[]{\frame{\includegraphics[width=0.4\textwidth]{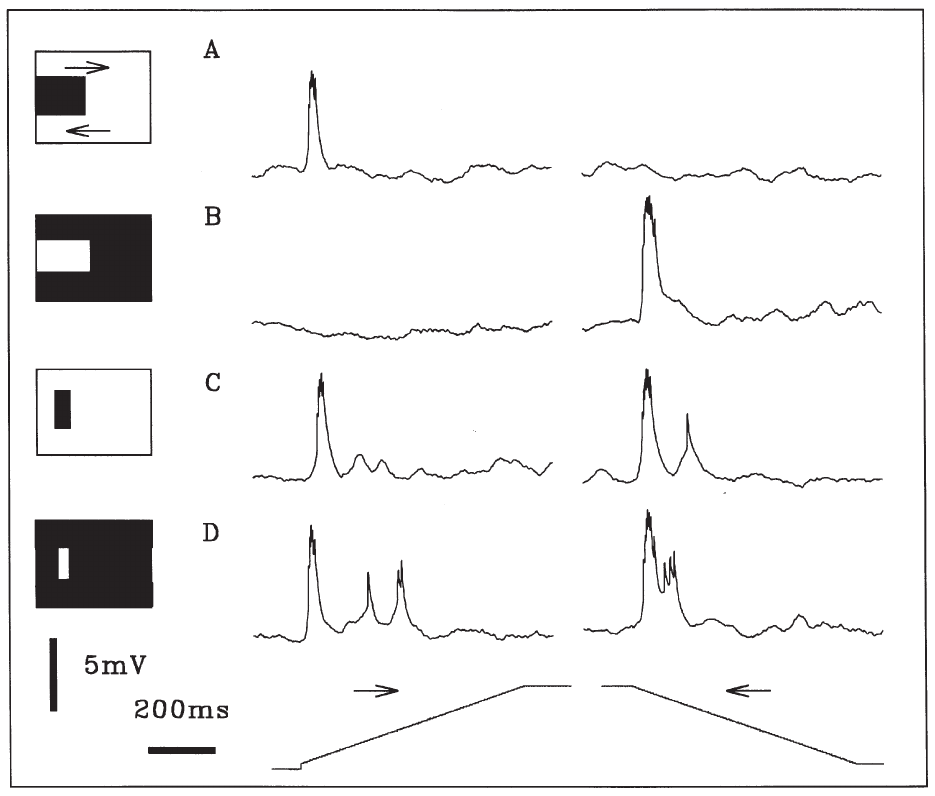}}
		\label{Fig: exp: bio-data: translating}}
	\caption{Biological data of LGMD2 neuron responses to (a) dark and light objects looming and (b) translating stimuli, adapted from \cite{Simmons-1997(LGMD2-neuron-locusts)}.}
	\label{Fig: lgmds-response-looming-translating}
\end{figure}

More precisely, the specific living environments for young locusts endow the LGMD2 a particular neural characteristic, that is, it is only sensitive to dark looming objects within bright background whilst not responding to white or light objects approaching within dark background, which represent a preference to the light-to-dark luminance change. 
Moreover, the biological functions of LGMD2 differ from the LGMD1 in a number of ways. 
First, the LGMD2 is not sensitive to a light or white looming object whereas the LGMD1 is. 
Second, the LGMD2 does not respond to dark objects that recede at all, while the LGMD1 is often excited though very briefly \cite{Simmons-1997(LGMD2-neuron-locusts)}. 
Both the LGMD1 and the LGMD2 responds shortly to translation regardless of the size and the direction of moving objects \cite{Simmons-1997(LGMD2-neuron-locusts)}. 
They are neither sensitive to wide-field luminance change and grating movements \cite{Simmons-1997(LGMD2-neuron-locusts)}. 
The Fig. \ref{Fig: lgmds-response-looming-translating} clearly illustrates these features. 
Furthermore, to investigate the place where the looming selectivity forms in such neurons, Rind et al. recently looked into the pre-synaptic neuropile layer of Medulla in the locust's visual brain. 
This research proposed that the specific looming selectivity of both LGMDs may generate within the pre-synaptic fields \cite{LGMDs-2016}: the lateral-and-self inhibition mechanism works effectively to form the selectivity.

\begin{figure}[H]
	\centering
	\includegraphics[width=0.7\textwidth]{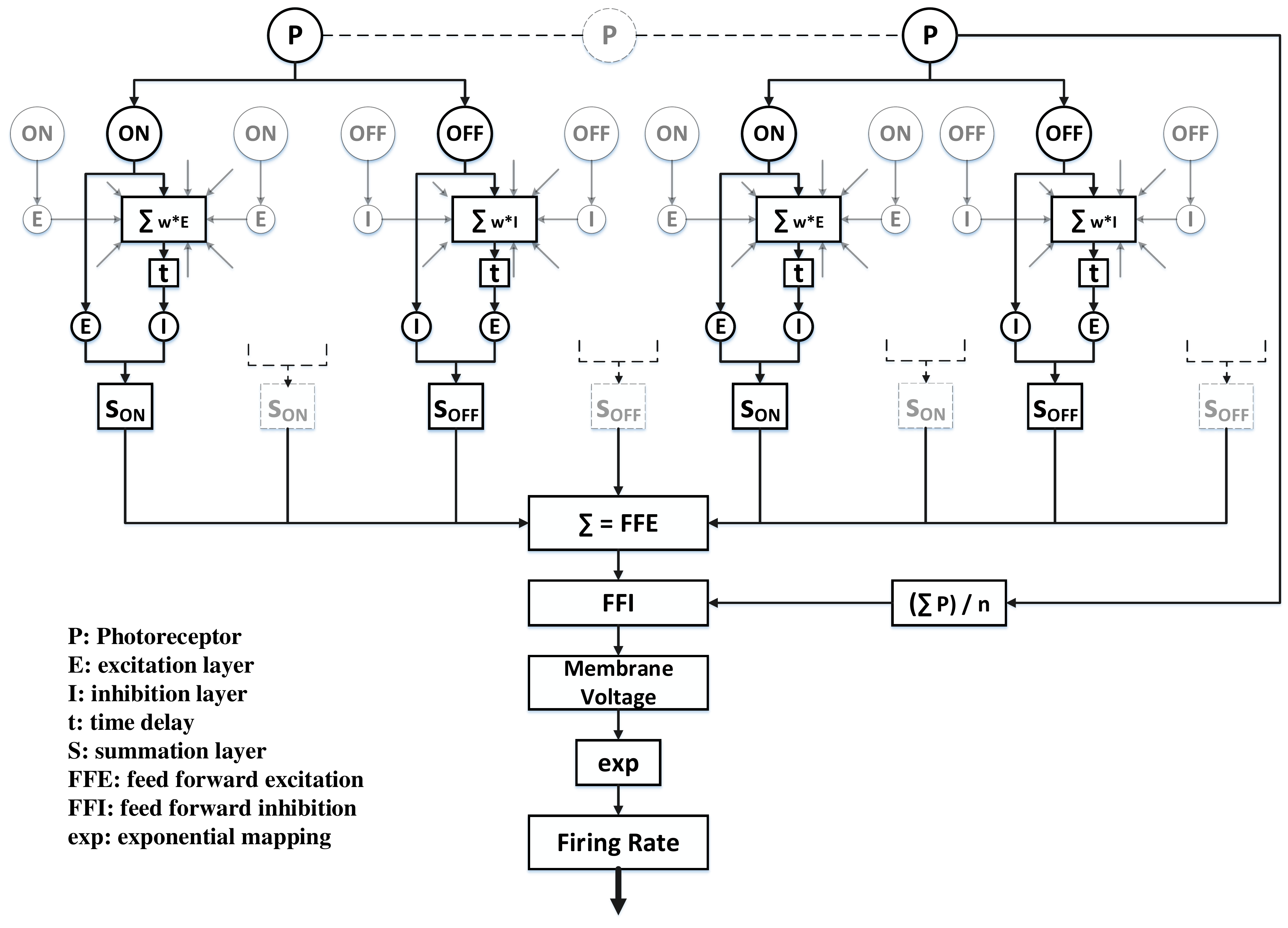}
	\caption{Schematic of a seminal LGMD2-based visual neural network, adapted from \cite{Fu-2015(LGMD2-MLSP),Fu-2016(LGMD2-BMVC)}:
		the model processes visual information with ON and OFF mechanisms that encode brightness increments and decrements, separately: the ON channels are rigorously suppressed to realise the LGMD2's specific looming selectivity to dark objects.
		In this model, the excitations are delayed in the OFF pathway and the inhibitions are delayed in the ON pathway, due to the ON (onset) and OFF (offset) mechanisms.
	}
	\label{Fig: LGMD2-models}
\end{figure}

\subsubsection{Computational models and applications}

\begin{figure}[H]
	\centering
	\includegraphics[width=0.9\textwidth]{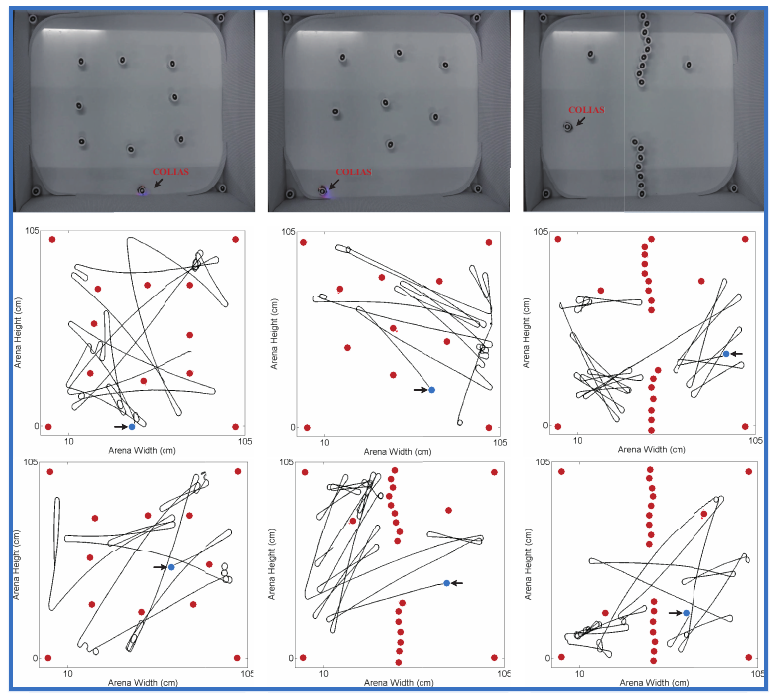}
	\caption{Micro-robot arena tests implemented with an LGMD2 model: the robot agent was tested by different layouts and densities of obstacles in an arena. 
		The black lines indicate robot overtime trajectories and the red circles denote the obstacles. 
		The blue circles indicate the start position of robot agent. 
		The experimental data is adapted from \cite{Fu-2016(LGMD2-BMVC)}.}
	\label{Fig: LGMD2-arena}
\end{figure}

\begin{figure}[H]
	\centering
	\subfloat[]{\includegraphics[width=0.7\textwidth]{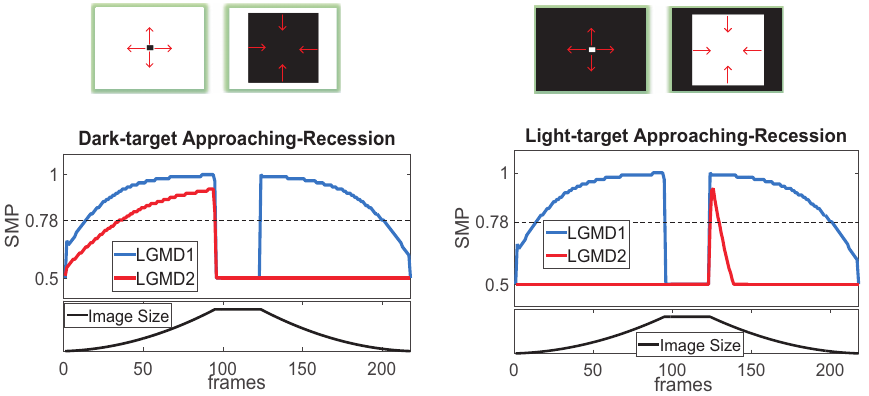}
		\label{LGMD2-response}}
	\vfil
	\vspace{-0.1in}
	\subfloat[]{\includegraphics[width=0.7\textwidth]{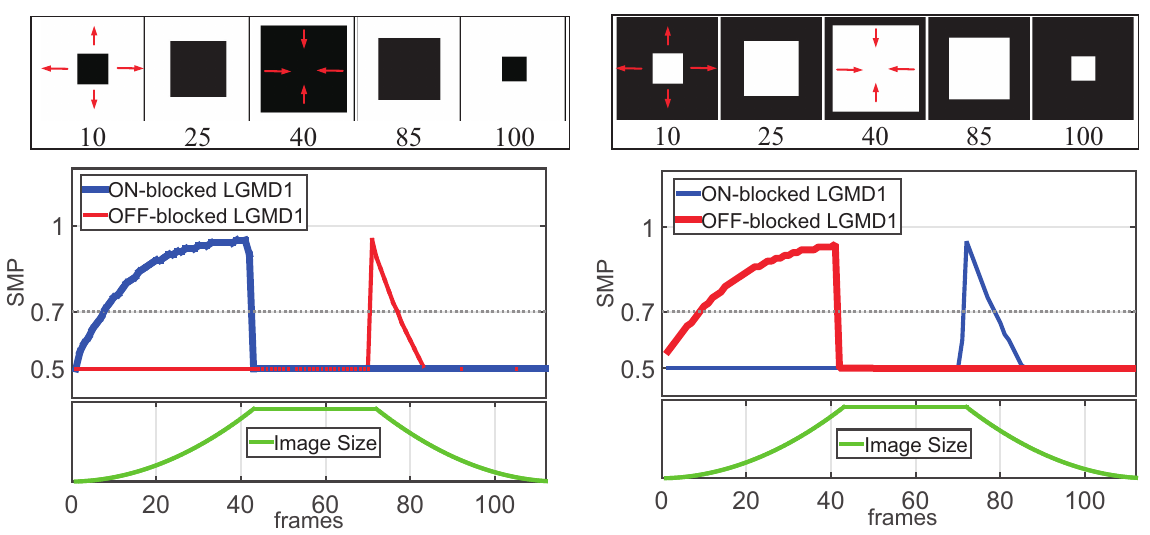}
		\label{LGMDs-shape}}
	\caption{Results of shaping the selectivity between the LGMD1 and the LGMD2 neuronal models via the modelling of ON and OFF pathways and spike frequency adaptation: (a) the LGMD2 responds to dark and light approach-recession movements compared with an LGMD1 model \cite{LGMD1-Glayer(feature-enhancement)}, adapted from \cite{Fu-2018(LGMD1-NN)}, (b) the effects of shaping the selectivity through blocking either the ON or the OFF pathways, adapted from \cite{Fu-2018(LGMD1-NN)}.
	The horizontal dashed lines indicate the spiking threshold.
	The image size change within the field of view is depicted at each bottom.}
	\label{Fig: LGMD2-response-shape}
\end{figure}

For computationally modelling the LGMD2, only a handful of studies have been proposed so far.
A seminal work appeared in the 2015: Fu and Yue proposed an LGMD2-based visual neural network to implement an LGMD2 in a vision-based micro-robot with similar selectivity to the light-to-dark luminance change via the modelling of ON and OFF mechanisms \cite{Fu-2015(LGMD2-MLSP)}. 
This model separates luminance change into parallel channels and encodes the excitations and the inhibitions via spatiotemporal computation similarly to the LGMD1 model \cite{Hu-2017(Colias-LGMD1)}, but with different delayed information, as illustrated in the Fig \ref{Fig: LGMD2-models}. 
More precisely, in order to achieve the specific looming selectivity of LGMD2 to dark objects only, the ON channels are rigorously sieved; the ON and OFF mechanisms also bring about different temporally delayed information in each separate pathway.
The effectiveness and flexibility of this LGMD2-based visual neural network has been validated by arena tests of an autonomous micro-robot \cite{Fu-2016(LGMD2-BMVC)}, as illustrated in the Fig. \ref{Fig: LGMD2-arena}.
For the first time, the specific functionality of an LGMD2 neuron revealed by biologists has been realised in a computational structure. 
Compared with all the aforementioned LGMD1 models, it only responds to dark looming objects and briefly to the receding of light objects, representing a preference to the light-to-dark luminance change, as depicted in the Fig. \ref{LGMD2-response}.

\begin{figure}[H]
	\centering
	\includegraphics[width=\textwidth]{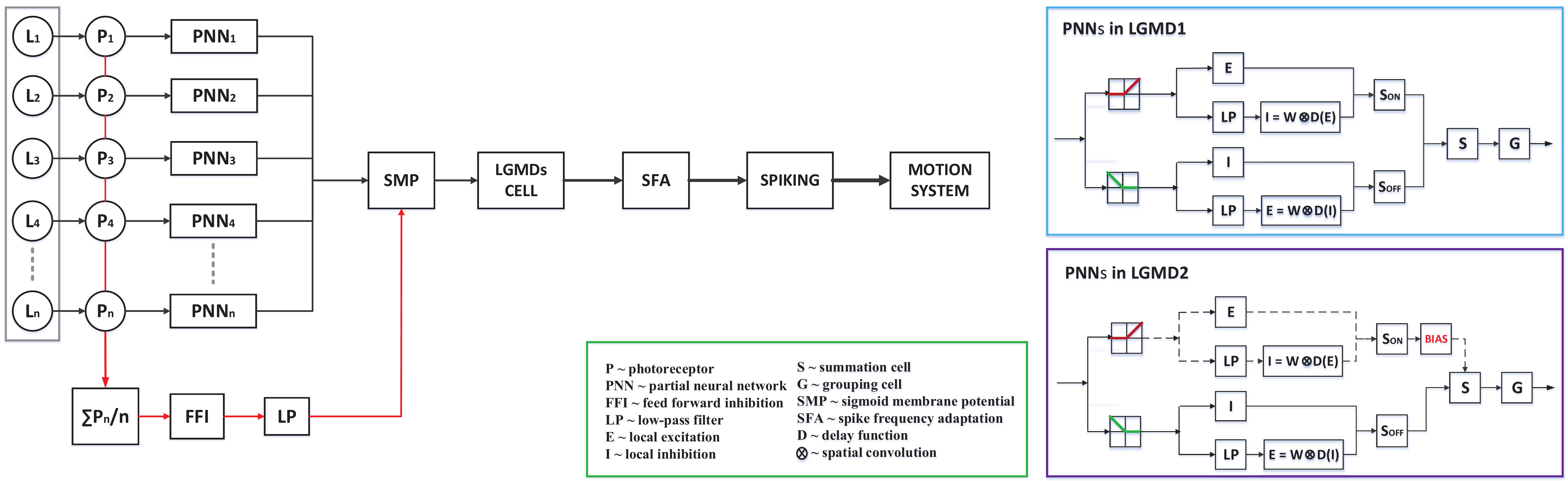}
	\caption{Schematic of a general LGMDs model adapted from \cite{Fu2017a(LGMDs-IROS)}: in this visual neural network, the functionality of ON and OFF pathways and a spike frequency adaptation mechanism are modelled. 
		This model can realise the characteristics of both the LGMD1 and the LGMD2 with different bias in the partial neural networks.}
	\label{Fig: LGMDs-generic-model}
\end{figure}

For further investigating the different looming selectivity between the LGMD1 and the LGMD2, Fu et al. proposed a hybrid visual neural model, which was smoothly implemented in an autonomous micro-robot for collision detection in an environment containing multiple dynamic robot agents \cite{Fu2017a(LGMDs-IROS)}. 
In this research, both the LGMDs were implemented in the robot agents and tested in both dark and bright environments. 
Each looming sensitive neuron handles a half region of the field of view for a bilateral control of robot reactive avoidance behaviours \cite{Fu2017a(LGMDs-IROS)}. 
This study has verified the effectiveness of such a strategy for guiding timely collision avoidance of mobile robots, and more importantly, demonstrated that the ON and OFF pathways would be a crucial structure for separating the different looming selectivity between the LGMD1 and the LGMD2. 
Although there is little physiological evidence showing the existence of such polarity pathways in the locust's visual systems \cite{LGMD-1976(ON-OFF),LGMD1-synaptic2015(first-stage-locust)}, the proposed computational models could evidence similar mechanisms in looming sensitive neural circuits or pathways \cite{Fu-2018(LGMD1-NN)}.

\subsection{Different mechanisms to mediate the looming selectivity}
\label{section: literature: looming: selectivity}

The looming selectivity to moving objects that approach over other kinds of visual challenges has been formed very well in locusts through millions of years of evolutionary development.
As a result, the locust LGMDs spike most strongly to looming stimuli. 
However, from the perspective of computational modellers, it is still an open challenge to shape the selectivity in looming perception neuron models. 
Though these state-of-the-art models can be applied as quick and efficient collision detectors, the performance, compared with the locusts, is far from acceptable, since they are greatly affected by some irrelevant motion patterns like recession and translation. 
They are also restricted by the complexity of background in real-world visual clutter. 
In the future, artificial machines should possess similar ability like the locusts with efficient and robust collision-free systems to better serve the human society.

There have been a few methodologies proposed to mediate the looming selectivity.
Rind et al. demonstrated that two kinds of inhibitions -- the pre-synaptic lateral inhibitions and the feed forward inhibition can cooperatively mediate the looming selectivity by cutting down the excitation caused by receding and translating stimuli \cite{LGMD1-1996(Rind-neural-network)}.
Gabbiani et al. presented the non-linear computations make the neuron liable to differentiate looming from receding stimuli \cite{Gabbiani2004(invariance-LGMD)}.
In addition, they revealed an intrinsic neural property of such looming sensitive neurons, that is, the spike frequency adaptation (SFA), which leads the LGMD to discriminate objects that approach from recession and translation \cite{SFA-2009,SFA-2009Role,Gabbiani2006(SFA-LGMD)}. 
As mentioned above, these methods have been implemented in mobile ground robots for collision detection in near range navigation, e.g. \cite{LGMD1-Glayer(feature-enhancement),Yue-2005(LGMD1-ICRA),Yue-2010(LGMD1-robot-bilateral),LGMD1-Yue2009(near-range-navigation),Badia-2010(LGMD1-nonlinear-model),Hu-2017(Colias-LGMD1)}. 
However, these theories have not been validated by challenge in more complex real-world scenarios, though some of them have been casually touched upon \cite{LGMD1-car2007(collision-detection-cars),LGMD1-car-2011(risk-collision-road),Yue-2006(LGMD1-car)}.

In addition, for computationally modelling these mechanisms, there is a trade-off between the algorithmic efficiency and the efficacy.
Computational modellers have always been trying to balance both, in order to achieve reliable and efficient performance in intelligent machines like mobile robots and vehicles.
Yue and Rind proposed a hybrid neural system incorporating a translating sensitive neural network, in order to extract colliding information from mixed motion cues \cite{Yue-2006(LGMD1-TSNN)}.
This method is effective in some driving scenarios; it nevertheless costs more computational power than an LGMD model alone.
Meng et al. designed an organisation of the LGMD1's post-synaptic field to monitor the gradient change of model output for discriminating approach from recession \cite{Meng-2009(IJCNN-LGMD1)} (Fig. \ref{LGMD1-Meng2010}). 
This method was smoothly implemented in the FPGA \cite{Meng-2010(LGMD1-FPGA)}; however such a structure is not biological reasonable. 
Moreover, a neural network of directional motion-detecting neurons in locusts was integrated with the LGMD1 neural network to ensure the recognition of imminent collision in some driving scenarios \cite{Zhang-2016(IJCNN-hybrid)}. 
In this research, the field of view was divided into different regions processed by specialised neurons, separately. 
More recently, Fu et al. has demonstrated efficacy of combining two bio-plausible mechanisms -- ON and OFF pathways and spike frequency adaptation to enhance the required selectivity of both the LGMD1 and the LGMD2 models \cite{Fu2017a(LGMDs-IROS),Fu-2018(LGMD1-NN)}. 
These models have been validated by bio-robotic tests on the embedded system.
To be more intuitive, some example results are illustrated in the Fig. \ref{LGMDs-shape} to clarify the effects of separating visual processing in parallel pathways in a computational structure to achieve different looming selectivity for detecting either dark or light objects. 
Notably, the neuron model with either ON or OFF pathway blocked is only briefly activated by light or dark object moving away.

\subsection{Further discussion}
\label{Sec: Looming: Discussion}

Within this section, we have reviewed the computational models and applications originating from locust visual systems research throughout the past several decades.
These computational models have been demonstrated effectiveness and flexibility for collision detection in some mobile machines, which shed light on building robust collision-detecting neuromorphic sensors for future intelligent machines for collision detection in both a cheap and reliable manner.

Biologists have also explored similar looming sensitive visual neurons in other animals; these include fruit flies (drosophila) \cite{FlyLooming-2012,Escape-2012(neural-computation-action),landing-2002,Science-2014(fly-banked-turn)} and arthropods like crabs \cite{Crab-2007(LGN-visual-escape)}. 
For instance, the lobula giant neurons (LGNs) in crabs have been identified as looming detectors that are located in the lobula layer and correspond to reactive collision avoidance behaviours \cite{Crab-2007(escape-from-looming),Crab-2008(neuronal-correlates-escape),Crab-2014(organization-columnar-visual)}.
The possible computational roles of such visual neurons have also been proposed in \cite{Crab2014(computation-approach-neurons)}.
However, there are no systematic studies on the modelling and applications of such fascinating looming detectors in crabs. 
Though the LGNs have different neuromorphology compared to the LGMDs, the computationally modelling of LGNs may learn from the practical experience of existing LGMDs models.

\section{Translation perception neural systems}
\label{Sec: Translational-Motion}

This section reviews computational models and applications of translation sensitive motion detectors and neural networks inspired by the insect visual neurons and pathways. 
First, the modelling of directionally selective motion-detecting neurons in locusts, namely the locust direction selective neurons (DSN) will be introduced in the Section \ref{section: literature: translating: locust-dsn}. 
Then, we will review a classic model category of fly elementary motion detector (EMD) and corresponding applications to robotics in the Section \ref{section: literature: translating: fly-emd}. 
After that, the cutting-edge biological findings and computational models of fly ON and OFF pathways and lobula plate tangential cells (LPTC), namely the fly DSNs, will be presented in the Section \ref{section: literature: translating: fly-lptc}.

Compared to the `non-directional' neurons such as the looming sensitive LGMD1 and LGMD2, the research on the DSN in animals has a much longer history; it can be even dated back to two centuries ago. 
Franceschini pointed out that an initial idea of `directionally selective motion sensitive cells' was proposed by Exner early in the 1894 \cite{Nicolas-1989(DSN-Insect-Neurons)}. 
The past several decades have witnessed much physiological progress in our understanding of the cellular mechanisms underlying the DS. 
More specifically the DSNs have been found in many animal species. 
These include invertebrates like flies \cite{Borst-2002(review-networks-fly)}, locusts \cite{DSNs-1990(Rind-locust-DSNs)}, as well as vertebrates like rabbits \cite{Barlow-1965(rabbit-DSNs)}, mice \cite{Borst2015(common-circuit-motion)}. 
Borst demonstrated the similarities of circuits and algorithms in design of insect and vertebrate motion detection systems for translational motion perception \cite{Borst2015(common-circuit-motion)}. 
Generally speaking, this field of research has been attracting much more attention from cross-disciplines. 
Here in this article, we will focus on presenting some milestone biological theories on translation perception neurons and pathways, and corresponding models with bio-robotic applications.

\begin{figure}[H]
	\centering
	\subfloat[locust DSNs neural network]{\includegraphics[width=0.35\textwidth]{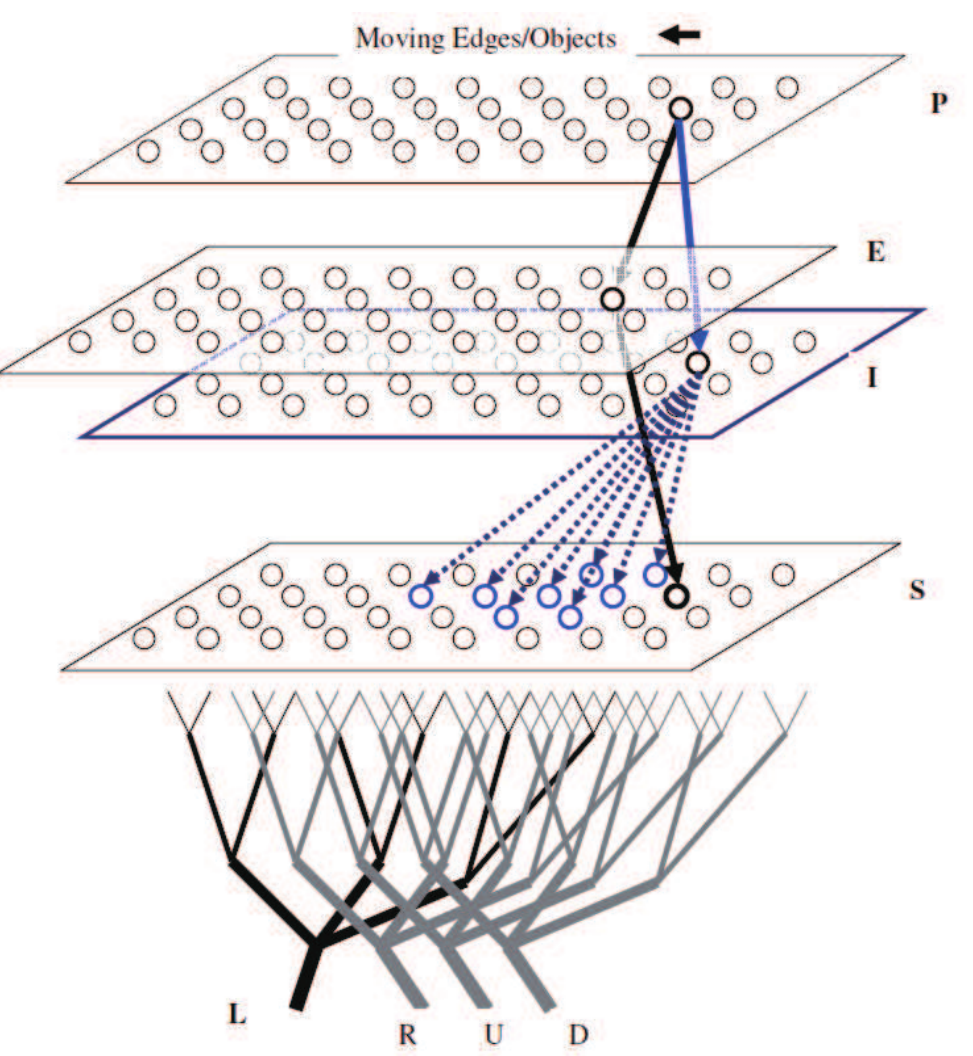}
		\label{DSNs-neural-network}}
	\hfil
	\subfloat[four DSNs responses to leftward translation]{\includegraphics[width=0.6\textwidth]{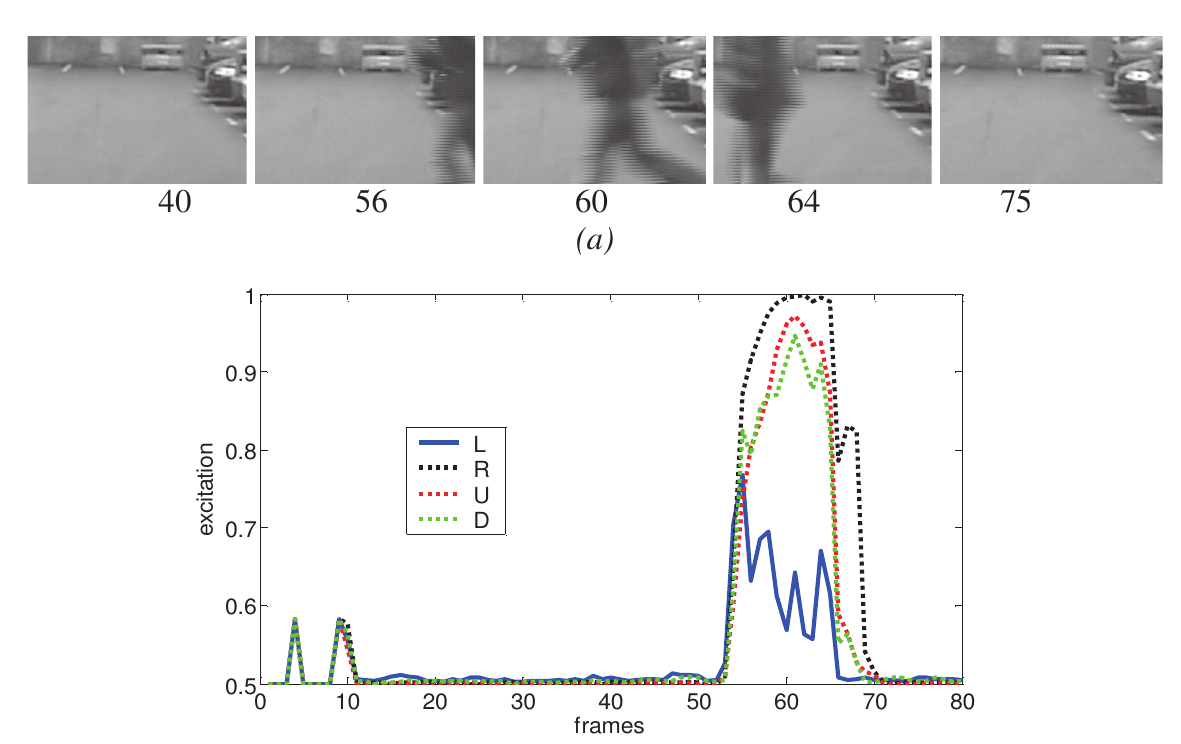}
		\label{DSNs-response}}
	\vfil
	\vspace{-0.1in}
	\subfloat[lateral inhibition structure]{\includegraphics[width=0.38\textwidth]{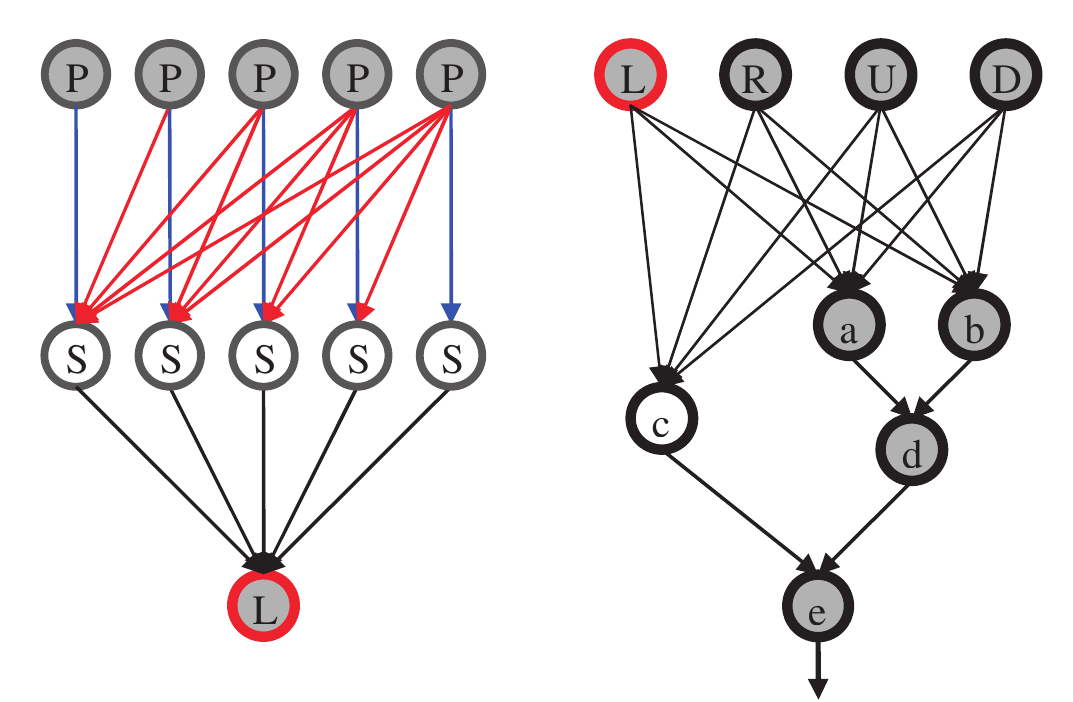}
		\label{DSNs-connection}}
	\hfil
	\subfloat[post-synaptic neurons organisation]{\includegraphics[width=0.6\textwidth]{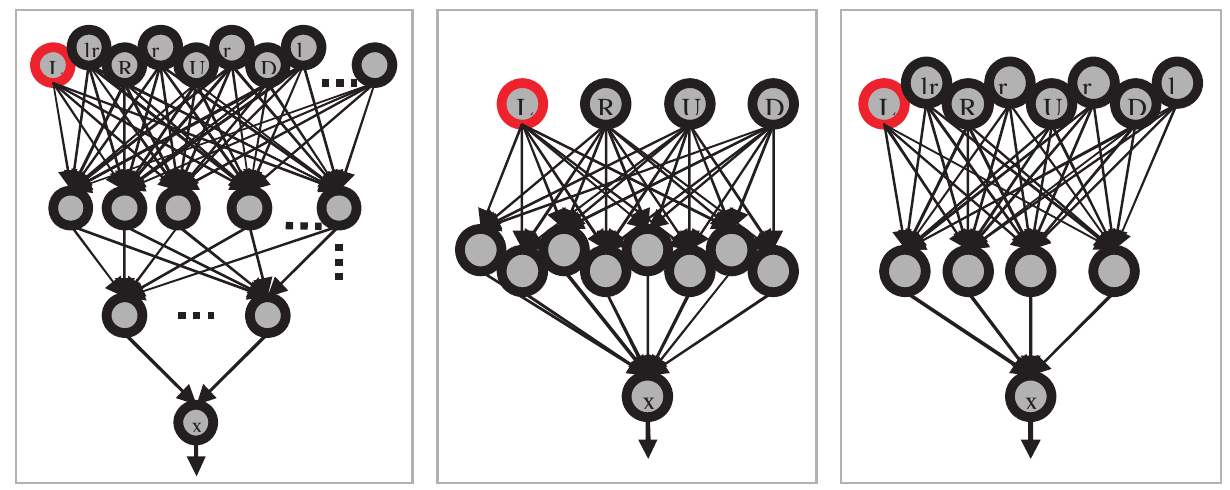}
		\label{DSNs-PNs}}
	\caption{Schematics of the locust DSNs based visual neural networks with both pre-synaptic and post-synaptic structures, adapted from \cite{Yue-2007(locust-DSNs),Yue-2013(locust-DSNs)}.}
	\label{Fig: Locust-DSNs-models}
\end{figure}

\subsection{Computational models of the locust DSNs}
\label{section: literature: translating: locust-dsn}

For locusts, Rind explored the characteristics of DSNs in the 1989, in both physiology and morphology \cite{DSNs-1990(Rind-locust-DSNs),Rind-1990(locusts-DSNs-morphology)}. 
These neurons respond to horizontal directional motion: they are rigorously excited by translation in preferred direction (PD) and inhibited by movements in non-preferred direction (ND).

On the basis of these biological findings, a few computational models have been proposed by Yue and Rind \cite{Yue-2007(locust-DSNs),Yue-2013(locust-DSNs)}. 
Generally speaking, these translation sensitive visual neural networks have been modelled with exquisite organisation of different post-synaptic DSNs for collision recognition, especially in driving scenarios. 
Interestingly, as illustrated in the Fig. \ref{Fig: Locust-DSNs-models}, these locust DSN models arise from the LGMD1-based visual neural networks: the computational structure of DSN looks similar to the LGMD1; they nevertheless possess different lateral-inhibition mechanisms. 
More concretely, in the LGMD1 neural networks, the inhibitions spread out to the surrounding areas of a corresponding excitatory cell, symmetrically (Fig. \ref{Fig: LGMD1-Glayer}); on the contrary, in the DSNs neural networks, the inhibitions spread out, asymmetrically and directionally, as shown in the Fig \ref{DSNs-neural-network} and \ref{DSNs-connection}. 
Therefore, the specific DS of locust DSNs could have been shaped by such a directional lateral-inhibition mechanism that cuts down local excitations caused by near-by translation. 
For example, with a strong inhibition from the right side with temporal delay, the excitation caused by left moving edges can be eliminated or attenuated (Fig. \ref{DSNs-response}). 
Likewise, each directionally sensitive cell is inhibited by the same directional motion.
With design of post-synaptic architecture, as illustrated in the Fig. \ref{DSNs-connection} and \ref{DSNs-PNs}, this model can detect looming objects, and moreover, recognise the direction of impending collision via activation of specific DSN. 
Furthermore, Yue and Rind extended the functionality of DSNs visual neural network to sense eight directional motion with similar methods, as depicted in the Fig. \ref{DSNs-PNs} \cite{Yue-2013(locust-DSNs)}: in this research, they also investigated the effects of different post-synaptic organisations on collision detection via evolutionary computation.

\begin{figure}[H]
	\centering
	\subfloat[LGMD1 model]{\includegraphics[width=0.26\textwidth]{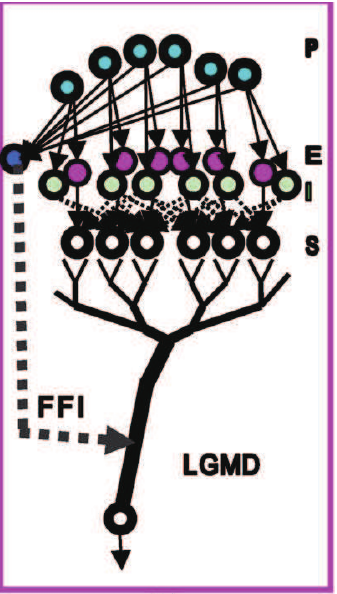}}
	\hfil
	\subfloat[DSNs model]{\includegraphics[width=0.244\textwidth]{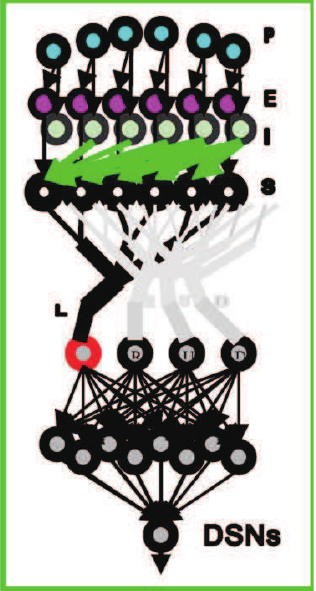}}
	\hfil
	\subfloat[Hybrid model]{\includegraphics[width=0.425\textwidth]{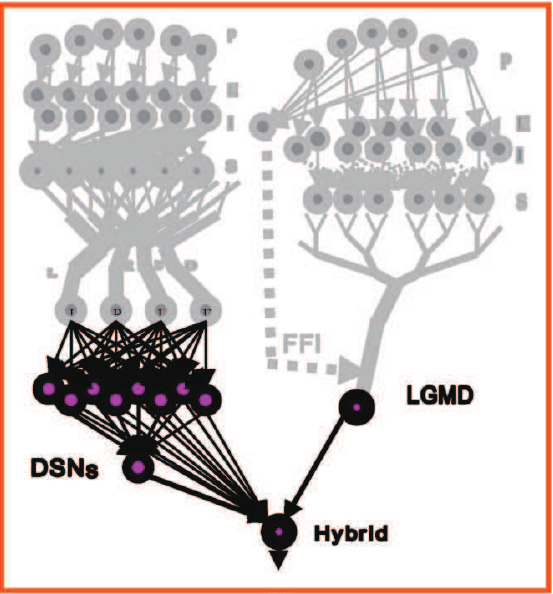}}
	\caption{Schematics of a hybrid visual neural network with three sub-models, i.e., the LGMD alone, the DSNs alone and the hybrid, competing for the collision recognition role via genetic algorithms, adapted from \cite{LGMD1-DSN-competing(LGMD1-DSNs-Hybrid)}.}
	\label{Fig: Locust-hybrid-DSNs-LGMD1}
\end{figure}

It appears that the locust DSNs and LGMDs models are both effective solutions to collision detection. 
A question emerges that which type of visual neurons in locusts could play dominant roles of collision recognition. 
To address this, Yue and Rind designed a hybrid visual neural network integrating the functionality of both neuron models \cite{LGMD1-DSN-competing(LGMD1-DSNs-Hybrid)}, as illustrated in the Fig. \ref{Fig: Locust-hybrid-DSNs-LGMD1}. 
In this research, two individual neural networks competed with the hybrid neural system via a switch gene and evolutionary computation. 
As a result, the LGMD model alone outperforms other candidates for collision recognition due to its computational simplicity and robustness.

\subsection{Fly EMDs models and OF-based strategy}
\label{section: literature: translating: fly-emd}

\subsubsection{Background}

\begin{figure}[H]
	\centering
	\subfloat[Fly compound eyes]{\includegraphics[width=0.4\textwidth]{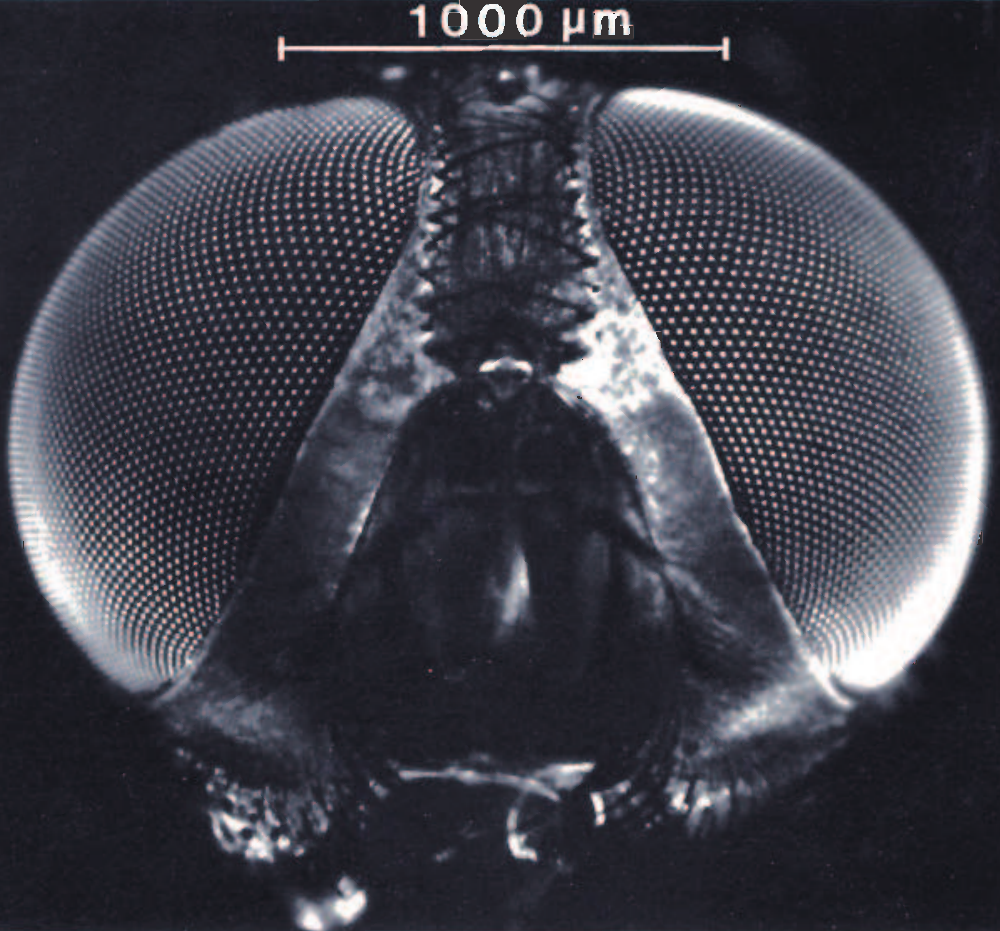}
		\label{Fly-compound-eyes}}
	\hfil
	\subfloat[Fly visual neuropile layers]{\includegraphics[width=0.5\textwidth]{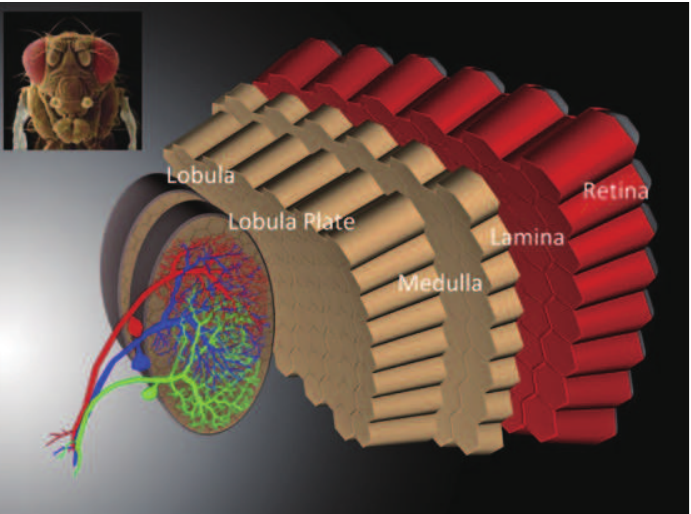}
		\label{Fly-neuro-layers}}
	\caption{Illustrations of fly compound eyes (a) and underlying neuropile layers (b) for motion perception, adapted from \cite{Nicolas-1989(DSN-Insect-Neurons)} and \cite{Borst2011(review-motion)}.}
	\label{Fig: Fly-eyes-morphology}
\end{figure}

The fly visual systems have been researched by a significant number of studies over a century since the first appearance of research early in the 1890s \cite{Nicolas-1989(DSN-Insect-Neurons)}. 
From early stages, the fly visual systems have become prominent model paradigms for studying biological visual processing methodologies and designing artificial motion perception vision systems \cite{HR-1956(EMD-1956),Borst-2002(review-networks-fly),Borst-2010(review-fly-vision),EMD-1989(principles-review),Egelhaaf1993(fly-algorithms-neurons)}. 
With development of biological techniques, the fundamental structures of neuropile layers and cellular implementations within the fly preliminary visual pathways have been better understood by biologists with numerous papers, e.g. \cite{Nicolas-1989(DSN-Insect-Neurons),Fly-1998(dendritic-optic-flow),Fly-2013(motion-circuit-connectomics),Fly2013(brain-visual-motion),Fly-Motion-2017(cellular-hybrid-detector),Fly-2006(feedback-control-photoreceptor),Fly-2013(neuronal-variability),Fly2017(object-detecting-neuron),Fly-Motion-2017(cellular-hybrid-detector)}. 
They have always been attempting to understand the mechanisms underlying motion perception from fly compound eyes, e.g. \cite{Horridge-1977-insect-compound-eye,Nicolas-principle-fly-compound-eye,Nicolas-fly-retinal-mosaic}, to internal pathways and neurons, e.g. \cite{Fly-DS-2016(preferred-null),Fly-DS-2017(emergence-direction-selectivity),Fly-DS-2017(preferred-null),Fly-Motion-2017(cellular-hybrid-detector),Fly2016(direction-selectivity-fly)}. 
The Fig. \ref{Fig: Fly-eyes-morphology} illustrates the fly compound eyes and the underlying neuropile layers of preliminary visual pathways for motion perception.

\begin{figure}[H]
	\centering
	\subfloat{\includegraphics[width=0.41\textwidth]{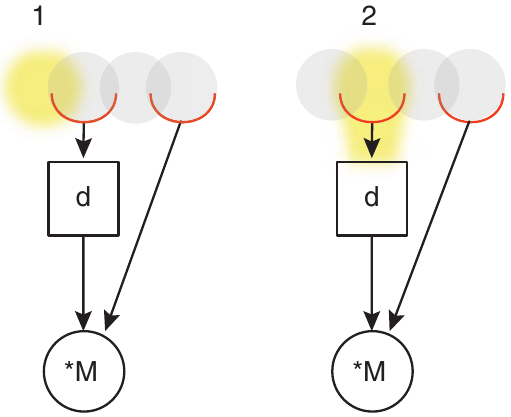}}
	\hfil
	\subfloat{\includegraphics[width=0.38\textwidth]{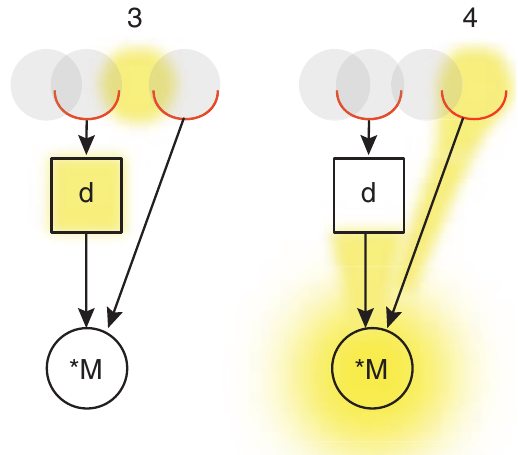}}
	\caption{Schematic of an EMD process in four steps: $d$ and $M$ indicate a time delay and a multiplication process on correlated signals from two neighbouring sensitive cells (red), adapted from \cite{Frye2015(EMDs-basic)}.}
	\label{Fig: Fly-EMDs}
\end{figure}

\subsubsection{Fly motion detectors}

\begin{figure}[H]
	\centering
	\subfloat[PD detector]{\includegraphics[width=0.37\textwidth]{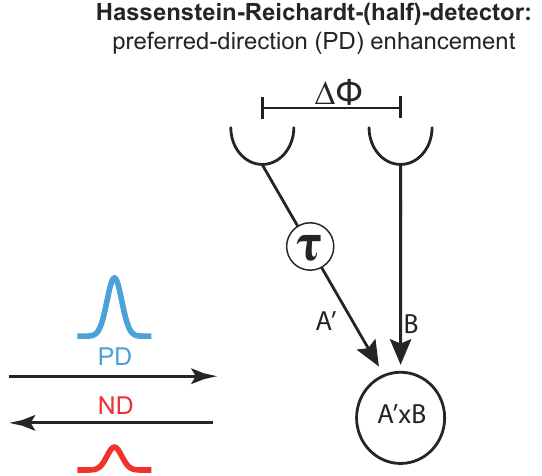}
		\label{fly-detector-PD}}
	\hfil
	\subfloat[ND detector]{\includegraphics[width=0.32\textwidth]{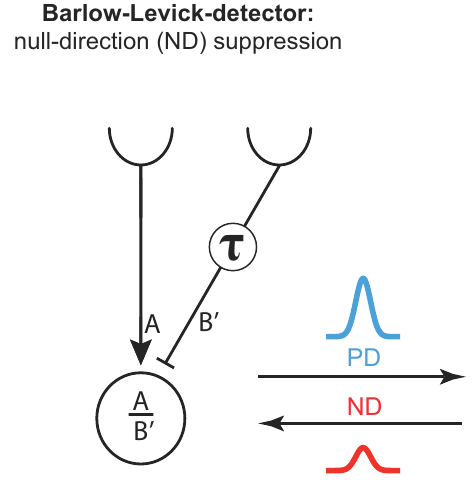}
		\label{fly-detector-ND}}
	\vfil
	\vspace{-0.1in}
	\subfloat[full-HR detector]{\includegraphics[width=0.3\textwidth]{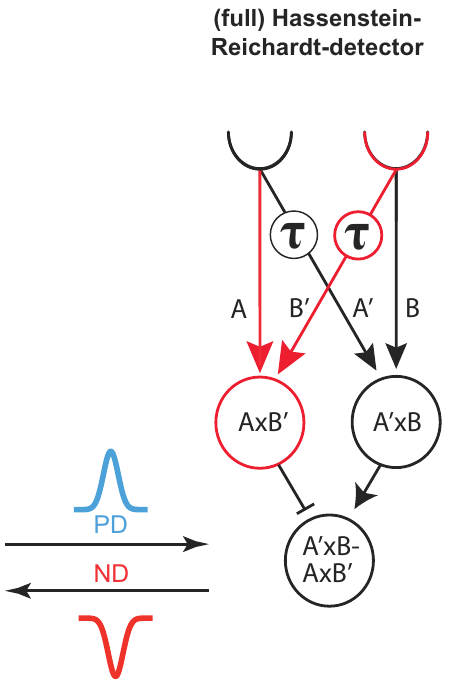}
		\label{fly-detector-full-HR}}
	\hfil
	\subfloat[HR-BL detector]{\includegraphics[width=0.35\textwidth]{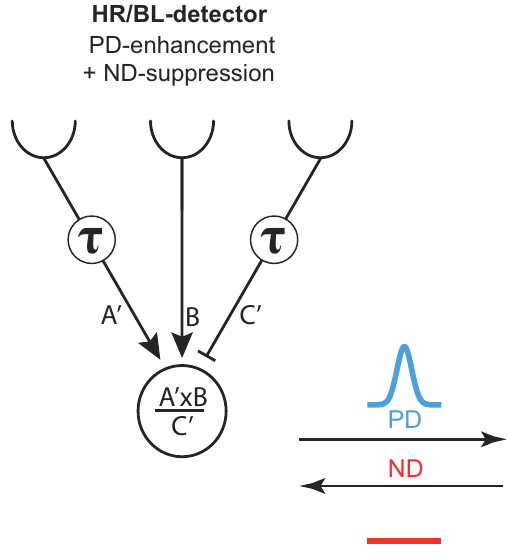}
		\label{fly-detector-HR-BL}}
	\caption{Schematics of a variety of classic fly motion detectors adapted from \cite{Fly-DS-2017(preferred-null)}.}
	\label{Fig: Fly-detectors}
\end{figure}

For defining the computational roles of fly motion detection, a few theories have been proposed in the past half-century, as summarised in \cite{Fly-DS-2017(preferred-null)} and illustrated in the Fig. \ref{Fig: Fly-detectors}. 
A classic and elegant mathematical model was proposed by Hassenstein and Reichardt to account for biological motion perception \cite{HR-1956(EMD-1956)}. 
It was named as the `HR detector' or `HRC' (Hassenstein-Reichardt correlator); this has become commonly referred to as Reichardt detectors or simply the EMD \cite{Reichardt-1987(evaluation-optical-motion),EMD-1989(principles-review),EMD-1999(adaptation-temporal-fly),Frye2015(EMDs-basic)}. 
As illustrated in the Fig. \ref{Fig: Fly-EMDs}, it has been used to explain motion perception strategy by the activity of neighbouring photoreceptors in the field of view. 
From these works, we can summarise that the EMD models have the following characteristics for translational motion detection:
\begin{enumerate}
	\item The direction of motion can be recognised by comparing the activity of at least two correlated photoreceptors in space.
	\item The EMD can not tell the true velocity of a translating pattern; it is sensitive to the spatiotemporal frequencies of stripes that pass over the stationary detectors.
	\item It is also affected by the contrast between a moving pattern and its background, that is, the model responds more strongly to higher contrast translating objects under an identical speed.
	\item There are two paramount parameters in the EMD -- the spacing between a pairwise detectors and the temporal delays for both detectors which can significantly influence the detection of motion direction and intensity.
\end{enumerate}

Recently, two studies on fly motion detectors \cite{Fly-DS-2017(preferred-null),Fly2016(direction-selectivity-fly)} brought together previous famous algorithmic models in the literature, like the HR half-detector (Fig. \ref{fly-detector-PD}) to enhance motion in the PDs, the Barlow-Levick (BL) detector (Fig. \ref{fly-detector-ND}) to suppress motion in the NDs, and the full HR detectors (Fig. \ref{fly-detector-full-HR}) that map PD and ND motion to positive and negative outputs. 
More importantly, both the papers proposed that the HR and the BL mechanisms may work in different sub-regions of the fly receptive field \cite{Fly-DS-2017(preferred-null),Fly2016(direction-selectivity-fly)}. 
It also appears that visual motion detection in flies could rely upon the processing of three input channels that integrates an HR detector with a BL detector \cite{Fly-DS-2017(preferred-null)}, as illustrated in the Fig. \ref{fly-detector-HR-BL}.

\begin{figure}[H]
	\centering
	\subfloat[]{\frame{\includegraphics[width=0.46\textwidth]{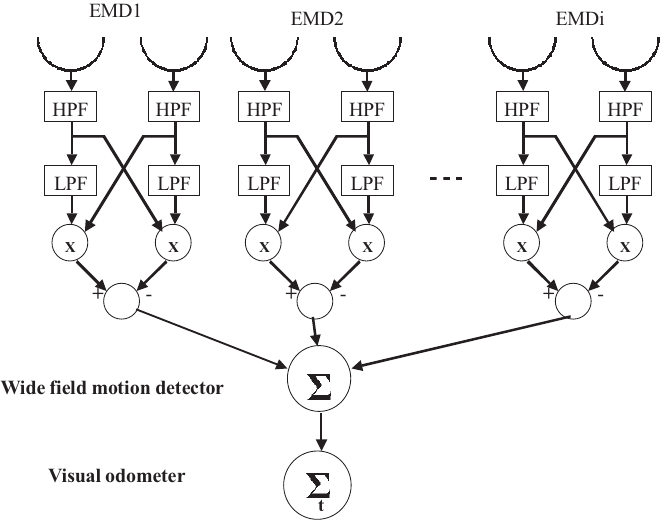}}
		\label{fly-emd-odometer}}
	\hfil
	\subfloat[]{\frame{\includegraphics[width=0.41\textwidth]{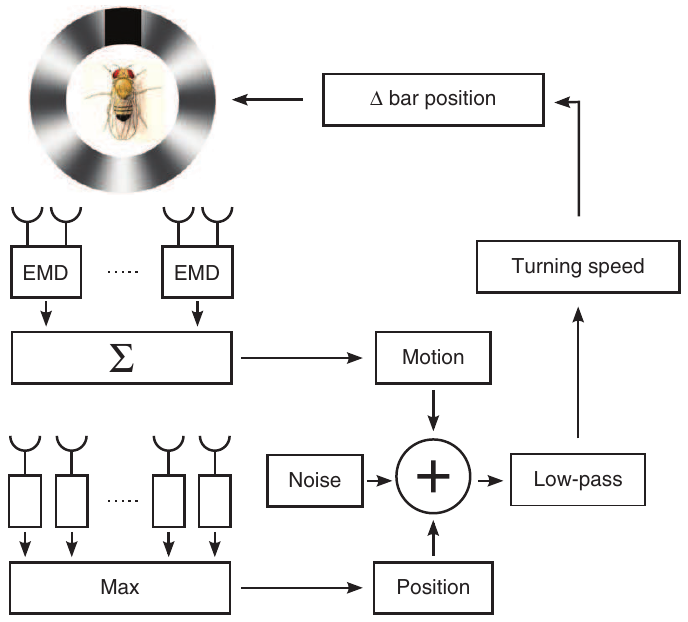}}
		\label{fly-emd-fixation}}
	\vfil
	\vspace{-0.1in}
	\subfloat[]{\frame{\includegraphics[width=0.49\textwidth]{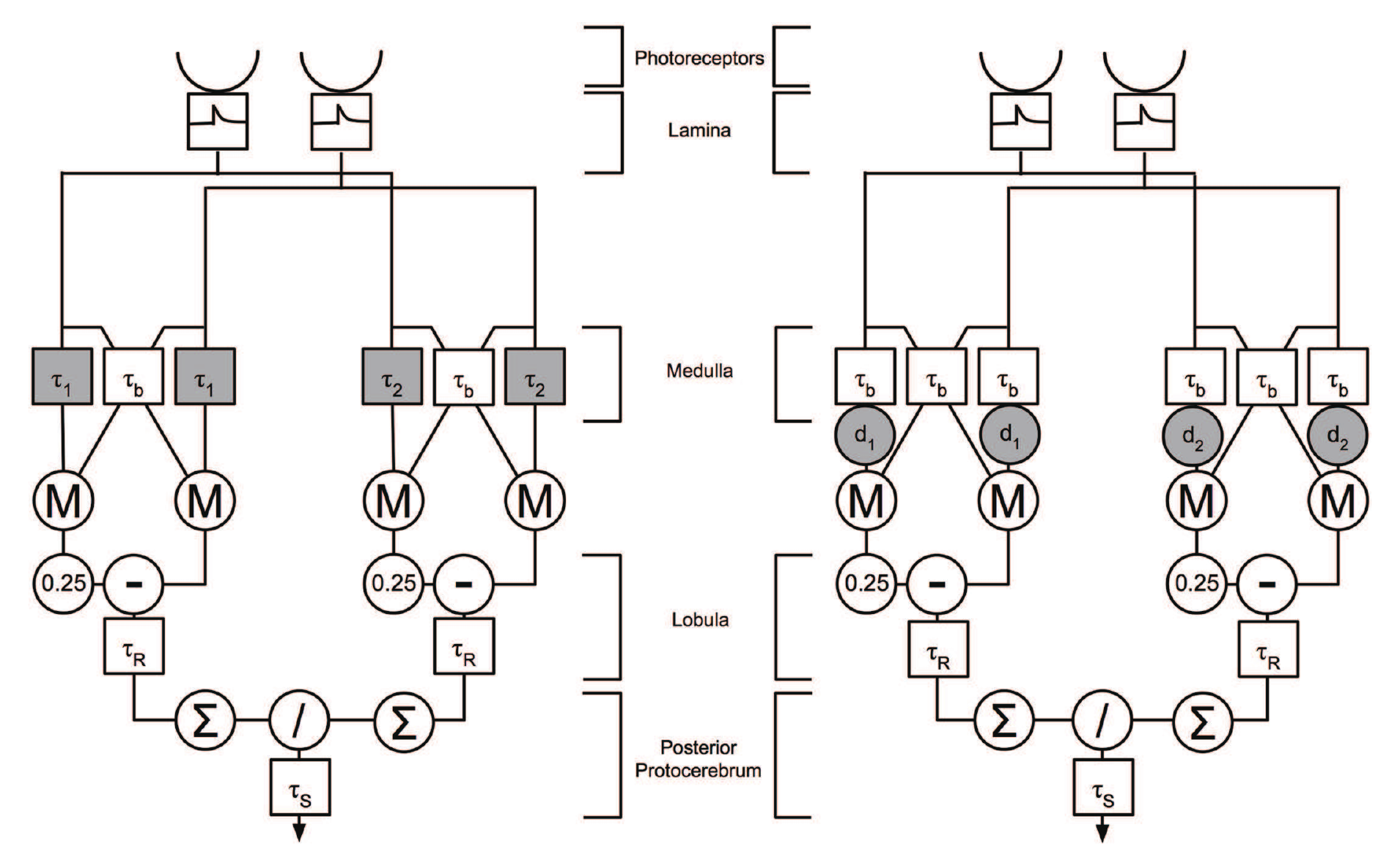}}
		\label{fly-emd-angular}}
	\hfil
	\subfloat[]{\frame{\includegraphics[width=0.46\textwidth]{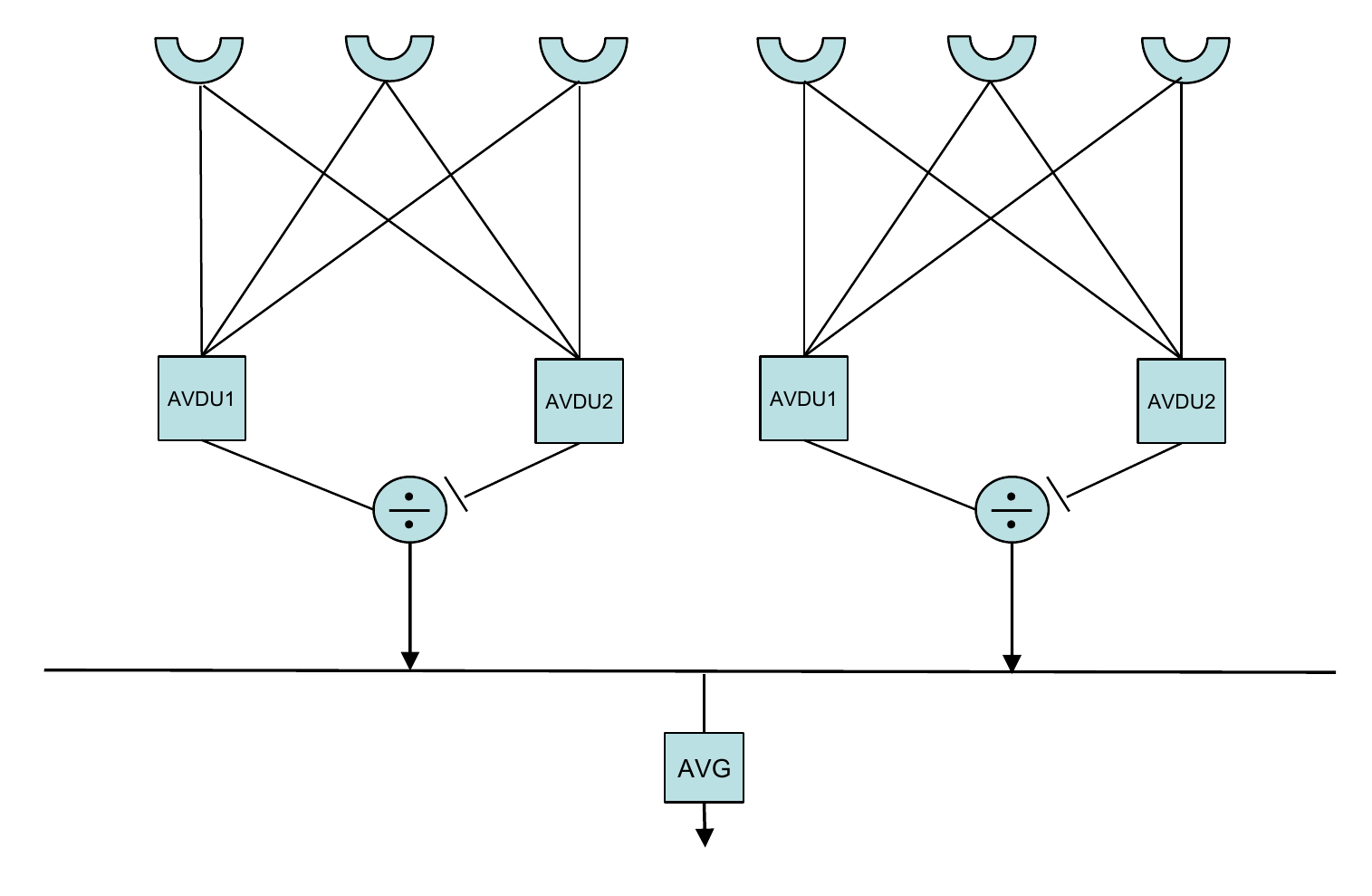}}
		\label{fly-avdu}}
	\caption{A variety of EMD models: (a) This model integrates multiple pairwise EMDs in a two-dimensional form to compute the visual odometer, adapted from \cite{Iida_2000(fly-visual-odometer)}. (b) This model is used to simulate fly fixation behaviour by combining the EMD with an individual location sensitive pathway, adapted from \cite{Fly-2013(motion-blind-tracking)}. (c)-(d) These are models for estimating angular velocity in the bee's brain, adapted from \cite{Cope2016(model-angular-bee)} and \cite{Huatian-ICANN-angular}, respectively.}
	\label{Fig: Fly-emd-models}
\end{figure}

\subsubsection{EMD models and OF-based applications to robotics}

\begin{figure}[H]
	\centering
	\subfloat[]{\frame{\includegraphics[width=0.5\textwidth]{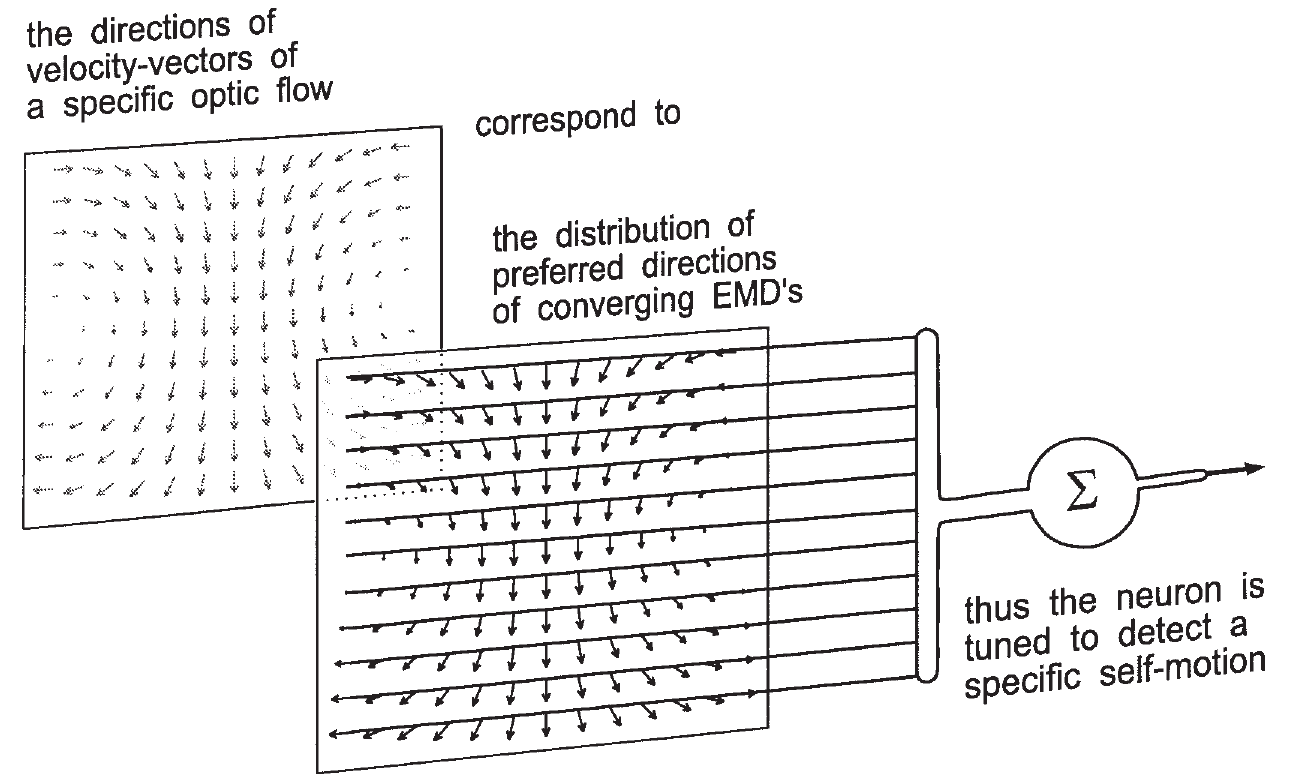}}
		\label{Fly-OF}}
	\hfil
	\subfloat[]{\includegraphics[width=0.45\textwidth]{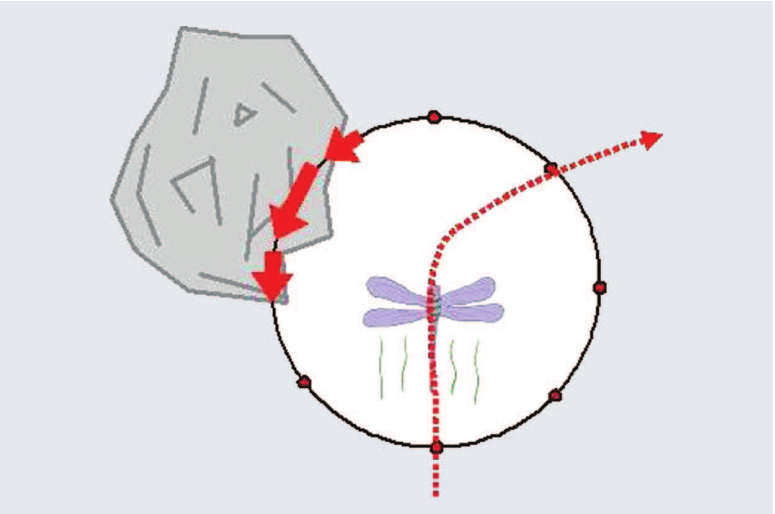}
		\label{Fly-saccade}}
	\caption{Illustrations of fly OF strategy based on the EMDs: (a) a classic method for the integration of local optic flow vectors for the estimation of self-motion, adapted from \cite{Fly-1998(dendritic-optic-flow)}, (b) a collision avoidance strategy based on the OF by a dragonfly, adapted from \cite{OpticFlow-2008(optic-flow-MAV)}.}
	\label{Fig: Fly-OF}
\end{figure}

There are a huge number of computational models and applications that arise from the EMD theories. 
A main utility of the EMDs is to mimic fly and bee optic flow (OF) strategy within the field of view, e.g. \cite{Opticflow-1996(self-motion-interneurons),Opticflow-2006(naturalistic-blowfly-motion),Opticflow-2009(brain-realistic-scenes),Boeddeker2005(blowfly-neurons-opticflow),Blowfly2001(outdoor-motion-neuron),Fly-1998(dendritic-optic-flow)}. 
As illustrated in the Fig. \ref{Fly-OF}, the OF can be defined as a flow vector-field perceived by the visual modality of either animals or machines \cite{Fly-1998(dendritic-optic-flow),Serres2017(review-optic-flow)}; this field is generated by the relative and apparent motion between observer and scene. 
The OF includes two sub-fields of the translational flow and the rotational flow: both are rigorously dependent of the structure of environment including textures and brightness and etc. 
Most Importantly, such a visual strategy can be used to conduct various forms of insect behaviours such as landing, e.g. \cite{landing-2002,Landing-2013(visually-guided-strategy),Ruffier-OF-MAV-unsteady-environments} and terrain following, e.g. \cite{Serres2017(review-optic-flow),Ruffier-OF-MAV-unsteady-environments,Netter-robotic-aircraft-terrain-neuromorphic,Expert-OF-uneven-terrain-following} and tunnel crossing or travelling, e.g. \cite{Baird2010(viewing-angle-bumblebees),Rahar-event-based-autopilot-OF-control} and corridor centring response, e.g. \cite{Robots-2004(mobile-panoramic-sensors),Cope2016(model-angular-bee),Serres-bee-centering-wall-following} and collision or obstacle avoidance, e.g. \cite{Fly-2015(model-collision-reichardt),OpticFlow-2015(colision-goal-direction),Martin-compound-eye-obstacle-avoidance-mobile-vehicle,Mura-scanning-retina-ground-robot,Portelli-honeybees-height-restore} and target tracking, e.g. \cite{Fly1995(network-tracking-fly),Ruffier-two-aerial-micro-robots} and fixation behaviours, e.g. \cite{Fly-2013(motion-blind-tracking)}.

A well-known type of velocity-tuned EMDs was proposed by Franceschini et al. in the 1992 \cite{Nicolas-insects-robot-vision}. 
Being different from the classic HRC, the output of the velocity-tuned EMDs demands on the ratio between the photoreceptors angle in space, as well as the time delay for each pairwise contrast detection photoreceptors. 
This model had been initially tested on a mobile robot \cite{Nicolas-insects-robot-vision}. 
Subsequently, this model has also been named `time of travel' \cite{Moeckel-Liu-time-to-travel-model} or `facilitate and sample' scheme which was implemented in pulse-based analogue VLSI velocity sensors \cite{Kramer-CMOS-VLSI-EMD}.

Furthermore, Iida proposed a method to integrate each pairwise local EMDs in a spatiotemporal manner to compute the visual odometer over time (Fig. \ref{fly-emd-odometer}): this approach was validated by navigation of a flying robot \cite{Iida_2000(fly-visual-odometer),Iida2003(visual-odometer-robot)}. 
Snippe and Koenderink demonstrated possible solutions to extract optical velocity by the design of ensembles of HR detectors \cite{Snippe_1994(optical-velocity-EMD)}. 
Zanker et al. investigated the speed tuning and estimation of EMDs \cite{Zanker-1999(EMD-speed-tuning),Zanker1999(noise-motion-speed)}. 
Subsequently, they analysed video sequences of outdoor scenes from a panoramic camera and performed optic flows as motion signal maps via two dimensional EMDs \cite{Zanker_2005(motion-signal-outdoor)}. 
Rajesh et al. modified the traditional HR detector to improve the velocity sensitivity and reduce the dependence on contrast and image structure: this work matched the neurobiological findings that an adaptive feedback mechanism is effective to normalise contrast of input signals in order to improve the reliability of velocity estimation \cite{Rajesh2005(velocity-estimators-insect)}. 
In addition, Bahl et al. incorporated in the EMDs a parallel position sensitive pathway to track a translating stripe in a simple background and mimic a \textit{Drosophila} behavioural response to fixation, as shown in the Fig. \ref{fly-emd-fixation}: this work was reconciled with electro-physiological recordings from motion-blind flies very well \cite{Fly-2013(motion-blind-tracking)}.

As a variation of the EMD, a few angular velocity estimation models were proposed to account for corridor-centring behaviours of bees, e.g. \cite{Cope2016(model-angular-bee),Huatian-ICANN-angular}, as illustrated in the Fig. \ref{fly-emd-angular} and \ref{fly-avdu}. 
Similarly to \cite{Iida_2000(fly-visual-odometer)}, the integrated response can be used as a visual odometer over time. 
Moreover, there are also studies on the temporal adaptation of EMDs \cite{AdaptiveFly-1996(Reichardt-detector),EMD-1999(adaptation-temporal-fly)}, the contrast sensitivity of EMDs \cite{EMD2005(contrast-saturation-EMD)}, and an EMDs-based algorithm for global motion estimation \cite{EMD-2002(fast-global-motion)}, as well as a non-directional HR detectors model for simulating insect speed-dependent behaviour \cite{Higgins2004(nondirectional-motion-insect)}, and so on.

\begin{figure}[H]
	\centering
	\subfloat[fixed-wing hovering MAV]{\includegraphics[width=0.36\textwidth]{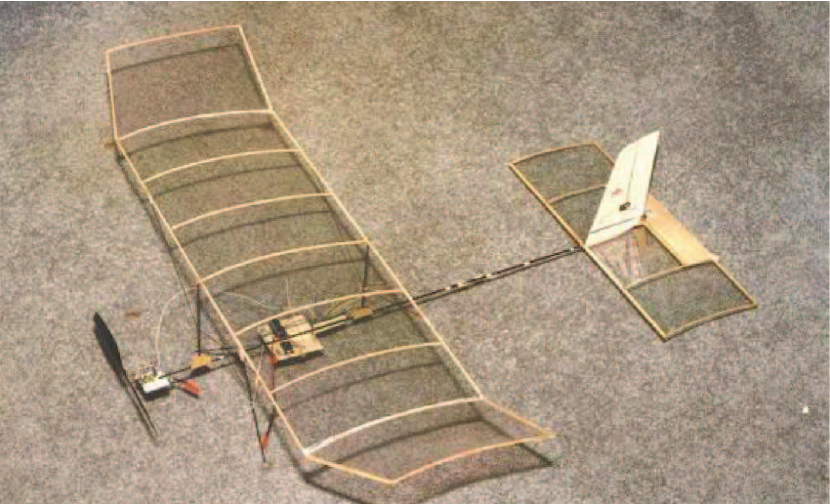}
		\label{Fly-OF-MAV}}
	\hfil
	\subfloat[autonomous sighted hovercraft]{\includegraphics[width=0.35\textwidth]{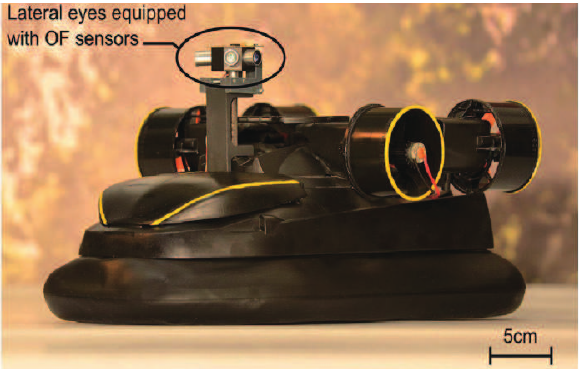}
		\label{Fly-OF-hovercraft}}
	\hfil
	\subfloat[`Robot Fly']{\includegraphics[width=0.14\textwidth]{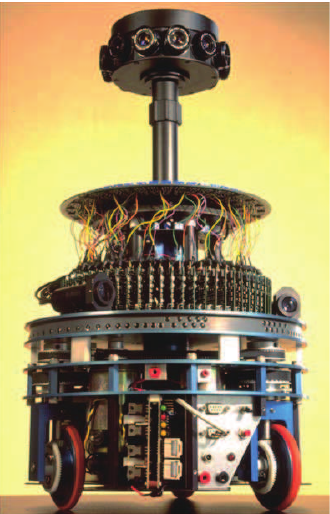}
		\label{Fly-EMD-robotfly}}
	\vfil
	\vspace{-0.1in}
	\subfloat[100-g micro-helicopter]{\includegraphics[width=0.31\textwidth]{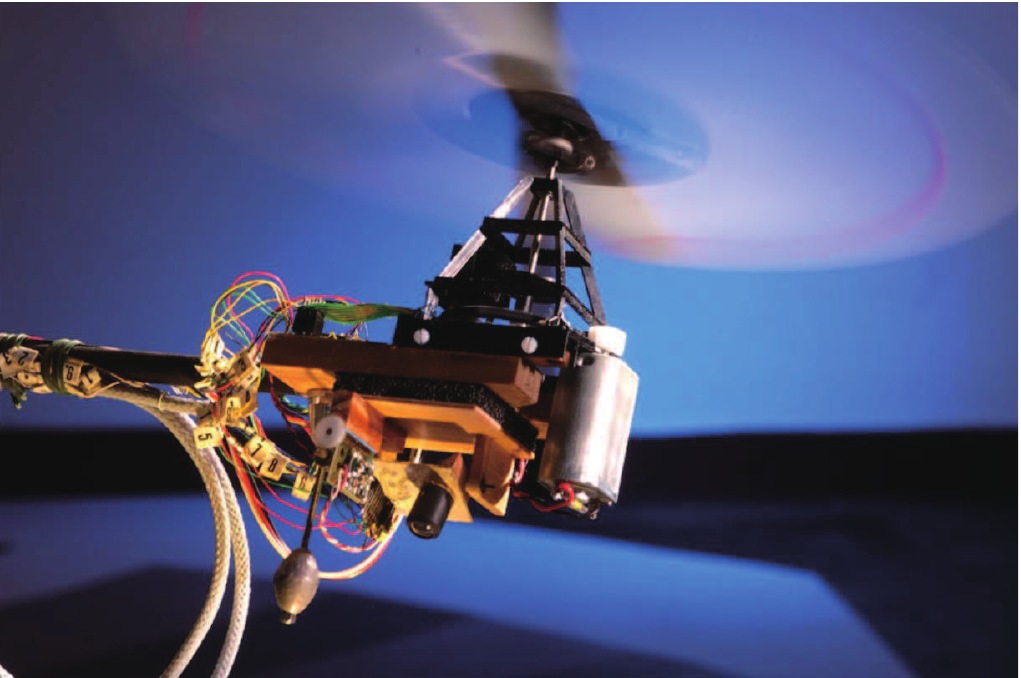}
		\label{Fly-EMD-MH}}
	\hfil
	\subfloat[flying robot `OSCAR I']{\includegraphics[width=0.305\textwidth]{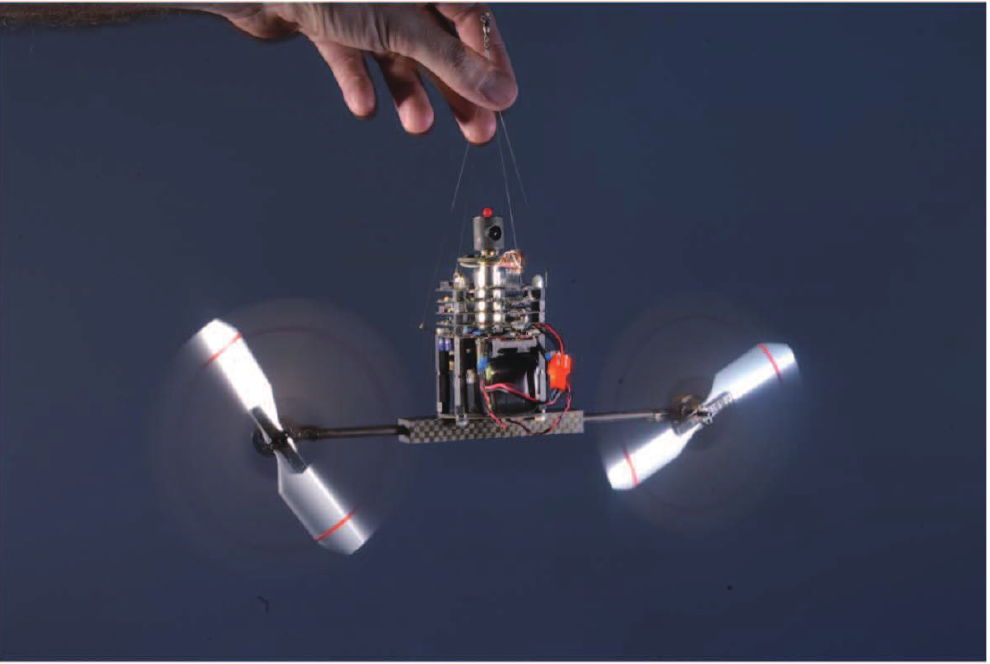}
		\label{Fly-EMD-OSCAR-I}}
	\hfil
	\subfloat[`OSCAR II']{\includegraphics[width=0.32\textwidth]{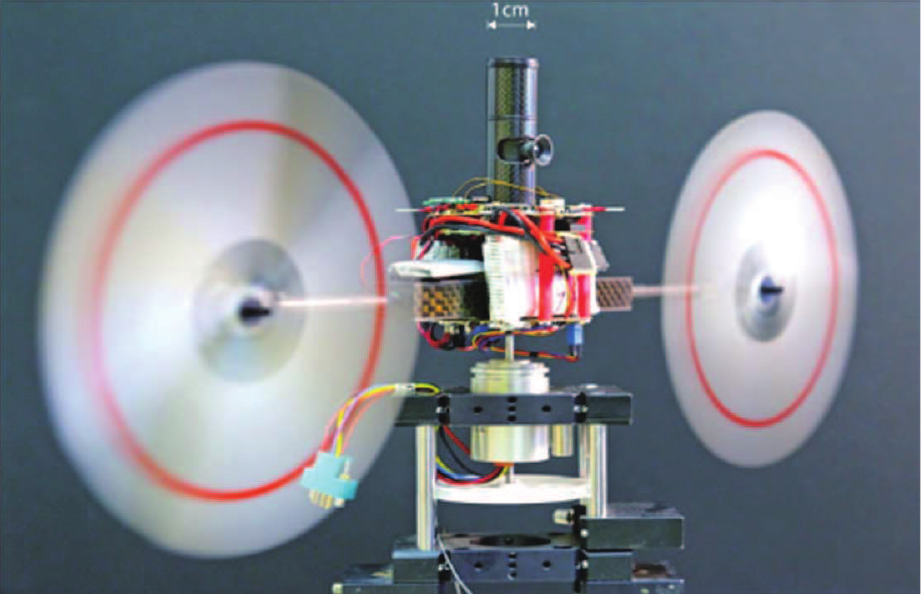}
		\label{Fly-EMD-OSCAR-II}}
	\vfil
	\vspace{-0.1in}
	\subfloat[silicon retina]{\includegraphics[width=0.5\textwidth]{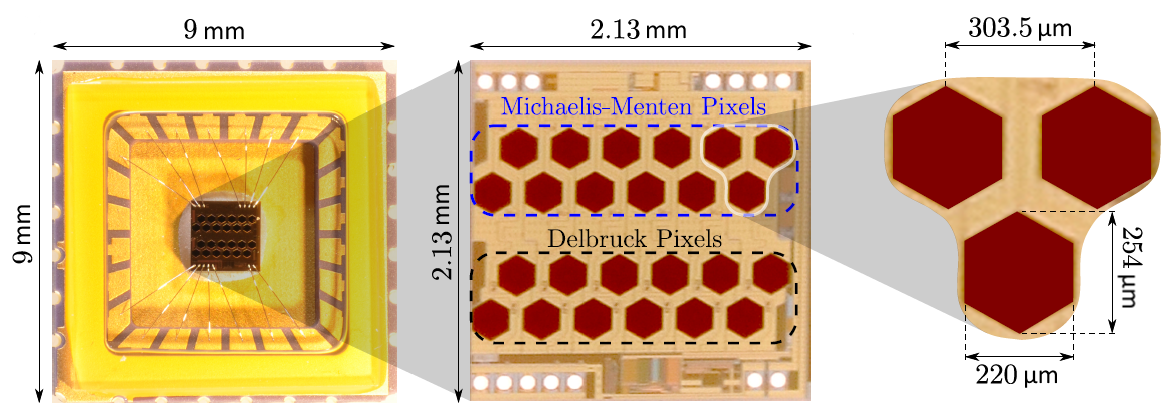}
		\label{Fly-silicon-retina}}
	\hfil
	\subfloat[`CurvACE' sensor (compound eye)]{\includegraphics[width=0.4\textwidth]{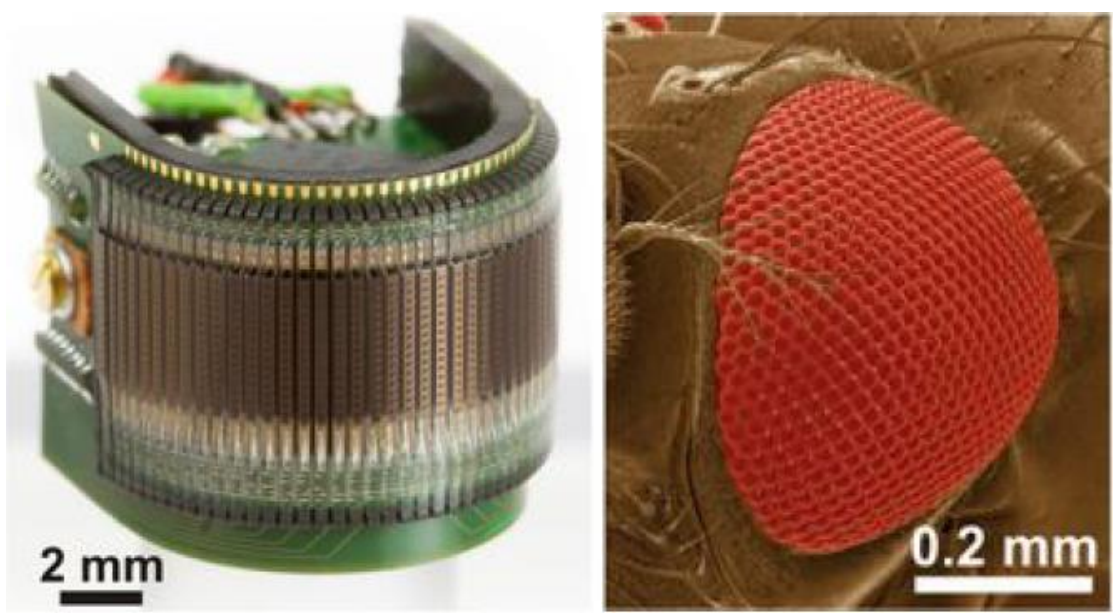}
		\label{Fly-CurvACE}}
	\caption{A few instances of bio-robotic applications of the fly EMDs and OF based control strategy: (a) is adapted from \cite{OpticFlow-2008(optic-flow-MAV)}, (b) is adapted from \cite{hovercraft2014(vision-bee-corridor)}, (c)-(f) are adapted from \cite{Nicolas-2014(Review-Fly-Robot)}, (g) is adapted from \cite{Vanhoutte-time-of-travel-OF-micro-flying-robot}, (h) is adapted from \cite{Nicolas-group-CurvACE}.}
	\label{Fig: Fly-biorobotic-apps}
\end{figure}

The aforementioned considerable amount of computational studies have demonstrated the effectiveness and computational simplicity of the fly EMDs models and OF-based control strategies. 
As a result, these models have been leading the design of small and efficient artificial translation perception sensors, such as silicon retina mimicking the structure of fly compound eye as shown in the Fig. \ref{Fly-silicon-retina} which was designed in very tiny size and used for on-board autopilots \cite{Vanhoutte-time-of-travel-OF-micro-flying-robot}, as well as implementation of CurvACE -- miniature curved artificial compound eyes in the Fig. \ref{Fly-CurvACE} which has the optical spatial filtering done by the bell-shape angular sensitivity created on each photoreceptor \cite{Nicolas-group-CurvACE}, as well as sensor implementation using thresholding-based digital processing on the MAV \cite{Ruffier-MAV-OF-circuits,Roubieu-1-gram-sensor-EMD}. 
In addition, as illustrated in the Fig. \ref{Fig: Fly-biorobotic-apps}, these models have been very widely used in bio-robotic applications. 
These bio-inspired robots can show similar visually-guided behaviours like insects. 
More concretely, there are many ground and flying robots benefiting from the OF-based sensors which are used for guiding the robots for autopilots, e.g. \cite{Sabiron-lightweight-sensor-OF-helicopter-outdoor,Ruffier-OF-regulation-autopilot,Rahar-event-based-autopilot-OF-control,Nicolas-bio-flying-robot-insect-piloting,Nicolas-insect-inspired-autopilots} , and collision avoidance (Fig. \ref{Fly-OF-MAV}, \ref{Fly-EMD-robotfly}), e.g. \cite{OpticFlow-2008(optic-flow-MAV),Nicolas-2014(Review-Fly-Robot),Serres2017(review-optic-flow),Pichon-mobile-robot-self-motion-sensor}, and tunnel crossing (Fig. \ref{Fly-OF-MAV}, \ref{Fly-EMD-robotfly}), e.g. \cite{hovercraft2012(autonomous-bees-control),hovercraft2014(vision-bee-corridor),Serres2017(review-optic-flow)}, and terrain following, e.g. \cite{Ruffier-two-aerial-micro-robots,Expert-OF-uneven-terrain-following}, take-off and landing behaviours (Fig. \ref{Fly-EMD-MH}, \ref{Fly-EMD-OSCAR-I}, \ref{Fly-EMD-OSCAR-II}), e.g. \cite{Nicolas-2014(Review-Fly-Robot)}, as well as indoor and outdoor visual odometry on a car-like robot \cite{Mafrica-OF-minisensor-robot}. 
There is also comparison on aerial robot between auto-adaptive retina based implementations using thresholding-based digital processing and cross-correlation digital processing \cite{Vanhoutte-time-of-travel-OF-micro-flying-robot}. 
Regarding this field, two prominent review papers \cite{Nicolas-2014(Review-Fly-Robot),Serres2017(review-optic-flow)} introduce, more systematically, the relevant bio-robotic approaches and applications of the fly EMDs and OF.

\subsubsection{Further discussion}

A shortcoming or unsolved problem of the fly EMD-based models for translation perception is the velocity tuning of motion detection. 
In other words, a biological motion-detecting circuit may not tell the true velocity of stimuli \cite{Frye2015(EMDs-basic)}. 
The reason is that for each combination of such `delay-and-correlate' motion detectors, it is advisable to decide the spacing between each pairwise detectors, and the time span for the delay in follow-up non-linear computation like the multiplication, each factor of which will affect the model's performance for sensing translation \cite{Zanker-1999(EMD-speed-tuning)}. 
For example, perceiving faster movements requires a larger spatial span between detectors if fixing the temporal delay; instead, it requires a shorter time delay when the spacing is unchanged.

Another defect is that the state-of-art models or strategies for motion perception still lack robust mechanisms for filtering out irrelevant motion from dynamic visual clutter of great complexity in the real world, so that they are always influenced by environmental noise such as windblown vegetation and shifting of background or surroundings caused by ego-motions. 
From a computational modeller's perspective, it is still a big challenge to make the motion perception visual systems robust to filter out irrelevant from relevant motion dealing with a complex and dynamic scene like the driving scenarios.

\subsection{Modelling of fly ON and OFF pathways and LPTCs}
\label{section: literature: translating: fly-lptc}

Within this subsection, we will continue to present the cutting-edge biological research in fly visual systems. 
We will focus on the understanding of underlying circuits or pathways in the fly visual brains for preliminary motion detection. 
We have now understood from the Section \ref{section: literature: translating: fly-emd} that visual neurons compute the direction of motion conforming to the HR or related theories: both flies and bees apply the EMDs to sense optical flows, which is very effective to deal with a variety of insect visually-guided behaviours. 
We have also known that the optical flow is sensed by the fly compound eyes after spatial filtering of motion information. 
However, a few questions still exist: where does the specific DS form to perceive translation stimuli within the internal structure of fly visual pathways? 
Which neurons carry out the corresponding neural computation? 
And how does the neural response connect to visuomotor control?

\subsubsection{Biological exploration}

\begin{figure}[H]
	\centering
	\subfloat[]{\includegraphics[width=0.25\textwidth]{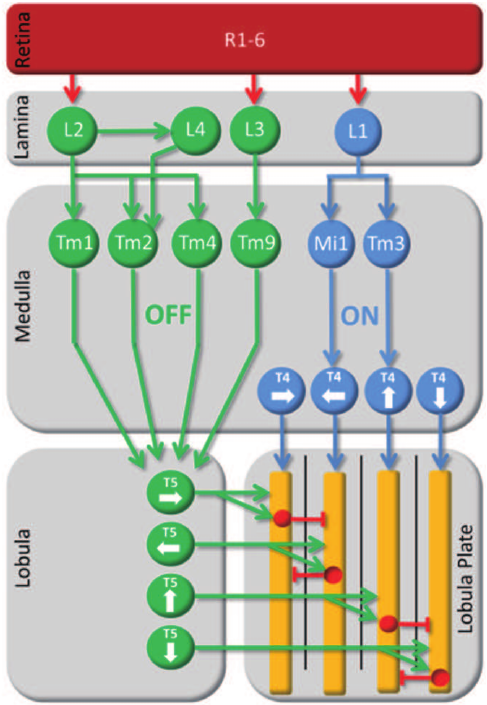}
		\label{Fly-visual-pathways}}
	\hfil
	\subfloat[]{\includegraphics[width=0.34\textwidth]{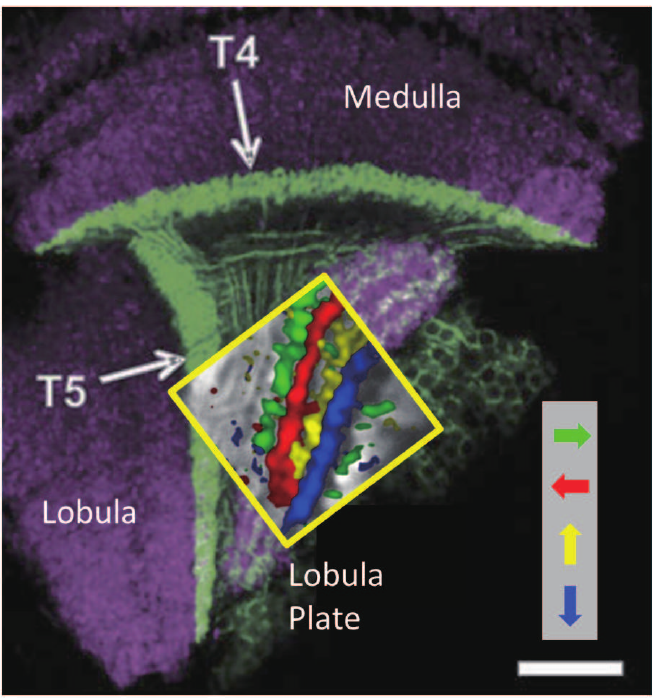}
		\label{Fly-neuromorphology}}
	\hfil
	\subfloat[]{\includegraphics[width=0.38\textwidth]{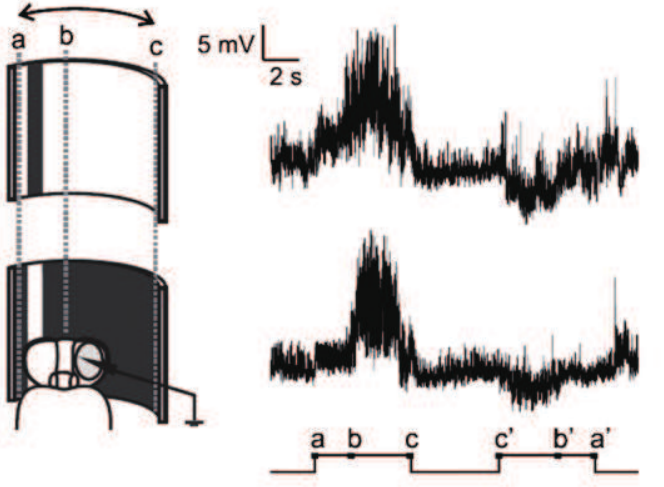}
		\label{Fly-LPTCs-response}}
	\caption{Schematic diagrams of fly preliminary motion-detecting neuropile layers: (a) The underlying ON and OFF pathways with interneurons and LPTCs perceive visual motion stimuli and generate the DS to four cardinal directional motion. (b) The neuromorphology of fly visual circuits -- the LPTCs pool directionally specific motion signal, individually into each sub-layers. (c) The LPTCs respond to directional translating stimuli by movements of dark and light bars, respectively. (a), (b) are adapted from \cite{Fly2016(direction-selectivity-fly)} and (c) is adapted from \cite{Borst2011(review-motion)}.}
	\label{Fig: Fly-Pathways-LPTCs}
\end{figure}

To address these questions, biologists have demonstrated that fly ON and OFF parallel visual pathways and LPTCs are the proper places to implement directionally selective signal processing, as illustrated in the Fig. \ref{Fly-visual-pathways} \cite{Borst-2010(review-fly-vision),Borst-2014(review-fly),Borst2011(review-motion),Borst2015(common-circuit-motion)}. 
A seminal work by Franceschini et al., early in the 1989, had proposed a transient ON-OFF nature of EMD response in the housefly: in this research, a splitting of an EMD into an ON-EMD and an OFF-EMD was presented to sense light and dark edge translation, separately \cite{Nicolas-1989(DSN-Insect-Neurons)}. 
Though many parts of the underlying neural mechanisms in the fly visual systems still remain unknown until today, many efforts have been made by biologists for exploring internal structures including cellular functionality underlying directional motion perception, particularly in the recent two decades \cite{Rister-2007(dissection-motion-channel),Fly-2014(processing-properties-motion),Fly-DS-2016(preferred-null),Fly-DS-2017(emergence-direction-selectivity),Fly-DS-2017(preferred-null),Fly2016(direction-selectivity-fly),Fly-Motion-2017(cellular-hybrid-detector),Fisher-L3(wide-field-local),Figure_2012(tracking-fly-parallel),Strother2014(direct-observation-on-off),Vogt-2007(first-steps-fly),Shinomiya2014(candidate-off-fly),CommonEvo-2015(ON-OFF-edge),Fly-2013(motion-circuit-connectomics)}. 
More concretely, a nice progress was made by Joesch et al., in the 2010. 
They looked into the downstream processing of photoreceptors and found that the visual signals are split into two parallel polarity pathways by L1 (ON) and L2 (OFF) inter-neurons in the neuropile layer of Lamina \cite{Joesch-2010(ON-OFF-fly)}, as illustrated in the Fig. \ref{Fly-visual-pathways}. 
After that, Maisak et al. revealed the characteristics of T4 and T5 cells in the neuropile layers of Medulla and Lobula \cite{Maisak_2013(T4-T5-fly)} in the Fig. \ref{Fly-neuromorphology}. 
Subsequently, a group of LPTCs in the neuropile layer of Lobula Plate has been also identified as wide-field detectors that integrate upstream visual signals \cite{Maisak_2013(T4-T5-fly),ColumnarCells-2012(fly-wide-field),Circuit-genetic(genetic-push-motion),Fly2013(brain-visual-motion)} (Fig. \ref{Fly-visual-pathways} and \ref{Fly-neuromorphology}). 
Importantly, the LPTCs have been demonstrated to process the optic flow field sensed by photoreceptors on a higher level corresponding to the control of visual flight course \cite{Serres2017(review-optic-flow)}. 
Furthermore, as shown in the Fig. \ref{Fly-LPTCs-response}, the horizontal translation sensitive LPTCs are rigorously activated by motion in the PD yet inhibited by motion in the ND underlying the specific DS of the fly DSNs.

On the basis of these biological findings, we can draw a directional tuning map of fly preliminary visual processing through multi-layers to demonstrate the following steps for translational motion perception, according to the Fig. \ref{Fly-visual-pathways}:
\begin{enumerate}
	\item The motion detection starts with splitting visual signals captured by photoreceptors (R1-6) in the first Retina layer into two parallel ON and OFF pathways; this is done by the large mono-polar cells (LMCs) -- L1 and L2 in the second Lamina layer.
	\item The L1 inter-neurons convey onset response by luminance increments to neurons in the Medulla layer; whilst the L2 inter-neurons relay offset response by luminance decrements to neurons in the Lobula layer.
	\item The EMDs are equipped to the T4 cells in the Medulla layer in the ON pathway, and the T5 cells in the Lobula layer in the OFF pathway in order to generate directionally specific motion signals. 
	It is worth emphasising that the four cardinal directions are formed in different groups of T4 and T5 cells, separately.
	\item Finally, the LPTCs integrate signals from ON and OFF channels in the Lobula Plate; motion information with an unanimous direction congregates at a same sub-layer and jointly flow downstream to circuits like motion-driven neural systems.
	\item There is another pathway, that is, the L3--Tm9--T5 (Fig. \ref{Fly-visual-pathways}) that provide locational information of objects to the OFF pathway and is regardless of the direction of motion \cite{Fisher-L3(wide-field-local)}; this pathway cooperates effectively with the ON and OFF pathways to conduct a fly fixation behaviour \cite{Fly-2013(motion-blind-tracking),Fu-2017(ROBIO-fixation)}.
\end{enumerate}

\subsubsection{ON and OFF motion detectors}

\begin{figure}[H]
	\centering
	\subfloat[three types of fly motion detectors]{\includegraphics[width=0.9\textwidth]{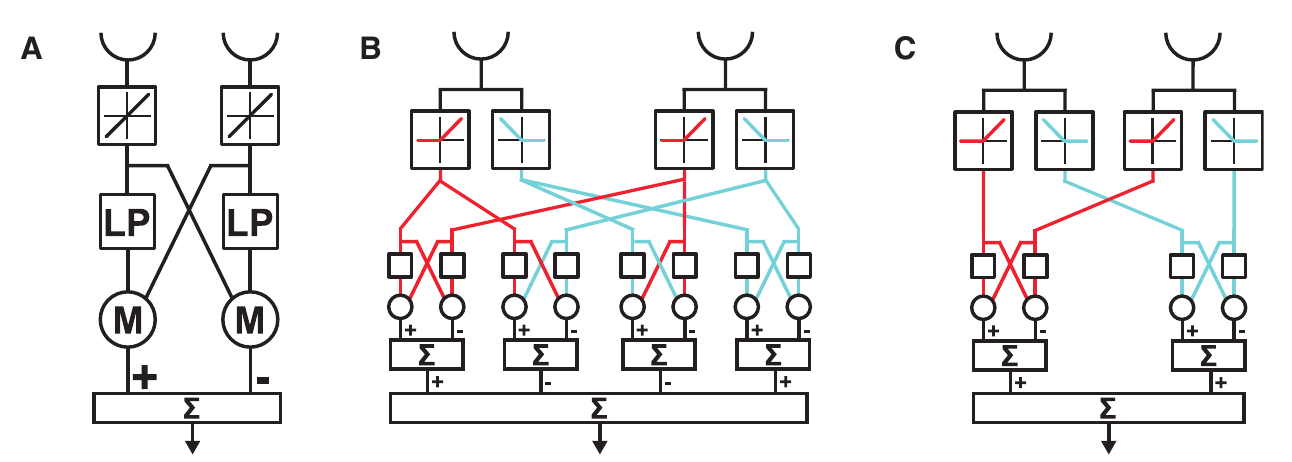}
		\label{fly-polarity-models}}
	\vfil
	\vspace{-0.1in}
	\subfloat[2-Q model]{\includegraphics[width=0.25\textwidth]{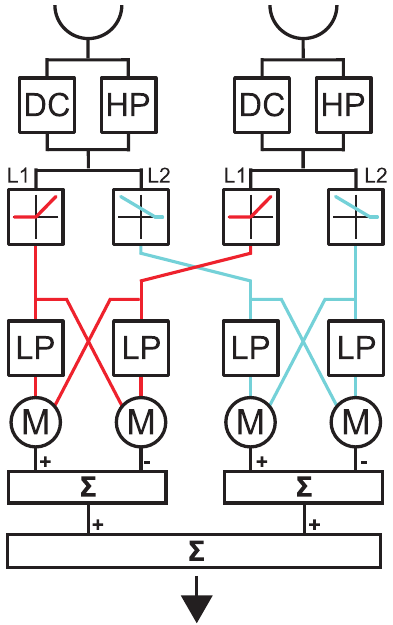}
		\label{fly-polarity-2Q}}
	\hfil
	\subfloat[6-Q model]{\includegraphics[width=0.5\textwidth]{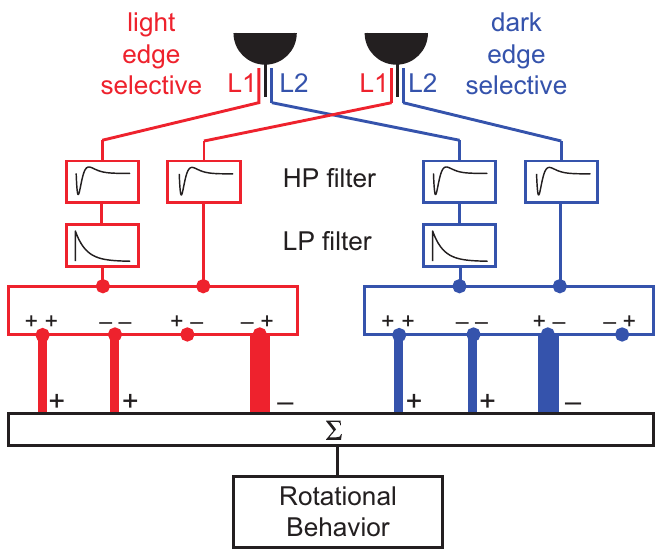}
		\label{fly-polarity-6Q}}
	\caption{Schematics of classic fly motion detectors with different combinations of ON and OFF detectors: (a) three basic types -- a pairwise EMDs (A), a 4-Q model (B) and a 2-Q model (C), adapted from \cite{Eichner2011(2Q-motion)}, (b) the 2-Q detector with input combinations of the same sign polarity detectors, adapted from \cite{Eichner2011(2Q-motion)}, (c) a 6-Q detector processes ON and OFF edge information in both pathways, adapted from \cite{Clark_2011(6Q-model-fly)}.}
	\label{Fig: Fly-polarity-models}
\end{figure}

Since the cellular implementations of EMDs have been demonstrated to be in the process of the ON and OFF pathways, there have been a few fly motion detectors with the ON and OFF mechanisms, as illustrated in the Fig. \ref{Fig: Fly-polarity-models}. 
The fundamental computation conform to the full HR detectors (Fig. \ref{fly-detector-full-HR}). 
More precisely, a 4-Quadrant (4-Q) detector encodes input combinations of ON-ON, OFF-OFF, ON-OFF and OFF-ON cells, which mathematically conform to the traditional EMDs (Fig. \ref{fly-polarity-models}). 
Eichner et al. proposed a 2-Quadrant (2-Q) motion detector, as illustrated in the Fig. \ref{fly-polarity-2Q}; this model processes input combinations of only the same sign signals, i.e., ON-ON in the ON pathway and OFF-OFF in the OFF pathway \cite{Eichner2011(2Q-motion)}. 
In addition, they revealed also a small fraction of original signals can pass through into the motion-detecting circuits which demonstrates that not only the transient luminance change but also the permanent brightness can be encoded by motion sensitive neurons. 
Moreover, Clark et al. presented a 6-Quadrant (6-Q) detector through behavioural experiments on fruit flies, as illustrated in the Fig. \ref{fly-polarity-6Q}; this model was constructed to match the behavioural observations in L1 and L2 blocked flies \cite{Clark_2011(6Q-model-fly)}. 
In this computational theory, either polarity pathways convey positive and negative contrast changes with edge selectivity inside motion detection circuits. 
To make a decision between these alternatives that determine motion detection strategy in flies, a case study suggested the existence of two over six quadrants detectors, by genetically silencing either the ON or the OFF pathway \cite{Joesch_2013(functional-ON-OFF)}.

\subsubsection{Computational models}

Following previous EMD and OF strategies, these different ON and OFF combinations of fly motion detectors have given us further understandings of the complex fly visual processing circuits; however, there is very few modelling studies, systematically, testing these polarity detectors and LPTC in real-world scenes with dynamic visual clutter. 
To fill this gap, Fu and Yue recently proposed a fly DSNs based visual neural network with ensembles of 2-Quadrant detectors to extract translational motion cues from a cluttered real physical background \cite{Fu2017(fly-DSNs-IJCNN)}. 
Subsequently, they extended this model to a behavioural response to fixation mimicking the fly behaviour, by incorporating in a newly-modelled position sensing system of the L3-Tm9-T5 pathway in the Fig. \ref{Fly-visual-pathways} \cite{Fu-2017(ROBIO-fixation)}. 
This model was then successfully implemented on the embedded system in an autonomous micro-robot \cite{Fu-ROBIO-2018}. 
Likewise, Wang et al. estimated the background motion by the LPTC responses to shifting of visually cluttered scenes. 
In this modelling study, a maximisation operation mechanism was proposed to simulate the functionality of the wide-field Tm9 neurons \cite{HWang-ICDL(LPTC-model)}, which effectively improves the performance of perceiving wide-field background translation.

\section{Small target motion perception models}
\label{Sec: Small-Target-Motion}

In previous sections, we have reviewed motion perception models that originate from insect visual systems and possess specific DS to looming and translation visual stimuli and corresponding applications to artificial machines. 
This section continues to review computational models of a specific group of visual neurons, which are sensitive to moving objects with small sizes only. 
These include small target motion detector (STMD) in the Section \ref{section: literature: small: stmd} and figure detection neuron (FDN) in the Section \ref{section: literature: small: fd} that have particular size selectivity to small target motion, with relevant biological findings about the STMD and the FDN in insect visual systems.

\subsection{Background}
\label{section: literature: small: background}

Due to the long sight distance, targets such as mates or preys, always appear as small dim speckles whose size may vary from one pixel to a few pixels in the field of view. 
Being able to perceive such small targets, in a distance and early, would endow insects with more competitiveness for survival. 
However, from a modeller's perspective, small target motion detection against naturally cluttered backgrounds has been considered as a challenging problem which includes the following several aspects: 
(1) small targets cannot provide sufficient visual features, such as shape, colour and texture, for target detection; 
(2) small targets are often buried in cluttered background and difficult to separate from background noise; 
(3) ego motion of the insects may bring further difficulties to small target motion detection. 
These challenges have been solved well by insects after long-term evolutionary development, and their small target motion detection visual systems are efficient and reliable.

\subsection{Small target motion detectors}
\label{section: literature: small: stmd}

\subsubsection{Biological research}

In the insect visual systems, a class of specific small-field motion sensitive neurons, called the small target motion detectors (STMD), is characterised by their exquisite sensitivity for small target motion. 
The STMD neurons have been observed in several insect groups, including hawk moths \cite{collett1971visual}, hover flies \cite{collett1975visual}, and dragonflies \cite{olberg1981object,olberg1986identified,o1993feature}. 
In the past two decades, the anatomy and physiology of STMD neurons have been further investigated with a good number of researches \cite{nordstrom2006small, nordstrom2006insect,barnett2007retinotopic,bolzon2009local,nordstrom2012neural,nordstrom2011spatial,geurten2007neural}.

\begin{figure}[H]
	\centering
	\subfloat[]{\includegraphics[width=0.45\textwidth]{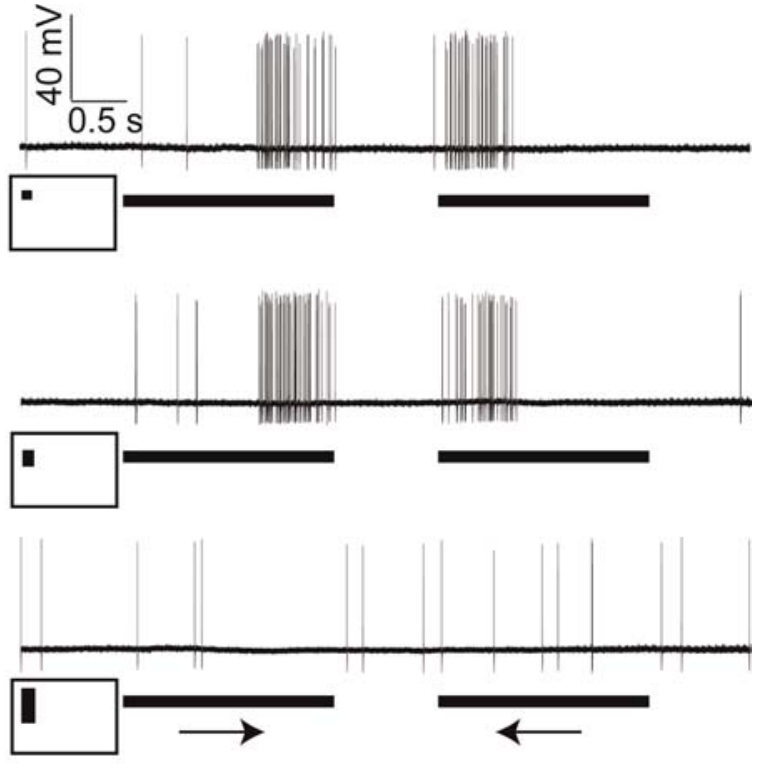}
		\label{STMD-Size-Tuning-2006-PlosBology}}
	\hfill
	\subfloat[]{\includegraphics[width=0.5\textwidth]{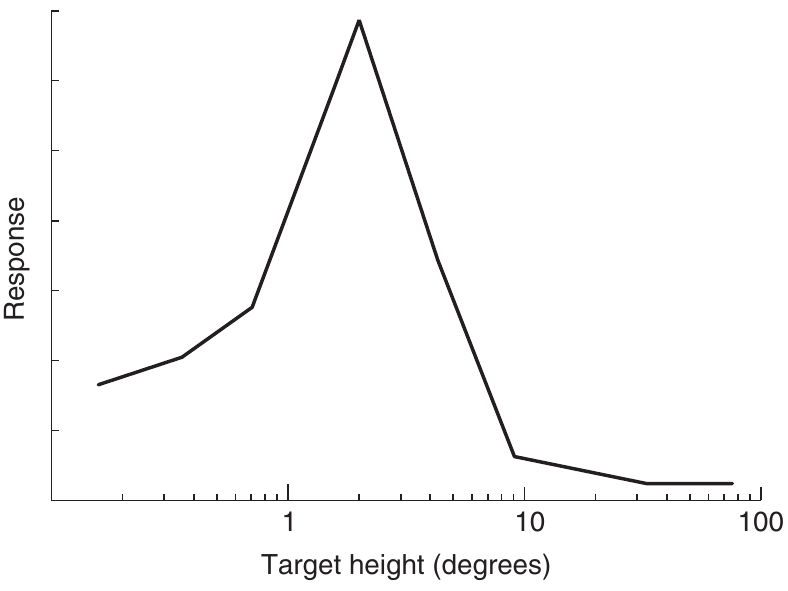}
		\label{STMD-Size-Tuning-2012-Neurobiology}}
	\caption{STMD neuronal raw responses: (a) neuronal responses to motion of three different-sized targets ($0.8^{\circ}$, $3^{\circ}$, or $15^{\circ}$ high by $0.8^{\circ}$ wide) drifted against bright backgrounds: the horizontal bars indicate the movement duration and the arrows denote the direction of target motion, adapted from \cite{nordstrom2006insect}. (b) the response of an STMD to targets of varying height, adapted from \cite{nordstrom2012neural}.}
	\label{Fig: STMD-Size-Tuning}
\end{figure}

\begin{figure}[H]
	\centering
	\subfloat[]{\includegraphics[width=0.45\textwidth]{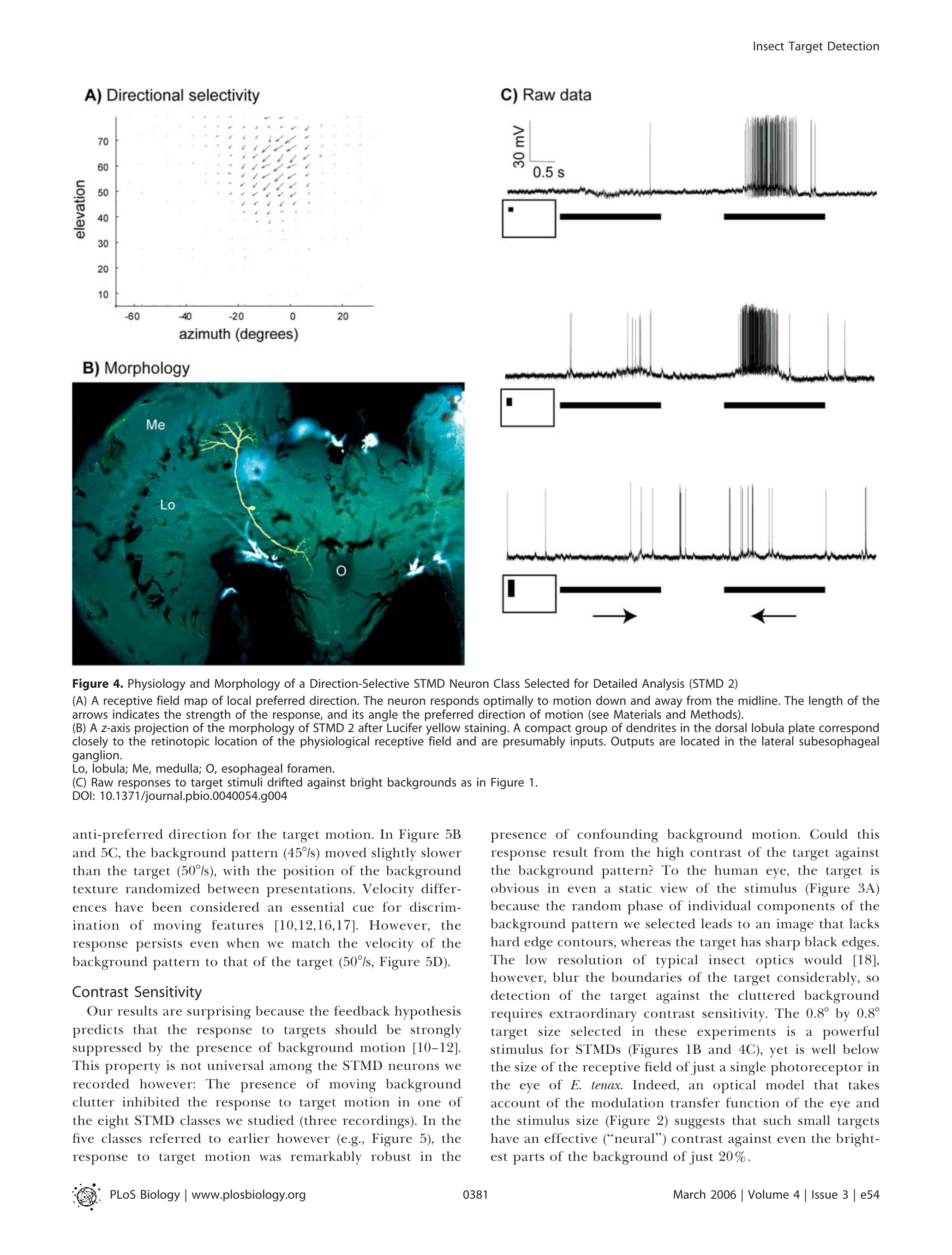}
		\label{STMD-Direction-Tuning-2006-PlosBology}}
	\hfill
	\subfloat[]{\includegraphics[width=0.49\textwidth]{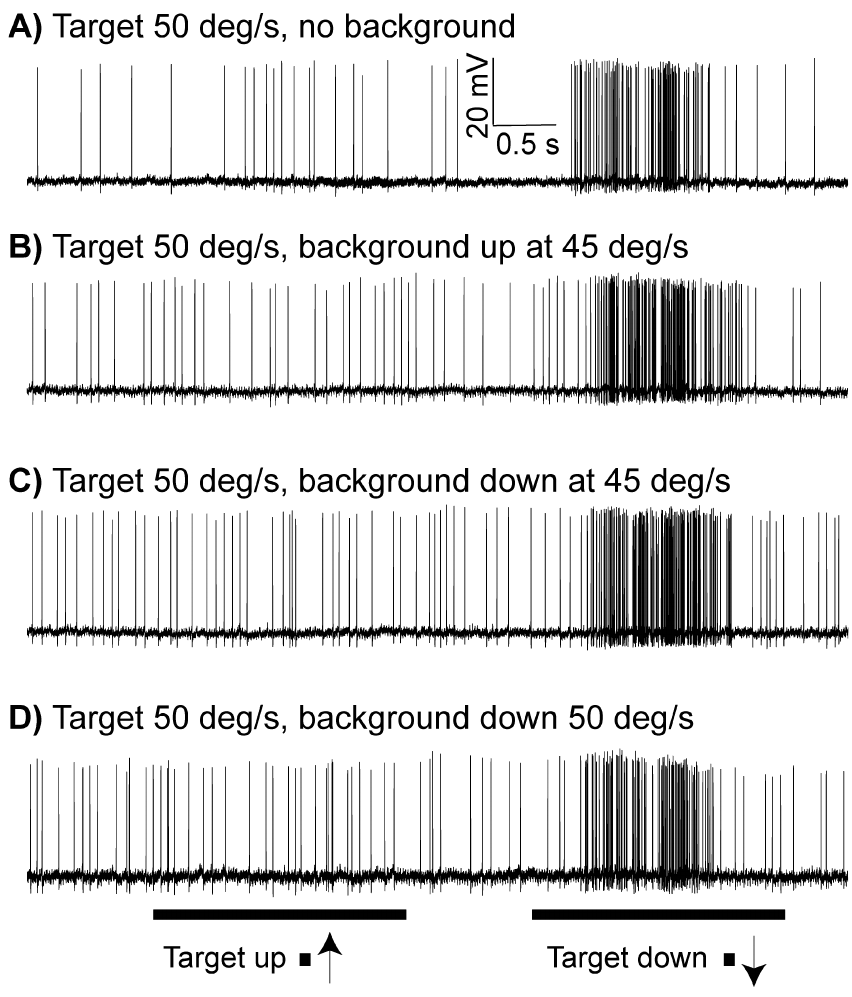}
		\label{STMD-Background-Resonse-2012-Neurobiology}}
	\caption{(a) Raw responses of the directionally selective STMD neuron which prefers target motion to left, tested by motion of three different-sized targets ($0.8^{\circ}$, $3^{\circ}$, or $15^{\circ}$ high by $0.8^{\circ}$ wide) drifted against bright backgrounds: the horizontal bars indicate the stimuli duration and the arrows denote the direction of target motion, adapted from \cite{nordstrom2006insect}. 
		(b) Responses of the STMD neuron which prefers target motion downward, to targets drifted against cluttered backgrounds, adapted from \cite{nordstrom2006insect}.}
	\label{Fig: STMD-Direction-Tuning-Background-Noise-2006-PlosBology}
\end{figure}


\begin{figure}[H]
	\centering
	\includegraphics[width=0.6\textwidth]{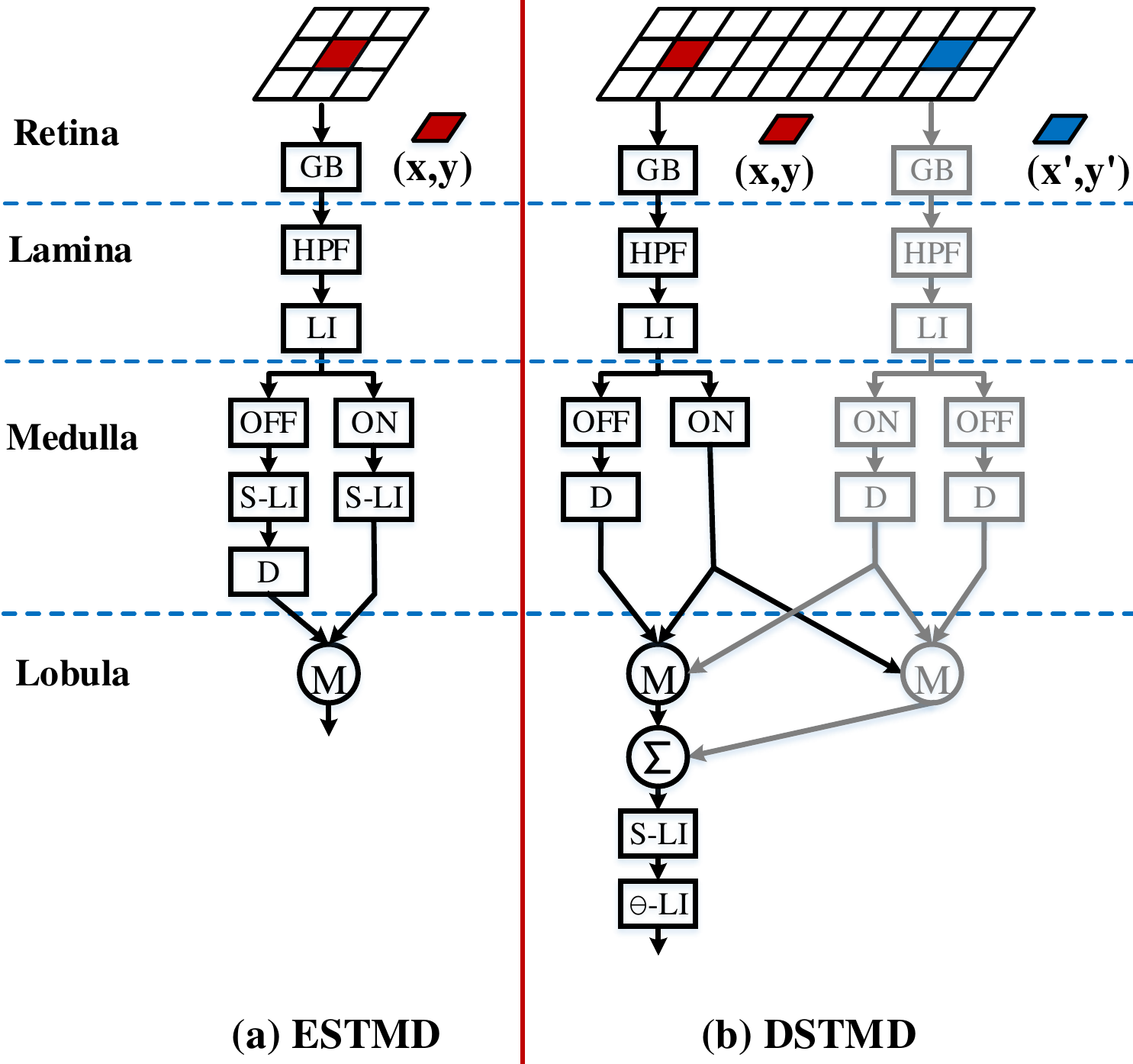}
	\caption{Schematics of an ESTMD (adapted from \cite{Wiederman2008(STMD-clutter)}) and a DSTMD (adapted from \cite{wang2018directionally}) computational models for the detection of small target motion: the ESTMD integrates signals from each single position, whilst the DSTMD has correlations between every two positions.}
	\label{Fig: STMD-ESTMD-DSTMD-Schematic}
\end{figure}

\begin{figure}[H]
	\centering
	\subfloat[]{\includegraphics[width=0.7\textwidth]{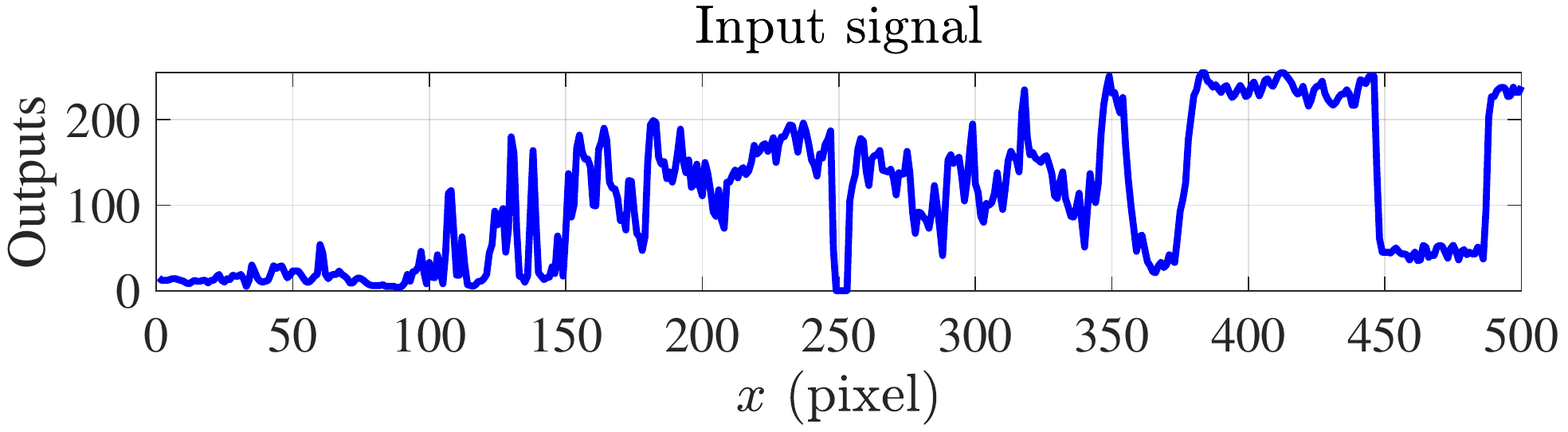}
		\label{STMD-Neural-Layer-Input-Signal}}
	\hspace{5cm}
	\subfloat[]{\includegraphics[width=0.49\textwidth]{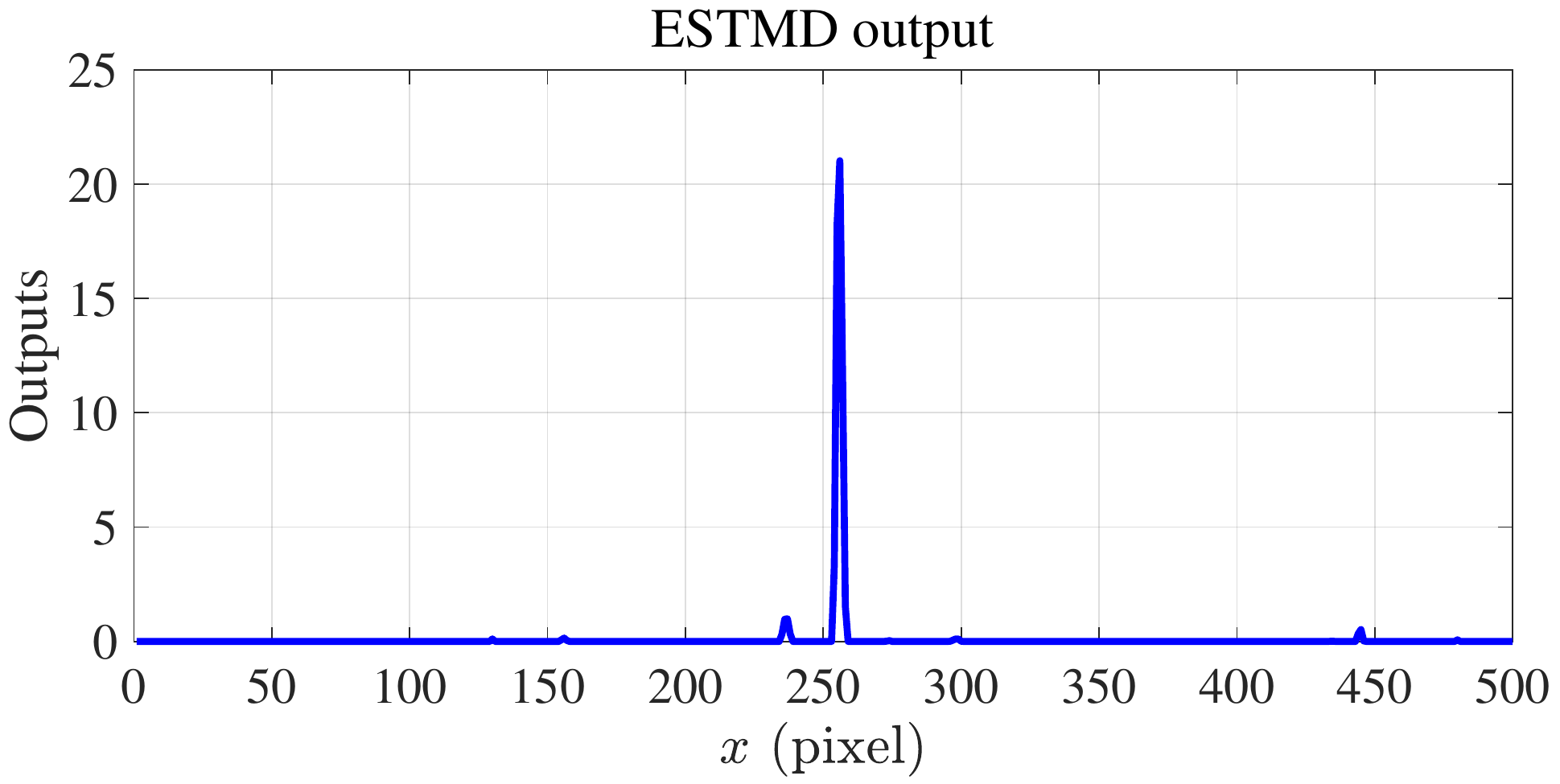}
		\label{STMD-Neural-Layer-ESTMD-Output}}
	\hfill
	\subfloat[]{\includegraphics[width=0.49\textwidth]{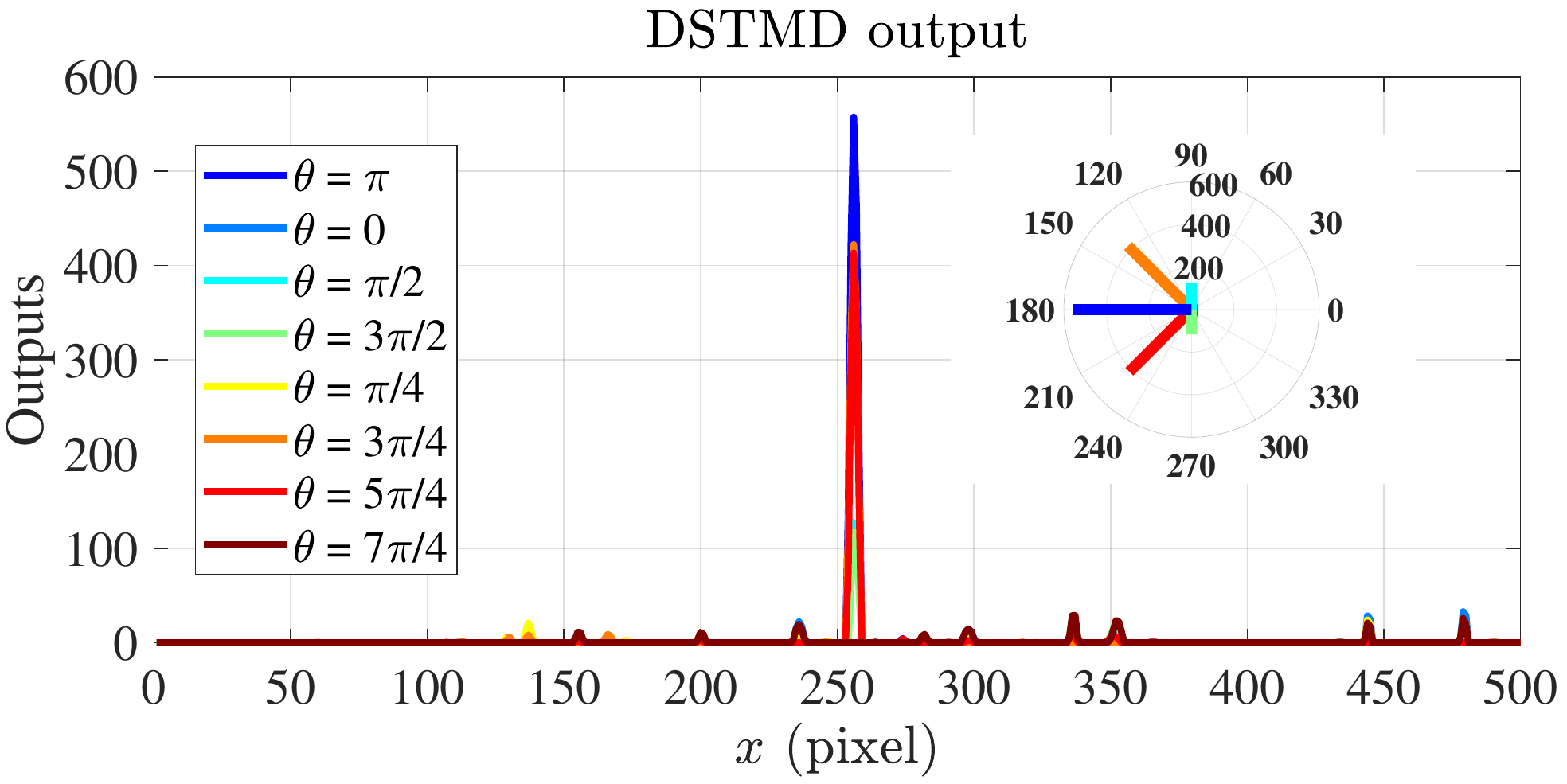}
		\label{STMD-Neural-Layer-DSTMD-Output}}
	\caption{The input signal (a) and the model outputs of the ESTMD (b) and the DSTMD (c): the DSTMD has eight outputs corresponding to eight PDs $\theta$ represented by eight colours in polar coordinate. 
		The angular coordinate denotes the PD motion $\theta$ while the radial coordinate denotes the strength of neural response along this PD.}
	\label{Fig: STMD-Neural-Layer-Output}
\end{figure}

The most significant difference between the STMD and other wide-field motion sensitive neurons, like the LGMD, the DSN, the LPTC and etc., is that the STMD has specific size selectivity to small-field movements. 
More precisely, the STMD represents peak responses to targets subtending $1-3^{\circ}$ of the field of view, yet have no response to larger bars (typically $>10^{\circ}$) or to wide-field grating stimuli \cite{nordstrom2006insect,nordstrom2012neural}. 
To clearly demonstrate the size selectivity of STMD neurons, the response of the STMD neuron to targets of varying heights is shown in the Fig. \ref{Fig: STMD-Size-Tuning}. 
From the Fig. \ref{STMD-Size-Tuning-2006-PlosBology}, we can see that the two smaller targets whose heights are equal to $0.8^{\circ}$ and $3^{\circ}$, respectively, can elicit stronger neural response of the STMD neurons. 
However, the response to the larger target whose height equals to $15^{\circ}$, is much weaker and indistinguishable from spontaneous activity. 
The selectivity of STMD to target height can be clearly seen in the Fig. \ref{STMD-Size-Tuning-2012-Neurobiology}. 
These results demonstrate the STMD has an optimal size sensitivity corresponding to the strongest neural response. 
When target height is higher or lower than that optimal one, the neural response will significantly decrease.

Some STMD neurons have also demonstrated the DS \cite{nordstrom2006small,nordstrom2006insect}. 
These directionally selective STMD neurons respond strongly to small target motion oriented along the PD, but show weaker or no, even fully opponent response to the ND motion. 
The Fig. \ref{STMD-Direction-Tuning-2006-PlosBology} illustrates raw responses of a directionally selective STMD neuron which prefers target motion to left, stimulated by three different-sized targets;
this demonstrates that the larger target whose height equals to $15^{\circ}$, cannot activate the STMD neurone though by the PD motion. 
However, for the smaller targets whose heights are equal to $0.8^{\circ}$ and $3^{\circ}$, the STMD neuron responds strongly to the PD motion. 
On the other hand, when the smaller targets move in the ND, the response of the STMD neuron is not significantly different from spontaneous activity, which means it is inactive. 
In the further research \cite{nordstrom2006insect,barnett2007retinotopic}, biologists assert that both the size and the DS of STMD is independent on background motion. 
More concretely, the STMD will rigorously respond to small target motion against visually cluttered background regardless of background motion direction and velocity. 
In the Fig. \ref{STMD-Background-Resonse-2012-Neurobiology}, we can see that the STMD neuron shows strong response to the small target moving along the PD (downward), but much weaker response to the ND (upward). 
Besides, the response to the small target motion is quite robust in spite of either the direction or the velocity of background motion. 
In another word, the STMD can recognise small target motion even without relative motion between the moving objects and the background.

\subsubsection{Computational models and applications}

On the basis of these biological findings, a few computational models have been put forward to simulate the STMD, in the past decade.
Wiederman et al. proposed a seminal work of an elementary small target motion detector (ESTMD in the Fig. \ref{Fig: STMD-ESTMD-DSTMD-Schematic}) to account for the specific size selectivity of the STMD \cite{Wiederman2008(STMD-clutter)}. 
However, the ESTMD model is unable to realise the DS of the STMD revealed by biologists. 
To implement the DS, two hybrid models, i.e. the ESTMD-EMD and the EMD-ESTMD, were proposed for achieving the DS of the STMD \cite{wiederman2013biologically}. 
More specifically, the ESTMD-EMD indicates that the ESTMD cascades with the EMD while the EMD-ESTMD indicates that the EMD cascades with the ESTMD. 
These two hybrid models have been successfully used for target tracking against cluttered backgrounds in an autonomous mobile ground robot \cite{bagheri2017autonomous, bagheri2017performance, bagheri2015properties}. 
Another directionally selective STMD model, the directionally selective small target motion detector (DSTMD in the Fig. \ref{Fig: STMD-ESTMD-DSTMD-Schematic}), was very recently proposed by Wang et al. \cite{wang2018directionally}. 
Compared to other STMD models, the DSTMD provides unified and rigorous mathematical descriptions. 
Besides, the DS of the DSTMD has been systematically studied and the motion direction of small targets can be estimated \cite{wang2018directionally}, as illustrated in the Fig. \ref{Fig: STMD-Neural-Layer-Output}. 
In comparison with the ESTMD, we can obtain from the Fig. \ref{STMD-Neural-Layer-ESTMD-Output} and \ref{STMD-Neural-Layer-DSTMD-Output} that the most significant difference between the DSTMD and the ESTMD is that the former can generate the DS to small target motion. 
More precisely, in the Fig. \ref{STMD-Neural-Layer-DSTMD-Output}, the DSTMD has eight outputs corresponding to eight PDs $\theta$, $\theta \in \{0, \frac{\pi}{4}, \frac{\pi}{2}, \frac{3\pi}{4}, \pi, \frac{5\pi}{4}, \frac{3\pi}{2}, \frac{7\pi}{4}\}$. 
On the other hand, in the Fig. \ref{STMD-Neural-Layer-ESTMD-Output}, the ESTMD produces only a single directional response. 
To clearly show the DS, the DSTMD responses to a small target are shown in polar coordinates as well in the Fig. \ref{STMD-Neural-Layer-DSTMD-Output}: the DSTMD exhibits the strongest response along the direction $\theta = \pi$ which is consistent with the motion direction of the detected small target translation. 
The other seven directional outputs of DSTMD decrease as the corresponding direction $\theta$ deviates from the target's motion direction.

\subsection{Figure detection neurons}
\label{section: literature: small: fd}

\subsubsection{Biological research}

\begin{figure}[H]
	\centering
	\subfloat[]{\includegraphics[width=0.5\textwidth]{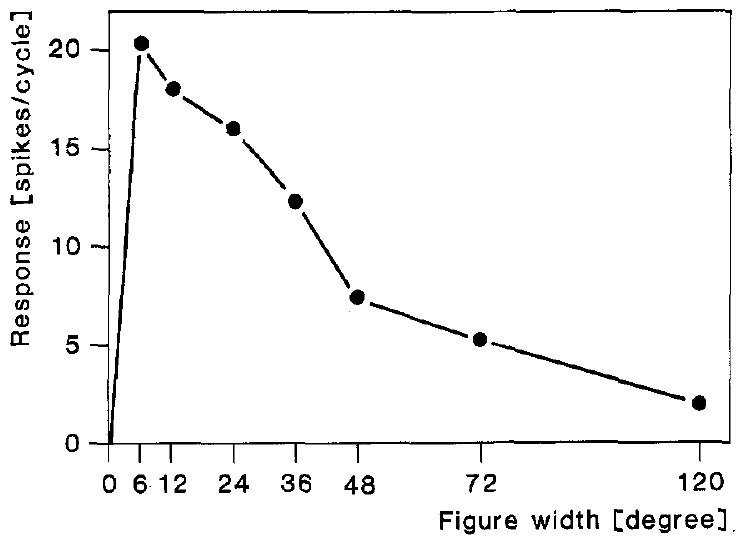}
		\label{FD-Cells-Size-Selectivity}}
	\hfill
	\subfloat[]{\includegraphics[width=0.45\textwidth]{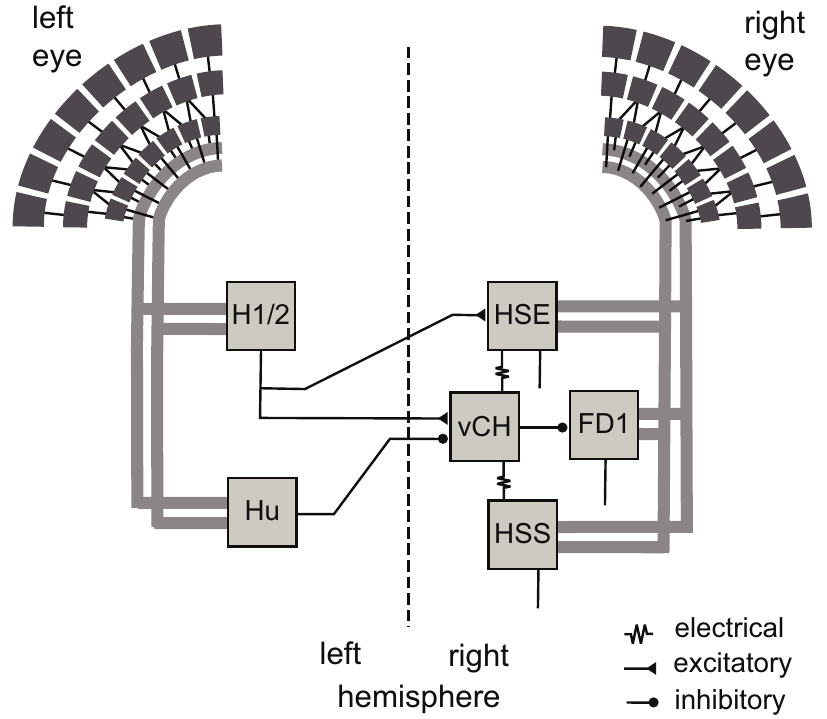}
		\label{FD-Cells-Inhibition-from-Background-Motion}}
	\caption{(a) The response of an FDN to targets of varying width: the horizontal axis denotes the target width (degree) while the vertical axis denotes the neural response (spikes/cycle), adapted from \cite{egelhaaf1985neuronal-3}. 
		(b) Wiring sketch of the FD1 cell input circuit, adapted from \cite{hennig2012neuronal}: the FD1 cell is a most thoroughly analysed FDN.}
	\label{Fig: FD-Cells-Size-Selectivity-Inhibition-from-Background-Motion}
\end{figure}

Moreover, one class of LPTCs, namely figure detection neurons (FDN), has also been demonstrated preference to small targets  \cite{egelhaaf1985neuronal-2,egelhaaf1985neuronal-3,gauck1999spatial,kimmerle2000detection,kimmerle1997object}. 
Although both the FDN and the STMD exhibit size selectivity to moving targets, they differ in the preferred size. 
More specifically, the STMD shows strongest response to targets with the size within $1\sim3$ degrees  \cite{nordstrom2006insect,nordstrom2012neural}. 
However, the FDN responds optimally to targets whose size is in the range of $6\sim12$ degrees \cite{egelhaaf1985neuronal-3,hennig2008distributed}. 
The Fig. \ref{FD-Cells-Size-Selectivity} presents responses of the FDN to targets with varying widths. 
In the Fig. \ref{FD-Cells-Size-Selectivity}, we can see that the optimal width for the FDN is $6$ degrees which is larger than that of the STMD ($1-3$ degrees). 
Another difference between these two types of small target motion sensitive neurons is the underlying mechanisms for size selectivity.	
To be more precise, the STMD does not receive inhibition from wide-field neurons \cite{nordstrom2006insect} and its size selectivity results from a second-order lateral inhibition mechanism \cite{bolzon2009local}.
However, the size selectivity of FDN is assumed to be the result of inhibition from wide-field neurons\cite{warzecha1993neural,hennig2008distributed}. 
The Fig. \ref{FD-Cells-Inhibition-from-Background-Motion} demonstrates a wiring sketch of an FDN input circuit; the FDN is inhibited by the vCH cell \cite{warzecha1993neural} which receives excitatory and inhibitory inputs from other motion sensitive LPTCs including HSE, HSS, H1, H2 and Hu cells \cite{krapp2001binocular,spalthoff2010localized,eckert1983centrifugal}.

\begin{figure}[H]
	\centering
	\includegraphics[width=0.5\textwidth]{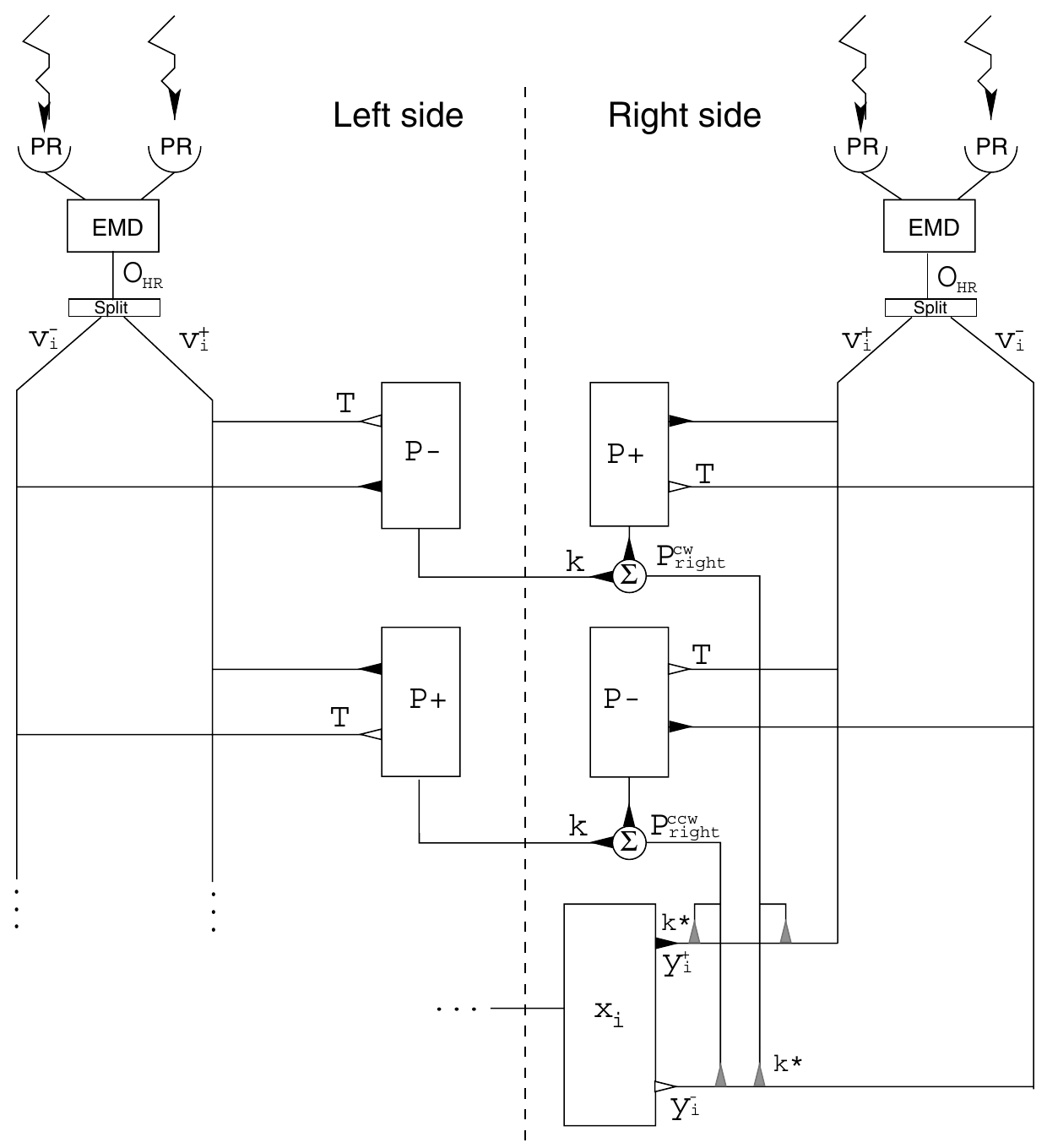}
	\caption{Schematic of an SFS based on the FDN: excitatory and inhibitory synapses are indicated by black and white triangles, respectively. 
		Shunting inhibition is indicated by grey triangles. 
		Responses from neighbouring photoreceptors (PR) are input to EMDs. 
		The EMD outputs are split into positive and negative components. 
		These components are aggregated into directionally selective monocular pool cells ($\text{P}^+,\text{P}^-$) and then into CW and CCW binocular pool cell responses ($\text{P}^{\text{CW}}_{\text{right}},\text{P}^{\text{CCW}}_{\text{right}}$). These directionally selective binocular pool cells interact via shunting inhibition with the individual motion detector output channels prior to their combination by unit $\text{x}_{\text{i}}$ . 
		For simplicity, only the right-side computation is diagrammed. 
		This figure is adapted from \cite{higgins2004elaborated}.}
	\label{Fig: FD-Cells-Model-Schematic}
\end{figure}

\begin{figure}[H]
	\centering
	\includegraphics[width=0.6\textwidth]{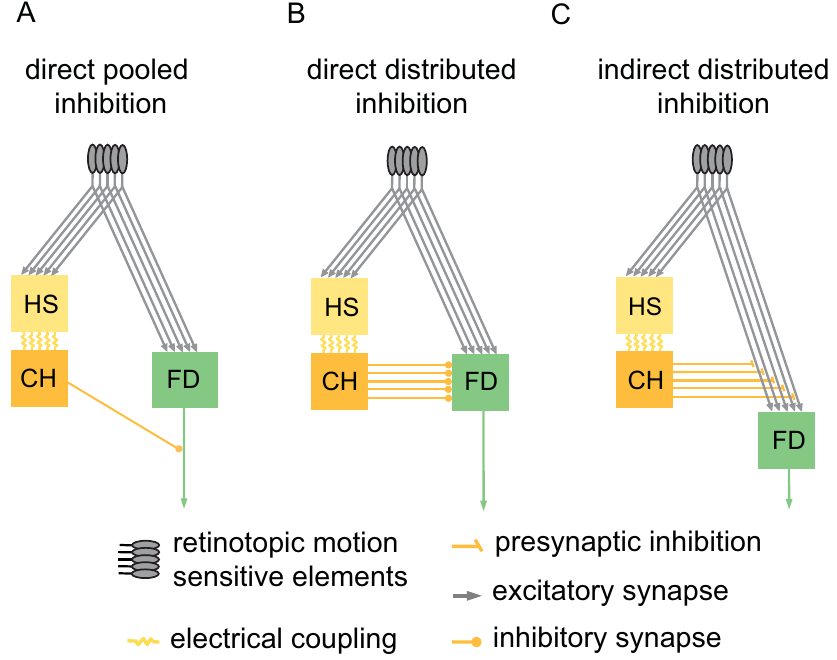}
	\caption{Schematics of potential circuits of the input organisation of an FDN:
		the small-field selective FDN receives excitatory retinotopic input from motion sensitive elements. 
		Inhibitory input of the FDN is mediated by the CH cell via HS cells. 
		For simplicity, only one of the two HS cells that provide input to the CH neuron is shown in this sketch. 
		(A) The CH inhibits the FDN after spatial pooling (`direct pooled inhibition'). 
		(B) The CH inhibits the FDN in a direct distributed way. 
		(C) The CH inhibits the retinotopic input elements of the FDN in an indirect distributed way. 
		This figure is adapted from \cite{hennig2008distributed}.}
	\label{Fig: FD-Cells-Wiring-Schemes}
\end{figure}

\subsubsection{Computational models and applications}

For computationally modelling the FDN, a few models called small field system (SFS) have been proposed to account for the specific size selectivity of FDN \cite{egelhaaf1985neuronal-3,reichardt1983figure,reichardt1989processing}. 
These SFS are quite similar; an instance is shown schematically in the Fig. \ref{Fig: FD-Cells-Model-Schematic}. 
This SFS is composed of an output neuron, the FDN ($\text{x}_{\text{i}}$) which receives retinotopic inputs ($\text{V}_{\text{i}}^{+},\text{V}_{\text{i}}^{-}$), as well as inputs from inhibitory neurons ($\text{P}^{\text{CW}}_{\text{right}},\text{P}^{\text{CCW}}_{\text{right}}$). 
The retinotopic inputs ($\text{V}_{\text{i}}^{+},\text{V}_{\text{i}}^{-}$) denote motion information which is detected by the EMDs. 
In \cite{hennig2012neuronal}, these SFS were modified to allow a simulated fly to track a small moving target in a virtual environment. 
Although the size selectivity of FDN results from the inhibitions from wide-filed neurons, the wiring scheme between the wide-field neurons and the FDNs is still unclear. 
In \cite{hennig2008distributed}, Hennig et al. analysed three kinds of wiring schemes between the wide-field neurons and the FDN, based on new anatomical and electro-physiological findings \cite{haag2002dendro,egelhaaf1993neural}, as illustrated in the Fig. \ref{Fig: FD-Cells-Wiring-Schemes}. 
The authors indicated that the latter two wiring schemes, i.e. direct distributed inhibition and indirect distributed inhibition in the Fig. \ref{Fig: FD-Cells-Wiring-Schemes}, can account well for the size selectivity of FDN and the dependence of FDN on the relative velocity between the small target and the background. 
In \cite{hennig2012neuronal}, Hennig et al. further improved the existing SFS by modelling the pre-synaptic neurons to the FDN, including H1, Hu, HSE and HSS \cite{hennig2011binocular}, as illustrated in the Fig. \ref{FD-Cells-Inhibition-from-Background-Motion}). 
They integrated the responses of pre-synaptic neurons in the proposed FDN circuit; this effectively matches the corresponding biological structure. 
Besides, comparing to the existing studies on modelling FDN that all use simple synthetic stimuli, the authors applied naturalistic stimuli to test the proposed FDN circuit and demonstrated its characteristics.

\section{Discussion}
\label{Sec: Discussion}

Through above survey, we have introduced motion perception visual models that possess different direction and size selectivity originating from insect visual systems, as well as their background biological research and corresponding applications in artificial mobile machines like robots, MAVs, UAVs and ground vehicles for motion perception. 
These models can sense different motion patterns including looming, directional translation and small target motion. 
This section will further discuss about the similarities in modelling of different insect motion detectors, and summarise the computational generation of both the direction and the size selectivity to proposed different motion patterns, and also point out existing and possible hardware implementations.

\subsection{Similarities in different motion perception models}
\label{Sec: Discussion: Similarity}

Though these motion detectors demonstrate different direction or size selectivity, there are similarities that can be summarised through these computational studies. 
Taken the fly and the locust inspired visual neural networks or models as examples, a great majority of these models have been focusing on structural modelling of internal circuits or pathways of insect visual systems. 
These models can share some similar visual processing methods:
\begin{enumerate}
	\item In most insect species, vision is mediated by compound eyes in the first layer of Retina \cite{Horridge-1977-insect-compound-eye,Cheng-review-compound-eye-imaging}, which consists of photoreceptors capturing grey-scale or single-channel (normally the green channel) images sequence. 
	This neuropile layer retrieves motion information by spatial filtering of input signals at the ommatidia level \cite{Horridge-1977-insect-compound-eye}, as shown in the LGMD models in the Fig. \ref{Fig: LGMD1-network-Rind-1996}, \ref{Fig: LGMD1-Glayer}, \ref{Fig: LGMD1-other-models}, \ref{Fig: LGMD1-Badia2010}, \ref{Fig: LGMD2-models}, \ref{Fig: LGMDs-generic-model}, and the fly LPTC models in the Fig. \ref{Fig: Fly-Pathways-LPTCs}, \ref{Fig: Fly-polarity-models}, and the STMD models in the Fig. \ref{Fig: STMD-ESTMD-DSTMD-Schematic}. 
	\item With respect to the biological findings of ON and OFF pathways in many animals including the various kinds of flies, the second layer of Lamina consists of rectifying transient cells separating visual signals into parallel channels.
	Although this structure has not yet been found in locusts, recent studies on the LGMD1 and the LGMD2 may evidence similar ON (onset) and OFF (offset) mechanisms in the locust visual systems \cite{Fu-2018(LGMD1-NN),Fu2017a(LGMDs-IROS),Fu-2015(LGMD2-MLSP),Fu-2016(LGMD2-BMVC),LGMDs-2016}.
	\item Within the computational layers of Medulla and Lobula, both the direction and the size selectivity is generated and sharpened up to specific motion patterns. 
	Both the lateral inhibition mechanism and the HR-like non-linear computation can contribute to shape the specific DS.
	\item The modelled LGMDs and LPTCs and DSNs are wide-field motion sensitive visual neurons which pool the intact pre-synaptic local directional motion information and then generate spikes toward further sensorimotor neural systems. 
	However, the small-field STMD models generate the DS to small target movements in every local pixel-scale field.
	\item We highlight the functionality of ON and OFF visual pathways that can explain biological visual processing in insect motion sensitive circuits. 
	Such a structure can be modelled in different motion perception neuron models including LGMDs, DSNs, LPTCs and STMDs and etc. 
	In addition, the modelling of motion sensitive visual systems in other animals like the LGNs in crabs \cite{Crab2014(computation-approach-neurons)} may learn from the existing models.
\end{enumerate}

\subsection{Realisation of direction and size selectivity to different motion patterns}
\label{Sec: Discussion: Direction}

This subsection summarises generation of both the direction and the size selectivity diversity in these computational motion perception neural networks or models.
Through above reviews of looming and translation sensitive neural systems, we summarise that 
1) the different direction selectivity of various models is shaped pre-synaptic to the wide-field motion detectors of the DSNs, the LPTCs and the LGMDs, that is, in the Medulla or the Lobula neuropile layers, as illustrated in the Fig. \ref{Fig: LGMD1-neuromorphology} and \ref{Fig: Fly-eyes-morphology}; 
2) the spatiotemporal computation including the lateral \cite{LGMD1-1996(Rind-neural-network),Yue-2013(locust-DSNs)} and the self \cite{LGMDs-2016,Fu2017a(LGMDs-IROS)} inhibition mechanisms and the non-linear interactions between neighbouring cells \cite{EMD-1989(principles-review)} can well mediate the specific DS to either looming or translation motion pattern.

Specifically for locusts, as reviewed in the Section \ref{Sec: Looming-Motion} and the Section \ref{section: literature: translating: locust-dsn}, there are two types of motion sensitive visual neurons, i.e., the LGMDs and the DSNs. 
Though they are applied as collision-detecting sensors, the different DS exists between the LGMDs and the DSNs, as the schematics shown in the Fig. \ref{Fig: LGMD1-Glayer}, \ref{Fig: LGMD2-models} and \ref{Fig: Locust-DSNs-models}. 
Firstly for realising the specific DS of the locust LGMDs to looming stimuli only, these neuron models or neural networks have been demonstrated the lateral only or the lateral-and-self inhibition mechanisms of spatiotemporal convolution processes. 
More precisely, the inhibitions in the computational Medulla layer are formed by convolving surrounding and symmetrically spreading out excitations with temporal delay \cite{LGMD1-Glayer(feature-enhancement),LGMD1-1996(Rind-neural-network),LGMD1-DSN-competing(LGMD1-DSNs-Hybrid),Yue-2006(LGMD1-car),LGMDs-2016}. 
That is,
\begin{equation}
I(x,y,t) = \iiint E(u,v,s)\ W(x-u,y-v,t-s)\ d u d v d s,
\label{lgmds-convolving}
\end{equation}
where $W$ is a local convolution kernel. 
$I$ and $E$ denote the inhibition and excitation cells both in a three-dimensional form. 
After that, the excitatory and inhibitory signals compete with each other at every local cell. 
That is,
\begin{equation}
S(x,y,t) = E(x,y,t) - w \cdot I(x,y,t),
\label{lgmds-competition}
\end{equation}
where $S$ denotes the summation cells and $w$ is a local bias. 
As a result, the excitations are cut down by the inhibitions. 
This mechanism plays a crucial role of shaping the selectivity in the LGMDs models which respond most strongly to looming (expanding of object edges) versus translating and receding objects. 
In addition, with regard to the modelling of ON and OFF mechanisms in the LGMDs models, the excitations can be also formed by convolving surrounding delayed inhibitions \cite{Fu-2016(LGMD2-BMVC),Fu-2018(LGMD1-NN),Fu2017a(LGMDs-IROS)}, the calculation of which is similar to the Eq. \ref{lgmds-convolving}.

Secondly for the locust DSNs, each directionally specific neuron responds to motion oriented along a particular PD. 
With similar ideas to the modelling of LGMD, the DS can be realised by a directional convolution process on asymmetrically spreading out inhibitions with temporal delay \cite{Yue-2007(locust-DSNs),Yue-2013(locust-DSNs),LGMD1-DSN-competing(LGMD1-DSNs-Hybrid)}. 
That is,
\begin{equation}
I(x,y,t) = \iint E(u,y,s)\ W(x-u,y,t-s)\ d u d s,
\label{dsns-convolving-hor}
\end{equation}
in horizontal directions and,
\begin{equation}
I(x,y,t) = \iint E(x,v,s)\ W(x,y-v,t-s)\ d v d s,
\label{dsns-convolving-ver}
\end{equation}
in vertical directions. 
After that, there are also competitions between every local excitation and inhibition cell. 
The calculation conforms to the Eq. \ref{lgmds-competition}.

With regard to the fly EMD and LPTC models for translation perception as reviewed in the Section \ref{section: literature: translating: fly-emd} and the Section \ref{section: literature: translating: fly-lptc}, the specific DS to four cardinal directions in the field of view (front-to-back, back-to-front, upward and downward) is implemented by non-linearly spatiotemporal computations according to the classic HRC \cite{EMD-1989(principles-review)}. 
That is,
\begin{equation}
R(t) = X_1(t-\epsilon) \cdot X_2(t) - X_1(t) \cdot X_2(t-\epsilon),
\label{hr}
\end{equation}
where $R$ is the output of each pairwise motion detectors in space. 
$X_1$ and $X_2$ are two adjacent motion sensitive cells, and $\epsilon$ is the temporal delay.
Such a theory or its derived versions have been very widely used in a variety of the fly EMDs models, e.g. \cite{Iida_2000(fly-visual-odometer),Zanker_2005(motion-signal-outdoor),EMD-MainArticle}, and the fly OF based strategies, e.g. \cite{Serres2017(review-optic-flow),Nicolas-2014(Review-Fly-Robot),Ruffier-OF-regulation-autopilot}, and the fly ON-OFF polarity motion detectors, e.g. \cite{Eichner2011(2Q-motion),Clark_2011(6Q-model-fly),Joesch_2013(functional-ON-OFF),Circuit-genetic(genetic-push-motion)}, and the fly LPTCs models, e.g. \cite{Fu2017(fly-DSNs-IJCNN),Fu-2017(ROBIO-fixation),HWang-ICDL(LPTC-model)}, and the insect directional STMD models, e.g. \cite{Wang-2016(IJCNN-STMD),Wiederman2008(STMD-clutter),wang2018directionally,Hongxin-feedback-stmd}, as well as the bee angular velocity detecting models, e.g. \cite{Cope2016(model-angular-bee),Huatian-ICANN-angular}.

Regarding to computational generation of the size selectivity, as reviewed in the Section \ref{Sec: Small-Target-Motion}, the STMD and the FDN are small-field motion sensitive neurons which have the specific size selectivity to small target motion that is different from these wide-field motion detectors. 
There have been two basic categories of STMD based visual neural networks, i.e., the ESTMD and the DSTMD. 
The latter one possesses the direction selectivity to small target motion that can be achieved by similar methods to the EMDs. 
Wiederman et al. have proposed that the lateral inhibition mechanism plays a crucial role to adjust the size selectivity via spatiotemporal neural computation \cite{Wiederman2008(STMD-clutter),Wiederman2013(ON-OFF-correlation)}. 
Derived from this theory, Wang et al. mathematically analysed the way of generating the size selectivity in motion sensitive visual pathways of insects: in this research, they applied a second-order lateral inhibition mechanism in the computational layer of Lobula which can be represented by an algorithm of `Difference of Gaussians' \cite{wang2018directionally}.

\subsection{Multiple neural systems integration}
\label{Sec: Discussion: Multi}

These proposed insect visual pathways or neurons are functionally specialised in recognising different motion patterns containing looming, directional translation and small target movements.
In animals' visual brains, evidence has been given that the complex visuomotor response is guided by collaboration of various visual neurons or circuits, rather than a single unit alone.
However, the underlying mechanisms still remain elusive. 
While the biological substrates are unknown, the computational modelling is of particular usefulness to help explain the mysterious biological visual processing. 
Most of the current state-of-art computational models implement a single kind of neural systems. 
From a modeller's perspective, integrating multiple neural pathways or neurons can undoubtedly benefit the motion perception within more complicated dynamic visual environments involving diverse motion patterns. 
In addition, this can also make the intelligent machines smarter for dealing with mixed visual cues and adopting more appropriate visually-guided behaviours like insects.

Taken some example computational studies, we discuss about the advantages of multiple systems integration for motion perception.
First, the locust DSNs based visual neural networks proposed in \cite{Yue-2007(locust-DSNs),Yue-2013(locust-DSNs),LGMD1-DSN-competing(LGMD1-DSNs-Hybrid)} themselves are paradigms of integrating multiple neural pathways, as illustrated in the Fig. \ref{Fig: Locust-DSNs-models}), since each directionally pathway is sensitive to a particular PD motion and the post-synaptic organisation of multiple DSNs can match well the requirements of collision recognition in dynamic scenes. 
Second, combining the translation sensitive neural pathways with the LGMD neuron model can effectively enhance the collision selectivity, especially in complex driving scenarios \cite{Yue-2006(LGMD1-TSNN),Zhang-2016(IJCNN-hybrid)}, since the translation and the looming perception pathways are perfectly complementary in functions. 
Third, as mentioned in the Section \ref{Sec: Looming-Motion}, the LGMD1 and the LGMD2 have different looming selectivity. 
We have suggested that combining the functionality of both neurons can enhance the collision-detecting performance in either dark or bright environments. 
With regard to this idea, a case study has initially demonstrated the usefulness of incorporating in the LGMD1 an LGMD2 neural pathway for collision detection in mobile ground robot scenarios under different illumination conditions \cite{Fu2017a(LGMDs-IROS)}. 
In addition, Fu and Yue recently investigated the possible method of integrating multiple visual pathways in the \textit{Drosophila}'s brain for fast motion tracking and implementing a closed-loop behavioural response to fixation \cite{Fu-2017(ROBIO-fixation)}. 
This approach has been also built on the embedded system in a miniaturised mobile robot \cite{Fu-ROBIO-2018}. 
Furthermore, a visual neural network that sense rotational or spiral motion patterns integrated mechanisms of the locust DSNs and the fly EMDs in a computational structure \cite{Hu2016(RMPNN),Hu2018(RMPNN)}: this model can well recognise both clockwise and anticlockwise rotations of an object in a simple background.

To sum up briefly, an artificial vision system that possesses robust functions to detect multiple motion patterns and extract more abundant features from a visually dynamic and cluttered environment is very necessary for further intelligent machines like self-driving cars to better serve the human society. 
The computational modelling and applications of insect visual systems can provide us with effective and efficient solutions.

\subsection{Hardware realisation of insect motion perception models}
\label{Sec: Discussion: Neuromorphic}

Continued with the surveys on computational models and applications of insect visual systems, this subsection discusses about the relevant hardware realisation of these models and the future trends. 
We propose that to achieve higher processing speed, larger scale or real-time solutions, the implementation of neuromorphic visual models on hardware could be extremely advantageous.

From an engineering perspective, the neuromorphic visual sensors are realised towards two different trends.	
One is single-chip solution featured by the compact size and specialised functions. 
Another trend is featured by high performance circuits such as the FPGA.

The single-chip solutions, e.g.\cite{Abbott1995,Indiveri1998(VLSI-model-approach),Harrison2005, Sarkar2013,Vanhoutte-time-of-travel-OF-micro-flying-robot,Nicolas-group-CurvACE}, are usually implemented by CMOS VLSI process with mixed-signal \cite{Neurorobotics2017(avoidance-acquisition-neuromorphic)}. 
The photoreceptors in compound eyes can also be integrated inside it like the former-mentioned silicon retina \cite{Vanhoutte-time-of-travel-OF-micro-flying-robot} and CurvACE sensor \cite{Nicolas-group-CurvACE}. 
Taking advantage from the compact design and low power-consumption, these silicon implementations could be widely deployed as individual sensors for distributed systems, e.g. \cite{Nicolas-group-CurvACE,Roubieu-1-gram-sensor-EMD,Kramer-CMOS-VLSI-EMD,Moeckel-Liu-time-to-travel-model,Viollet-OF-VLSI-sensors}, or as components on size-sensitive platforms such as micro robots, e.g. \cite{Vanhoutte-time-of-travel-OF-micro-flying-robot,Mafrica-OF-minisensor-robot} and MAVs, e.g. \cite{Expert-OF-uneven-terrain-following,Ruffier-MAV-OF-circuits,Ruffier-MAV-OF-ICRA,Kerhuel-VODKA-sensor} and UAVs, e.g. \cite{UAV2017(LGMD1-spiking),Sabiron-lightweight-sensor-OF-helicopter-outdoor}. 
This kind of integrated chip can also be utilised as an optical sensor for further applications. 
For instance, the dynamic vision sensor (DVS) \cite{Posch2011, Milde2015} technology is featured by its low-latency and low-data volume.																																																	

On the other hand, the high-performance solutions aim to capture images from commercial cameras with high resolution and high frame rate, and to be established the signal processing within the FPGA \cite{Zhang2008, Meng-2010(LGMD1-FPGA), Koehler2015} or even the application specific integrated circuits (ASICs). 
Due to the feature that data array can be dealt in parallel, the total frame rate can reach even up to $350$ fps at the resolution of $256\times256$ \cite{Zhang2008}, or $5$~kHz with $12$ photo-diodes \cite{Aubepart2007}. 
These high-performance approaches could significantly enhance the visual model's spatial sensitivity and temporal response for further researches with critical requirements. 

As presented above, these bio-inspired motion perception models could be ideal choices for design of neuromorphic vision sensors as a future trend of hardware realisation of visual processing. 
Furthermore, these low-energy and miniaturised visual sensing modalities would be able to incorporated in some control systems for much broader applications in robotics, such as the under-actuated systems \cite{Pengcheng-NonlinearDynamics}, and corresponding bio-inspired robot applications like the vibro-driven robot \cite{Pengcheng-IROS18}.

\section{Conclusion}
\label{Sec: Conclusion}

In this article, we have provided an overview of computational motion perception models originating from insect visual systems research, as well as corresponding applications to artificial mobile machines for visual motion detection and insect-like behaviours control like obstacle avoidance, landing, tunnel crossing, terrain following, fixation and etc. 
We have reviewed these motion perception models according to their specific direction and size selectivity to different motion patterns including looming, translation and small target motion. 
To a large extent, the physiology underlying motion perception in insect visual systems is still unknown. 
However, the diversity in direction and size selectivity in different types of visual neurons can be realised by spatiotemporal computation within the neural circuits or pathways. 
We have summarised different methodologies including lateral inhibition mechanisms and non-linear computation to implement different selectivity. 
In addition, both biological and modelling studies, over decades, have demonstrated the similarities in different insect motion detectors. 
The effectiveness and efficiency of these bio-inspired models have been validated by a variety of applications to bio-robotics and other vision-based platforms for motion perception in a both low-power and fast mode. 
Through the existing modelling studies, we have pointed out the great potential of these dynamic vision systems in building neuromorphic sensors for volume production and utility in future intelligent machines.

\section*{Acknowledgement}
This research was supported by the EU Horizon 2020 projects STEP2DYNA (691154) and ULTRACEPT (778062).

\section*{References}

\bibliography{reviewbib}

\begin{thebibliography}{250}
\expandafter\ifx\csname natexlab\endcsname\relax\def\natexlab#1{#1}\fi
\expandafter\ifx\csname url\endcsname\relax
  \def\url#1{{\tt #1}}\fi
\expandafter\ifx\csname urlprefix\endcsname\relax\def\urlprefix{URL }\fi

\bibitem[{Abbott et~al.(1995)Abbott, Moini, Yakovleff, Nguyen, Blanksby, Kim,
  Bouzerdoum, Bogner, \& Eshraghian}]{Abbott1995}
Abbott, D., Moini, A., Yakovleff, A., Nguyen, X.~T., Blanksby, A., Kim, G.,
  Bouzerdoum, A., Bogner, R.~E., \& Eshraghian, K. (1995).
\newblock New {VLSI} smart sensor for collision avoidance inspired by insect
  vision.
\newblock In {\em Proc. SPIE 2344, Intelligent vehicle highway systems\/}, vol.
  2344, (pp. 105--115). SPIE.

\bibitem[{Aptekar et~al.(2012)Aptekar, A.Shoemaker, \&
  Frye}]{Figure_2012(tracking-fly-parallel)}
Aptekar, J.~W., A.Shoemaker, P., \& Frye, M.~A. (2012).
\newblock Figure tracking by flies is supported by parallel visual streams.
\newblock {\em Current Biology\/}, {\em 22\/}(6), 482--487.

\bibitem[{Aptekar \& Frye(2013)}]{Motion-2013(higher-order-discrimination)}
Aptekar, J.~W., \& Frye, M.~A. (2013).
\newblock Higher-order figure discrimination in fly and human vision.
\newblock {\em Current Biology\/}, {\em 23\/}(16), R694--R700.

\bibitem[{Arenz et~al.(2017)Arenz, Drews, Richter, Ammer, \&
  Borst}]{Fly-DS-2017(preferred-null)}
Arenz, A., Drews, M.~S., Richter, F.~G., Ammer, G., \& Borst, A. (2017).
\newblock The temporal tuning of the drosophila motion detectors is determined
  by the dynamics of their input elements.
\newblock {\em Current Biology\/}, {\em 27\/}(7), 929--944.

\bibitem[{{Argyros} et~al.(2004){Argyros}, {Tsakiris}, \&
  {Groyer}}]{Robots-2004(mobile-panoramic-sensors)}
{Argyros}, A.~A., {Tsakiris}, D.~P., \& {Groyer}, C. (2004).
\newblock Biomimetic centering behavior [mobile robots with panoramic sensors].
\newblock {\em IEEE Robotics Automation Magazine\/}, {\em 11\/}(4), 21--30.

\bibitem[{Arkin et~al.(2000)Arkin, Ali, Weitzenfeld, \&
  Cervantes-P{\'{e}}rez}]{Mantis-2000(behavioural-model-robot)}
Arkin, R.~C., Ali, K., Weitzenfeld, A., \& Cervantes-P{\'{e}}rez, F. (2000).
\newblock Behavioral models of the praying mantis as a basis for robotic
  behavior.
\newblock {\em Robotics and Autonomous Systems\/}, {\em 32\/}(1), 39--60.

\bibitem[{Aub{\'e}part \& Franceschini(2007)}]{Aubepart2007}
Aub{\'e}part, F., \& Franceschini, N. (2007).
\newblock Bio-inspired optic flow sensors based on {FPGA}: Application to
  micro-air-vehicles.
\newblock {\em Microprocessors and Microsystems\/}, {\em 31\/}(6), 408--419.

\bibitem[{Badia et~al.(2010)Badia, Bernardet, \&
  Verschure}]{Badia-2010(LGMD1-nonlinear-model)}
Badia, S. B.~I., Bernardet, U., \& Verschure, P.~F. (2010).
\newblock Non-linear neuronal responses as an emergent property of afferent
  networks: A case study of the locust lobula giant movement detector.
\newblock {\em PLoS Computational Biology\/}, {\em 6\/}(3), e1000701.

\bibitem[{Bagheri et~al.(2017{\natexlab{a}})Bagheri, Cazzolato, Grainger,
  O’Carroll, \& Wiederman}]{bagheri2017autonomous}
Bagheri, Z.~M., Cazzolato, B.~S., Grainger, S., O’Carroll, D.~C., \&
  Wiederman, S.~D. (2017{\natexlab{a}}).
\newblock An autonomous robot inspired by insect neurophysiology pursues moving
  features in natural environments.
\newblock {\em Journal of Neural Engineering\/}, {\em 14\/}(4), 046030.

\bibitem[{Bagheri et~al.(2015)Bagheri, Wiederman, Cazzolato, Grainger, \&
  O'Carroll}]{bagheri2015properties}
Bagheri, Z.~M., Wiederman, S.~D., Cazzolato, B.~S., Grainger, S., \& O'Carroll,
  D.~C. (2015).
\newblock Properties of neuronal facilitation that improve target tracking in
  natural pursuit simulations.
\newblock {\em Journal of The Royal Society Interface\/}, {\em 12\/}(108),
  20150083.

\bibitem[{Bagheri et~al.(2017{\natexlab{b}})Bagheri, Wiederman, Cazzolato,
  Grainger, \& O’Carroll}]{bagheri2017performance}
Bagheri, Z.~M., Wiederman, S.~D., Cazzolato, B.~S., Grainger, S., \&
  O’Carroll, D.~C. (2017{\natexlab{b}}).
\newblock Performance of an insect-inspired target tracker in natural
  conditions.
\newblock {\em Bioinspiration {\&} Biomimetics\/}, {\em 12\/}(2), 025006.

\bibitem[{Bahl et~al.(2013)Bahl, Ammer, Schilling, \&
  Borst}]{Fly-2013(motion-blind-tracking)}
Bahl, A., Ammer, G., Schilling, T., \& Borst, A. (2013).
\newblock Object tracking in motion-blind flies.
\newblock {\em Nature Neuroscience\/}, {\em 16\/}(6), 730--738.

\bibitem[{Baird et~al.(2013)Baird, Boeddeker, Ibbotson, \&
  Srinivasan}]{Landing-2013(visually-guided-strategy)}
Baird, E., Boeddeker, N., Ibbotson, M.~R., \& Srinivasan, M.~V. (2013).
\newblock A universal strategy for visually guided landing.
\newblock {\em Proceedings of the National Academy of Sciences\/}, {\em
  110\/}(46), 18686--18691.

\bibitem[{Baird et~al.(2010)Baird, Kornfeldt, \&
  Dacke}]{Baird2010(viewing-angle-bumblebees)}
Baird, E., Kornfeldt, T., \& Dacke, M. (2010).
\newblock Minimum viewing angle for visually guided ground speed control in
  bumblebees.
\newblock {\em Journal of Experimental Biology\/}, {\em 213\/}(10), 1625--1632.

\bibitem[{Barlow \& Levick(1965)}]{Barlow-1965(rabbit-DSNs)}
Barlow, H., \& Levick, W. (1965).
\newblock The mechanism of directionally selective units in rabbit's retina.
\newblock {\em Journal of Physiology\/}, {\em 178\/}, 477--504.

\bibitem[{Barnett et~al.(2007)Barnett, Nordstr{\"o}m, \&
  O'Carroll}]{barnett2007retinotopic}
Barnett, P.~D., Nordstr{\"o}m, K., \& O'Carroll, D.~C. (2007).
\newblock Retinotopic organization of small-field-target-detecting neurons in
  the insect visual system.
\newblock {\em Current Biology\/}, {\em 17\/}(7), 569--578.

\bibitem[{Behnia et~al.(2014)Behnia, Clark, Carter, Clandinin, \&
  Desplan}]{Fly-2014(processing-properties-motion)}
Behnia, R., Clark, D.~A., Carter, A.~G., Clandinin, T.~R., \& Desplan, C.
  (2014).
\newblock Processing properties of on and off pathways for drosophila motion
  detection.
\newblock {\em Nature\/}, {\em 512\/}(7515), 427--430.

\bibitem[{Bengochea \& {Ber{\'{o}}n de
  Astrada}(2014)}]{Crab-2014(organization-columnar-visual)}
Bengochea, M., \& {Ber{\'{o}}n de Astrada}, M. (2014).
\newblock Organization of columnar inputs in the third optic ganglion of a
  highly visual crab.
\newblock {\em Journal of Physiology Paris\/}, {\em 108\/}(2-3), 61--70.

\bibitem[{Bermudez~i Badia \& Verschure(2004)}]{LGMD1-2004(Badia)}
Bermudez~i Badia, S., \& Verschure, P.~F. (2004).
\newblock A collision avoidance model based on the lobula giant movement
  detector ({LGMD}) neuron of the locust.
\newblock In {\em Proceedings of the 2004 IEEE international joint conference
  on neural networks (IJCNN)\/}, vol.~3, (pp. 1757--1761). IEEE.

\bibitem[{Bertrand et~al.(2015)Bertrand, Lindemann, \&
  Egelhaaf}]{OpticFlow-2015(colision-goal-direction)}
Bertrand, O. J.~N., Lindemann, J.~P., \& Egelhaaf, M. (2015).
\newblock A bio-inspired collision avoidance model based on spatial information
  derived from motion detectors leads to common routes.
\newblock {\em PLOS Computational Biology\/}, {\em 11\/}(11), 1--28.

\bibitem[{Biswas \& Lee(2017)}]{Fly-Motion-2017(cellular-hybrid-detector)}
Biswas, T., \& Lee, C.~H. (2017).
\newblock Visual motion: Cellular implementation of a hybrid motion detector.
\newblock {\em Current Biology\/}, {\em 27\/}(7), R274--R276.

\bibitem[{Blanchard et~al.(2000)Blanchard, Rind, \&
  Verschure}]{LGMD1-2000(model)}
Blanchard, M., Rind, F.~C., \& Verschure, P.~F. (2000).
\newblock Collision avoidance using a model of the locust lgmd neuron.
\newblock {\em Robotics and Autonomous Systems\/}, {\em 30\/}(1), 17--38.

\bibitem[{Blanchard et~al.(2001)Blanchard, Rind, \&
  Verschure}]{LGMD1-2001(sensory-coding-robot)}
Blanchard, M., Rind, F.~C., \& Verschure, P.~F. (2001).
\newblock How accurate need sensory coding be for behaviour? experiments using
  a mobile robot.
\newblock {\em Neurocomputing\/}, {\em 38-40\/}, 1113--1119.

\bibitem[{Boeddeker et~al.(2005)Boeddeker, Lindemann, Egelhaaf, \&
  Zeil}]{Boeddeker2005(blowfly-neurons-opticflow)}
Boeddeker, N., Lindemann, J.~P., Egelhaaf, M., \& Zeil, J. (2005).
\newblock Responses of blowfly motion-sensitive neurons to reconstructed optic
  flow along outdoor flight paths.
\newblock {\em Journal of Comparative Physiology A: Neuroethology, Sensory,
  Neural, and Behavioral Physiology\/}, {\em 191\/}, 1143--1155.

\bibitem[{Bolzon et~al.(2009)Bolzon, Nordstr{\"o}m, \&
  O'Carroll}]{bolzon2009local}
Bolzon, D.~M., Nordstr{\"o}m, K., \& O'Carroll, D.~C. (2009).
\newblock Local and large-range inhibition in feature detection.
\newblock {\em Journal of Neuroscience\/}, {\em 29\/}(45), 14143--14150.

\bibitem[{Borst(2014)}]{Borst-2014(review-fly)}
Borst, A. (2014).
\newblock Fly visual course control: Behaviour, algorithms and circuits.
\newblock {\em Nature Reviews Neuroscience\/}, {\em 15\/}, 590--599.

\bibitem[{Borst \& Egelhaaf(1989)}]{EMD-1989(principles-review)}
Borst, A., \& Egelhaaf, M. (1989).
\newblock Principles of visual motion detection.
\newblock {\em Trends in Neurosciences\/}, {\em 12\/}, 297--306.

\bibitem[{Borst \& Euler(2011)}]{Borst2011(review-motion)}
Borst, A., \& Euler, T. (2011).
\newblock Seeing things in motion: Models, circuits, and mechanisms.
\newblock {\em Neuron\/}, {\em 71\/}(6), 974--994.

\bibitem[{Borst \& Haag(2002)}]{Borst-2002(review-networks-fly)}
Borst, A., \& Haag, J. (2002).
\newblock Neural networks in the cockpit of the fly.
\newblock {\em Journal of Comparative Physiology\/}, {\em 188\/}(6), 419--437.

\bibitem[{Borst et~al.(2010)Borst, Haag, \&
  Reiff}]{Borst-2010(review-fly-vision)}
Borst, A., Haag, J., \& Reiff, D.~F. (2010).
\newblock Fly motion vision.
\newblock {\em The Annual Review of Neuroscience\/}, {\em 33\/}, 49--70.

\bibitem[{Borst \& Helmstaedter(2015)}]{Borst2015(common-circuit-motion)}
Borst, A., \& Helmstaedter, M. (2015).
\newblock Common circuit design in fly and mammalian motion vision.
\newblock {\em Nature Neuroscience\/}, {\em 18\/}(8), 1067--1076.

\bibitem[{Browning et~al.(2009)Browning, Grossberg, \&
  Mingolla}]{Opticflow-2009(brain-realistic-scenes)}
Browning, N.~A., Grossberg, S., \& Mingolla, E. (2009).
\newblock A neural model of how the brain computes heading from optic flow in
  realistic scenes.
\newblock {\em Cognitive Psychology\/}, {\em 59\/}(4), 320--356.

\bibitem[{Card(2012)}]{Escape-2012(insects-behaviours)}
Card, G.~M. (2012).
\newblock Escape behaviors in insects.
\newblock {\em Current Opinion in Neurobiology\/}, {\em 22\/}(2), 180--186.

\bibitem[{Cheng et~al.(2019)Cheng, Cao, Zhang, \&
  Hao}]{Cheng-review-compound-eye-imaging}
Cheng, Y., Cao, J., Zhang, Y., \& Hao, Q. (2019).
\newblock Review of state-of-the-art artificial compound eye imaging systems.
\newblock {\em Bioinspiration {\&} Biomimetics\/}, {\em 14\/}(3), 031002.

\bibitem[{Cizek et~al.(2017)Cizek, Milicka, \& Faigl}]{LGMD1-walking-robot}
Cizek, P., Milicka, P., \& Faigl, J. (2017).
\newblock Neural based obstacle avoidance with {CPG} controlled hexapod walking
  robot.
\newblock In {\em Proceedings of the 2017 IEEE international joint conference
  on neural networks (IJCNN)\/}, (pp. 650--656). IEEE.

\bibitem[{Clark et~al.(2011)Clark, Bursztyn, Horowitz, Schnitzer, \&
  Clandinin}]{Clark_2011(6Q-model-fly)}
Clark, D.~A., Bursztyn, L., Horowitz, M.~A., Schnitzer, M.~J., \& Clandinin,
  T.~R. (2011).
\newblock Defining the computational structure of the motion detector in
  drosophila.
\newblock {\em Neuron\/}, {\em 70\/}(6), 1165--1177.

\bibitem[{Clifford \&
  Ibbotson(2002)}]{MotionReview2002(models-cells-functions)}
Clifford, C., \& Ibbotson, M. (2002).
\newblock Fundamental mechanisms of visual motion detection: models, cells and
  functions.
\newblock {\em Progress in Neurobiology\/}, {\em 68\/}(6), 409--437.

\bibitem[{CLIFFORD \& LANGLEY(1996)}]{AdaptiveFly-1996(Reichardt-detector)}
CLIFFORD, C. W.~G., \& LANGLEY, K. (1996).
\newblock A model of temporal adaptation in fly motion vision.
\newblock {\em Vision Research\/}, {\em 36\/}, 2595--2608.

\bibitem[{Collett(1971)}]{collett1971visual}
Collett, T. (1971).
\newblock Visual neurones for tracking moving targets.
\newblock {\em Nature\/}, {\em 232\/}(5306), 127.

\bibitem[{Collett \& Land(1975)}]{collett1975visual}
Collett, T., \& Land, M. (1975).
\newblock Visual control of flight behaviour in the hoverflysyritta pipiens l.
\newblock {\em Journal of Comparative Physiology\/}, {\em 99\/}(1), 1--66.

\bibitem[{Cope et~al.(2016)Cope, Sabo, Gurney, Vasilaki, \&
  Marshall}]{Cope2016(model-angular-bee)}
Cope, A.~J., Sabo, C., Gurney, K., Vasilaki, E., \& Marshall, J.~A. (2016).
\newblock A model for an angular velocity-tuned motion detector accounting for
  deviations in the corridor-centering response of the bee.
\newblock {\em PLoS Computational Biology\/}, {\em 12\/}(5), 1--22.

\bibitem[{{De Vries} \& Clandinin(2012)}]{FlyLooming-2012}
{De Vries}, S. E.~J., \& Clandinin, T.~R. (2012).
\newblock Loom-sensitive neurons link computation to action in the drosophila
  visual system.
\newblock {\em Current Biology\/}, {\em 22\/}(5), 353--362.

\bibitem[{Deng et~al.(2007)Deng, Inoue, Shibata, Sekiguchi, \&
  Ueki}]{LGMD1-2007(obstacle-avoidance-robot)}
Deng, M., Inoue, A., Shibata, Y., Sekiguchi, K., \& Ueki, N. (2007).
\newblock An obstacle avoidance method for two wheeled mobile robot.
\newblock In {\em Proceedings of the 2007 IEEE international conference on
  networking sensing and control\/}, (pp. 689--692). IEEE.

\bibitem[{DeSouza \& Kak(2002)}]{DeSouza_2002(survey-vision-robot)}
DeSouza, G.~N., \& Kak, A.~C. (2002).
\newblock Vision for mobile robot navigation: A survey.
\newblock {\em IEEE Transactions on Pattern Analysis and Machine
  Intelligence\/}, {\em 24\/}, 237--267.

\bibitem[{Dewell \& Gabbiani(2012)}]{Escape-2012(neural-computation-action)}
Dewell, R.~B., \& Gabbiani, F. (2012).
\newblock Escape behavior: Linking neural computation to action.
\newblock {\em Current Biology\/}, {\em 22\/}(5), R152--R153.

\bibitem[{Eckert \& Dvorak(1983)}]{eckert1983centrifugal}
Eckert, H., \& Dvorak, D.~R. (1983).
\newblock The centrifugal horizontal cells in the lobula plate of the blowfly,
  phaenicia sericata.
\newblock {\em Journal of Insect Physiology\/}, {\em 29\/}(7), 547--560.

\bibitem[{Egelhaaf(1985{\natexlab{a}})}]{egelhaaf1985neuronal-2}
Egelhaaf, M. (1985{\natexlab{a}}).
\newblock On the neuronal basis of figure-ground discrimination by relative
  motion in the visual system of the fly. 2: Figure-detection cells a new class
  of visual interneurons.
\newblock {\em Biological Cybernetics\/}, {\em 52\/}(2), 123--140.

\bibitem[{Egelhaaf(1985{\natexlab{b}})}]{egelhaaf1985neuronal-3}
Egelhaaf, M. (1985{\natexlab{b}}).
\newblock On the neuronal basis of figure-ground discrimination by relative
  motion in the visual system of the fly. 3: Possible input circuitries and
  behavioural significance of the {FD}-cells.
\newblock {\em Biological Cybernetics\/}, {\em 52\/}(4), 267--280.

\bibitem[{Egelhaaf \& Borst(1993)}]{Egelhaaf1993(fly-algorithms-neurons)}
Egelhaaf, M., \& Borst, A. (1993).
\newblock A look into the cockpit of the fly: Visual orientation, algorithms,
  and identified neurons.
\newblock {\em Journal of Neuroscience\/}, {\em 13\/}(11), 4563--4574.

\bibitem[{Egelhaaf et~al.(1993)Egelhaaf, Borst, Warzecha, Flecks, \&
  Wildemann}]{egelhaaf1993neural}
Egelhaaf, M., Borst, A., Warzecha, A.-K., Flecks, S., \& Wildemann, A. (1993).
\newblock Neural circuit tuning fly visual neurons to motion of small objects.
  ii. input organization of inhibitory circuit elements revealed by
  electrophysiological and optical recording techniques.
\newblock {\em Journal of Neurophysiology\/}, {\em 69\/}(2), 340--351.

\bibitem[{Egelhaaf et~al.(2001)Egelhaaf, Grewe, Kern, \&
  Warzecha}]{Blowfly2001(outdoor-motion-neuron)}
Egelhaaf, M., Grewe, J., Kern, R., \& Warzecha, A.~K. (2001).
\newblock Outdoor performance of a motion-sensitive neuron in the blowfly.
\newblock {\em Vision Research\/}, {\em 41\/}(27), 3627--3637.

\bibitem[{Eichner et~al.(2011)Eichner, Joesch, Schnell, Reiff, \&
  Borst}]{Eichner2011(2Q-motion)}
Eichner, H., Joesch, M., Schnell, B., Reiff, D.~F., \& Borst, A. (2011).
\newblock Internal structure of the fly elementary motion detector.
\newblock {\em Neuron\/}, {\em 70\/}(6), 1155--1164.

\bibitem[{Expert \& Ruffier(2015)}]{Expert-OF-uneven-terrain-following}
Expert, F., \& Ruffier, F. (2015).
\newblock Flying over uneven moving terrain based on optic-flow cues without
  any need for reference frames or accelerometers.
\newblock {\em Bioinspiration {\&} Biomimetics\/}, {\em 10\/}, 026003.

\bibitem[{Fisher et~al.(2015)Fisher, Leong, Sporar, Ketkar, Gohl, Clandinin, \&
  Silies}]{Fisher-L3(wide-field-local)}
Fisher, Y.~E., Leong, J. C.~S., Sporar, K., Ketkar, M.~D., Gohl, D.~M.,
  Clandinin, T.~R., \& Silies, M. (2015).
\newblock A class of visual neurons with wide-field properties is required for
  local motion detection.
\newblock {\em Current Biology\/}, {\em 25\/}(24), 3178--3189.

\bibitem[{Floreano et~al.(2013)Floreano, Pericet-Camara, Viollet, Ruffier,
  Bruckner, Leitel, Buss, Menouni, Expert, Juston, Dobrzynski, Eplattenier,
  Recktenwald, Mallot, \& Franceschini}]{Nicolas-group-CurvACE}
Floreano, D., Pericet-Camara, R., Viollet, S., Ruffier, F., Bruckner, A.,
  Leitel, R., Buss, W., Menouni, M., Expert, F., Juston, R., Dobrzynski, M.~K.,
  Eplattenier, G.~L., Recktenwald, F., Mallot, H.~A., \& Franceschini, N.
  (2013).
\newblock Miniature curved artificial compound eyes.
\newblock {\em Proceedings of the National Academy of Sciences\/}, {\em
  110\/}(23).

\bibitem[{Floreano et~al.(2005)Floreano, Zufferey, \&
  Nicoud}]{Floreano2005(wheel-wing-spiking)}
Floreano, D., Zufferey, J.-C., \& Nicoud, J.-D. (2005).
\newblock From wheels to wings with evolutionary spiking circuits.
\newblock {\em Artificial Life\/}, {\em 11\/}(1-2), 121--138.

\bibitem[{Franceschini(1975)}]{Nicolas-principle-fly-compound-eye}
Franceschini, N. (1975).
\newblock {\em Sampling of the visual environment by the compound eye of the
  fly: Fundamentals and applications\/}, (pp. 98--125).
\newblock Springer Berlin Heidelberg.

\bibitem[{Franceschini(1984)}]{Nicolas-fly-retinal-mosaic}
Franceschini, N. (1984).
\newblock {\em Chromatic Organization and Sexual Dimorphism of the Fly Retinal
  Mosaic\/}, (pp. 319--350).
\newblock Boston, MA: Springer US.

\bibitem[{Franceschini(2014)}]{Nicolas-2014(Review-Fly-Robot)}
Franceschini, N. (2014).
\newblock Small brains, smart machines: From fly vision to robot vision and
  back again.
\newblock {\em Proceedings of the IEEE\/}, {\em 102\/}, 751--781.

\bibitem[{Franceschini et~al.(1992)Franceschini, Pichon, \&
  Blanes}]{Nicolas-insects-robot-vision}
Franceschini, N., Pichon, J., \& Blanes, C. (1992).
\newblock From insect vision to robot vision.
\newblock {\em Philosophical Transactions of the Royal Society B\/}, {\em
  337\/}(1281), 283--294.

\bibitem[{Franceschini et~al.(1989)Franceschini, Riehle, \&
  Le~Nestour}]{Nicolas-1989(DSN-Insect-Neurons)}
Franceschini, N., Riehle, A., \& Le~Nestour, A. (1989).
\newblock Directionally selective motion detection by insect neurons.
\newblock In D.~G. Stavenga, \& R.~C. Hardie (Eds.) {\em Facets of Vision\/},
  (pp. 360--390). Springer Berlin Heidelberg.

\bibitem[{Franceschini et~al.(2007)Franceschini, Ruffier, \&
  Serres}]{Nicolas-bio-flying-robot-insect-piloting}
Franceschini, N., Ruffier, F., \& Serres, J. (2007).
\newblock A bio-inspired flying robot sheds light on insect piloting abilities.
\newblock {\em Current Biology\/}, {\em 17\/}, 329--335.

\bibitem[{Franceschini et~al.(2010)Franceschini, Ruffier, \&
  Serres}]{Nicolas-insect-inspired-autopilots}
Franceschini, N., Ruffier, F., \& Serres, J. (2010).
\newblock Insect inspired autopilots.
\newblock {\em Journal of Aero Aqua Bio-mechanisms\/}, {\em 1\/}(1), 2--10.

\bibitem[{Frye(2015)}]{Frye2015(EMDs-basic)}
Frye, M. (2015).
\newblock Elementary motion detectors.
\newblock {\em Current Biology\/}, {\em 25\/}(6), R215--R217.

\bibitem[{{Fu} et~al.(2018){Fu}, {Bellotto}, {Hu}, \& {Yue}}]{Fu-ROBIO-2018}
{Fu}, Q., {Bellotto}, N., {Hu}, C., \& {Yue}, S. (2018).
\newblock Performance of a visual fixation model in an autonomous micro robot
  inspired by drosophila physiology.
\newblock In {\em Proceedings of the 2018 IEEE international conference on
  robotics and biomimetics (ROBIO)\/}, (pp. 1802--1808). IEEE.

\bibitem[{Fu et~al.(2018{\natexlab{a}})Fu, Hu, Liu, \& Yue}]{Fu-TAROS(review)}
Fu, Q., Hu, C., Liu, P., \& Yue, S. (2018{\natexlab{a}}).
\newblock Towards computational models of insect motion detectors for robot
  vision.
\newblock In M.~Giuliani, T.~Assaf, \& M.~E. Giannaccini (Eds.) {\em Towards
  autonomous robotic systems conference\/}, (pp. 465--467). Springer
  International Publishing.

\bibitem[{Fu et~al.(2017)Fu, Hu, Liu, \& Yue}]{Fu2017a(LGMDs-IROS)}
Fu, Q., Hu, C., Liu, T., \& Yue, S. (2017).
\newblock Collision selective {LGMDs} neuron models research benefits from a
  vision-based autonomous micro robot.
\newblock In {\em Proceedings of the 2017 IEEE/RSJ international conference on
  intelligent robots and systems (IROS)\/}, (pp. 3996--4002). IEEE.

\bibitem[{Fu et~al.(2018{\natexlab{b}})Fu, Hu, Peng, \&
  Yue}]{Fu-2018(LGMD1-NN)}
Fu, Q., Hu, C., Peng, J., \& Yue, S. (2018{\natexlab{b}}).
\newblock Shaping the collision selectivity in a looming sensitive neuron model
  with parallel {ON} and {OFF} pathways and spike frequency adaptation.
\newblock {\em Neural Networks\/}, {\em 106\/}, 127--143.

\bibitem[{Fu \& Yue(2015)}]{Fu-2015(LGMD2-MLSP)}
Fu, Q., \& Yue, S. (2015).
\newblock Modelling {LGMD2} visual neuron system.
\newblock In {\em Proceedings of the 2015 IEEE 25th international workshop on
  machine learning for signal processing\/}, (pp. 1--6). IEEE.

\bibitem[{Fu \& Yue(2017{\natexlab{a}})}]{Fu-2017(ROBIO-fixation)}
Fu, Q., \& Yue, S. (2017{\natexlab{a}}).
\newblock Mimicking fly motion tracking and fixation behaviors with a hybrid
  visual neural network.
\newblock In {\em Proceedings of the 2017 IEEE international conference on
  robotics and biomimetics (ROBIO)\/}, (pp. 1636--1641). IEEE.

\bibitem[{Fu \& Yue(2017{\natexlab{b}})}]{Fu2017(fly-DSNs-IJCNN)}
Fu, Q., \& Yue, S. (2017{\natexlab{b}}).
\newblock Modeling direction selective visual neural network with on and off
  pathways for extracting motion cues from cluttered background.
\newblock In {\em Proceedings of the 2017 international joint conference on
  neural networks (IJCNN)\/}, (pp. 831--838). IEEE.

\bibitem[{Fu et~al.(2016)Fu, Yue, \& Hu}]{Fu-2016(LGMD2-BMVC)}
Fu, Q., Yue, S., \& Hu, C. (2016).
\newblock Bio-inspired collision detector with enhanced selectivity for ground
  robotic vision system.
\newblock In E.~R.~H. Richard C.~Wilson, \& W.~A.~P. Smith (Eds.) {\em British
  machine vision conference\/}, (pp. 1--13). BMVA Press.

\bibitem[{Gabbiani et~al.(2005)Gabbiani, Cohen, \&
  Laurent}]{Gabbiani2005(feed-forward-inhibition)}
Gabbiani, F., Cohen, I., \& Laurent, G. (2005).
\newblock Time-dependent activation of feed-forward inhibition in a
  looming-sensitive neuron.
\newblock {\em Journal of Neurophysiology\/}, {\em 94\/}(May 2005), 2150--2161.

\bibitem[{Gabbiani \& Jones(2011)}]{Circuit-genetic(genetic-push-motion)}
Gabbiani, F., \& Jones, P.~W. (2011).
\newblock A genetic push to understand motion detection.
\newblock {\em Neuron\/}, {\em 70\/}(6), 1023--1025.

\bibitem[{Gabbiani \& Krapp(2006)}]{Gabbiani2006(SFA-LGMD)}
Gabbiani, F., \& Krapp, H.~G. (2006).
\newblock Spike-frequency adaptation and intrinsic properties of an identified,
  looming-sensitive neuron.
\newblock {\em Journal of Neurophysiology\/}, {\em 96\/}(6), 2951--2962.

\bibitem[{Gabbiani et~al.(2004)Gabbiani, Krapp, Hatsopoulos, Mo, Koch, \&
  Laurent}]{Gabbiani2004(invariance-LGMD)}
Gabbiani, F., Krapp, H.~G., Hatsopoulos, N., Mo, C.~H., Koch, C., \& Laurent,
  G. (2004).
\newblock Multiplication and stimulus invariance in a looming-sensitive neuron.
\newblock {\em Journal of Physiology Paris\/}, {\em 98\/}(1-3 SPEC. ISS.),
  19--34.

\bibitem[{Gabbiani et~al.(2002{\natexlab{a}})Gabbiani, Krapp, Koch, \&
  Laurent}]{Gabbiani-2002(LGMD-multiplicative-computation)}
Gabbiani, F., Krapp, H.~G., Koch, C., \& Laurent, G. (2002{\natexlab{a}}).
\newblock Multiplicative computation by a looming-sensitive neuron.
\newblock In {\em Proceedings of the second joint 24th annual conference and
  the annual fall meeting of the biomedical engineering society] [engineering
  in medicine and biology\/}, (pp. 1968--1969). IEEE.

\bibitem[{Gabbiani et~al.(2002{\natexlab{b}})Gabbiani, Krapp, Koch, \&
  Laurent}]{Gabbiani2002(multiplicative-computation-LGMD)}
Gabbiani, F., Krapp, H.~G., Koch, C., \& Laurent, G. (2002{\natexlab{b}}).
\newblock Multiplicative computation in a visual neuron sensitive to looming.
\newblock {\em Nature\/}, {\em 420\/}(6913), 320--324.

\bibitem[{Gabbiani et~al.(1999{\natexlab{a}})Gabbiani, Krapp, \&
  Laurent}]{Gabbiani1999(computation-LGMD)}
Gabbiani, F., Krapp, H.~G., \& Laurent, G. (1999{\natexlab{a}}).
\newblock Computation of object approach by a wide-field, motion-sensitive
  neuron.
\newblock {\em Journal of Neuroscience\/}, {\em 19\/}(3), 1122--1141.

\bibitem[{Gabbiani et~al.(1999{\natexlab{b}})Gabbiani, Laurent, Hatsopoulos,
  Krapp, Rind, \& Simmons}]{Gabbiani-1999(many-ways-LGMD)}
Gabbiani, F., Laurent, G., Hatsopoulos, N., Krapp, H.~G., Rind, F.~C., \&
  Simmons, P.~J. (1999{\natexlab{b}}).
\newblock The many ways of building collision-sensitive neurons.
\newblock {\em Trends in Neurosciences\/}, {\em 22\/}(10), 437--438.

\bibitem[{Gabbiani et~al.(2001)Gabbiani, Mo, \&
  Laurent}]{Gabbiani-2001(LGMD-invariance-angular)}
Gabbiani, F., Mo, C., \& Laurent, G. (2001).
\newblock Invariance of angular threshold computation in a wide-field
  looming-sensitive neuron.
\newblock {\em The Journal of neuroscience : the official journal of the
  Society for Neuroscience\/}, {\em 21\/}(1), 314--329.

\bibitem[{Gauck \& Borst(1999)}]{gauck1999spatial}
Gauck, V., \& Borst, A. (1999).
\newblock Spatial response properties of contralateral inhibited lobula plate
  tangential cells in the fly visual system.
\newblock {\em Journal of Comparative Neurology\/}, {\em 406\/}(1), 51--71.

\bibitem[{Geurten et~al.(2007)Geurten, Nordstr{\"o}m, Sprayberry, Bolzon, \&
  O'Carroll}]{geurten2007neural}
Geurten, B.~R., Nordstr{\"o}m, K., Sprayberry, J.~D., Bolzon, D.~M., \&
  O'Carroll, D.~C. (2007).
\newblock Neural mechanisms underlying target detection in a dragonfly
  centrifugal neuron.
\newblock {\em Journal of Experimental Biology\/}, {\em 210\/}(18), 3277--3284.

\bibitem[{Gilbert(2013)}]{Fly2013(brain-visual-motion)}
Gilbert, C. (2013).
\newblock Brain connectivity: Revealing the fly visual motion circuit.
\newblock {\em Current Biology\/}, {\em 23\/}(18), R851--R853.

\bibitem[{Gray et~al.(2001)Gray, Lee, \&
  Robertson}]{Gray2001(DCMD-headon-stimuli)}
Gray, J.~R., Lee, J.~K., \& Robertson, R.~M. (2001).
\newblock Activity of descending contralateral movement detector neurons and
  collision avoidance behaviour in response to head-on visual stimuli in
  locusts.
\newblock {\em Journal of Comparative Physiology - A Sensory, Neural, and
  Behavioral Physiology\/}, {\em 187\/}(2), 115--129.

\bibitem[{Green \& Oh(2008)}]{OpticFlow-2008(optic-flow-MAV)}
Green, W.~E., \& Oh, P.~Y. (2008).
\newblock Optic-flow-based collision avoidance.
\newblock {\em IEEE Robotics Automation Magazine\/}, {\em 15\/}(1), 96--103.

\bibitem[{Haag et~al.(2016)Haag, Arenz, Serbe, Gabbiani, \&
  Borst}]{Fly2016(direction-selectivity-fly)}
Haag, J., Arenz, A., Serbe, E., Gabbiani, F., \& Borst, A. (2016).
\newblock Complementary mechanisms create direction selectivity in the fly.
\newblock {\em eLife\/}, {\em 5\/}, 1--15.

\bibitem[{Haag \& Borst(2002)}]{haag2002dendro}
Haag, J., \& Borst, A. (2002).
\newblock Dendro-dendritic interactions between motion-sensitive large-field
  neurons in the fly.
\newblock {\em Journal of Neuroscience\/}, {\em 22\/}(8), 3227--3233.

\bibitem[{Harris et~al.(1999)Harris, O'Carroll, \&
  Laughlin}]{EMD-1999(adaptation-temporal-fly)}
Harris, R.~A., O'Carroll, D.~C., \& Laughlin, S.~B. (1999).
\newblock Adaptation and the temporal delay filter of fly motion detectors.
\newblock {\em Vision Research\/}, {\em 39\/}(16), 2603--2613.

\bibitem[{Harrison(2005)}]{Harrison2005}
Harrison, R.~R. (2005).
\newblock A biologically inspired analog {IC} for visual collision detection.
\newblock {\em IEEE Transactions on Circuits and Systems I: Regular Papers\/},
  {\em 52\/}(11), 2308--2318.

\bibitem[{Hartbauer(2017)}]{LGMD-car-2017(bionic-vehicle-collision)}
Hartbauer, M. (2017).
\newblock Simplified bionic solutions: A simple bio-inspired vehicle collision
  detection system.
\newblock {\em Bioinspiration {\&} Biomimetics\/}, {\em 12\/}(2), 026007.

\bibitem[{Hassenstein \& Reichardt(1956)}]{HR-1956(EMD-1956)}
Hassenstein, B., \& Reichardt, W. (1956).
\newblock Systemtheoretische analyse der zeit-, reihenfolgen- und
  vorzeichenauswertung bei der bewegungsperzeption des riisselkiifers
  chlorophanus.
\newblock {\em Zeitschrift fur Naturforschung\/}, (pp. 513--524).

\bibitem[{Hennig \& Egelhaaf(2012)}]{hennig2012neuronal}
Hennig, P., \& Egelhaaf, M. (2012).
\newblock Neuronal encoding of object and distance information: A model
  simulation study on naturalistic optic flow processing.
\newblock {\em Frontiers in Neural Circuits\/}, {\em 6\/}, 14.

\bibitem[{Hennig et~al.(2011)Hennig, Kern, \& Egelhaaf}]{hennig2011binocular}
Hennig, P., Kern, R., \& Egelhaaf, M. (2011).
\newblock Binocular integration of visual information: A model study on
  naturalistic optic flow processing.
\newblock {\em Frontiers in Neural Circuits\/}, {\em 5\/}, 4.

\bibitem[{Hennig et~al.(2008)Hennig, M{\"o}ller, \&
  Egelhaaf}]{hennig2008distributed}
Hennig, P., M{\"o}ller, R., \& Egelhaaf, M. (2008).
\newblock Distributed dendritic processing facilitates object detection: A
  computational analysis on the visual system of the fly.
\newblock {\em PLoS One\/}, {\em 3\/}(8), e3092.

\bibitem[{Higgins(2004)}]{Higgins2004(nondirectional-motion-insect)}
Higgins, C.~M. (2004).
\newblock Nondirectional motion may underlie insect behavioral dependence on
  image speed.
\newblock {\em Biological Cybernetics\/}, {\em 91\/}(5), 326--332.

\bibitem[{Higgins \& Pant(2004)}]{higgins2004elaborated}
Higgins, C.~M., \& Pant, V. (2004).
\newblock An elaborated model of fly small-target tracking.
\newblock {\em Biological Cybernetics\/}, {\em 91\/}(6), 417--428.

\bibitem[{Horridge(1977)}]{Horridge-1977-insect-compound-eye}
Horridge, G.~A. (1977).
\newblock The compound eye of insects.
\newblock {\em Scientific American\/}, {\em 237\/}(1), 108--121.

\bibitem[{Hu et~al.(2017{\natexlab{a}})Hu, Yue, \& Zhang}]{Hu2016(RMPNN)}
Hu, B., Yue, S., \& Zhang, Z. (2017{\natexlab{a}}).
\newblock A rotational motion perception neural network based on asymmetric
  spatiotemporal visual information processing.
\newblock {\em IEEE Transactions on Neural Networks and Learning Systems\/},
  {\em 28\/}(11), 2803--2821.

\bibitem[{Hu \& Zhang(2018)}]{Hu2018(RMPNN)}
Hu, B., \& Zhang, Z. (2018).
\newblock Bio-plausible visual neural network for spatio-temporally spiral
  motion perception.
\newblock {\em Neurocomputing\/}, {\em 310\/}, 96--114.

\bibitem[{Hu et~al.(2017{\natexlab{b}})Hu, Arvin, Xiong, \&
  Yue}]{Hu-2017(Colias-LGMD1)}
Hu, C., Arvin, F., Xiong, C., \& Yue, S. (2017{\natexlab{b}}).
\newblock Bio-inspired embedded vision system for autonomous micro-robots: The
  lgmd case.
\newblock {\em IEEE Transactions on Cognitive and Developmental Systems\/},
  {\em 9\/}(3), 241--254.

\bibitem[{Hu et~al.(2014)Hu, Arvin, \& Yue}]{Hu-2014(LGMD1-ICDL)}
Hu, C., Arvin, F., \& Yue, S. (2014).
\newblock Development of a bio-inspired vision system for mobile micro-robots.
\newblock In {\em Proceedings of the 4th IEEE international conference on
  development and learning and on epigenetic robotics\/}, (pp. 81--86). IEEE.

\bibitem[{Huber \& B\"{u}lthoff(2003)}]{Huber-2003(visuomotor-book-review)}
Huber, S.~A., \& B\"{u}lthoff, H.~H. (2003).
\newblock Visuomotor control in flies and behavior - based agents.
\newblock In R.~J. Duro, J.~Santos, M.~Gra\~{n}a, \& J.~Kacprzyk (Eds.) {\em
  Biologically inspired robot behavior engineering\/}, (pp. 89--117).
  Heidelberg, Germany, Germany: Physica-Verlag GmbH.

\bibitem[{Iida(2003)}]{Iida2003(visual-odometer-robot)}
Iida, F. (2003).
\newblock Biologically inspired visual odometer for navigation of a flying
  robot.
\newblock {\em Robotics and Autonomous Systems\/}, {\em 44\/}(3-4), 201--208.

\bibitem[{Iida(2012)}]{Iida-2012(book-review)}
Iida, F. (2012).
\newblock Book review: Flying insects and robots.
\newblock {\em Artificial Life\/}, {\em 18\/}, 125--127.

\bibitem[{Iida \& Lambrinos(2000)}]{Iida_2000(fly-visual-odometer)}
Iida, F., \& Lambrinos, D. (2000).
\newblock Navigation in an autonomous flying robot by using a biologically
  inspired visual odometer.
\newblock {\em Sensor Fusion and Decentralized Control in RoboticSystem III
  Photonics East\/}, {\em 4196\/}, 86--97.

\bibitem[{Indiveri(1998)}]{Indiveri1998(VLSI-model-approach)}
Indiveri, G. (1998).
\newblock Analog vlsi model of locust dcmd neuron for computation of object
  approach.
\newblock {\em Neuromorphic Systems. Engineering Silicon from Neurobiology\/},
  {\em 10\/}, 47--60.

\bibitem[{Joesch et~al.(2010)Joesch, Schnell, Raghu, Reiff, \&
  Borst}]{Joesch-2010(ON-OFF-fly)}
Joesch, M., Schnell, B., Raghu, S.~V., Reiff, D.~F., \& Borst, A. (2010).
\newblock {ON} and {OFF} pathways in drosophila motion vision.
\newblock {\em Nature\/}, {\em 468\/}(7321), 300--304.

\bibitem[{Joesch et~al.(2013)Joesch, Weber, Eichner, \&
  Borst}]{Joesch_2013(functional-ON-OFF)}
Joesch, M., Weber, F., Eichner, H., \& Borst, A. (2013).
\newblock Functional specialization of parallel motion detection circuits in
  the fly.
\newblock {\em Journal of Neuroscience\/}, {\em 33\/}(3), 902--905.

\bibitem[{Judge \& Rind(1997)}]{DCMD-1997(collision-trajectories)}
Judge, S., \& Rind, F. (1997).
\newblock The locust {DCMD}, a movement-detecting neurone tightly tuned to
  collision trajectories.
\newblock {\em The Journal of Experimental Biology\/}, {\em 200\/}, 2209--16.

\bibitem[{Karmeier(2006)}]{Opticflow-2006(naturalistic-blowfly-motion)}
Karmeier, K. (2006).
\newblock Encoding of naturalistic optic flow by a population of blowfly
  motion-sensitive neurons.
\newblock {\em Journal of Neurophysiology\/}, {\em 96\/}(3), 1602--1614.

\bibitem[{Keil(2011)}]{Keil-2011(LGMD-algorithm-NIPS)}
Keil, M.~S. (2011).
\newblock Emergence of multiplication in a biophysical model of a wide-field
  visual neuron for computing object approaches: Dynamics, peaks, \& fits.
\newblock In J.~Shawe-Taylor, R.~S. Zemel, P.~L. Bartlett, F.~Pereira, \& K.~Q.
  Weinberger (Eds.) {\em Advances in Neural Information Processing Systems
  24\/}, (pp. 469--477). Curran Associates, Inc.

\bibitem[{Keil(2015)}]{Keil-2015(LGMD-dendritic-noisy)}
Keil, M.~S. (2015).
\newblock Dendritic pooling of noisy threshold processes can explain many
  properties of a collision-sensitive visual neuron.
\newblock {\em PLoS Computational Biology\/}, {\em 11\/}(10), 1--17.

\bibitem[{Keil et~al.(2004)Keil, Roca-Moreno, \&
  Rodriguez-Vazquez}]{LGMD1-ONOFF-2004}
Keil, M.~S., Roca-Moreno, E., \& Rodriguez-Vazquez, A. (2004).
\newblock A neural model of the locust visual system for detection of object
  approaches with real-world scenes.
\newblock In {\em Proceedings of the fourth IASTED international conference on
  visualization, imaging, and image processing\/}, (pp. 340--345). IASTED.

\bibitem[{Keleş \& Frye(2017)}]{Fly2017(object-detecting-neuron)}
Keleş, M.~F., \& Frye, M.~A. (2017).
\newblock Object-detecting neurons in drosophila.
\newblock {\em Current Biology\/}, {\em 27\/}(5), 680--687.

\bibitem[{Kerhuel et~al.(2012)Kerhuel, Viollet, \&
  Franceschini}]{Kerhuel-VODKA-sensor}
Kerhuel, L., Viollet, S., \& Franceschini, N. (2012).
\newblock The {VODKA} sensor: A bio-inspired hyperacute optical position
  sensing device.
\newblock {\em IEEE Sensors Journal\/}, {\em 12\/}(2), 315--324.

\bibitem[{Kimmerle \& Egelhaaf(2000)}]{kimmerle2000detection}
Kimmerle, B., \& Egelhaaf, M. (2000).
\newblock Detection of object motion by a fly neuron during simulated flight.
\newblock {\em Journal of Comparative Physiology A\/}, {\em 186\/}(1), 21--31.

\bibitem[{Kimmerle et~al.(1997)Kimmerle, Warzecha, \&
  Egelhaaf}]{kimmerle1997object}
Kimmerle, B., Warzecha, A.-K., \& Egelhaaf, M. (1997).
\newblock Object detection in the fly during simulated translatory flight.
\newblock {\em Journal of Comparative Physiology A\/}, {\em 181\/}(3),
  247--255.

\bibitem[{K{\"o}hler(2015)}]{Koehler2015}
K{\"o}hler, T. (2015).
\newblock {\em Bioinspired Motion Detection Based on an {FPGA} Platform\/},
  chap.~17, (pp. 405--424).
\newblock John Wiley {\&} Sons, Ltd.

\bibitem[{Kramer et~al.(1997)Kramer, Sarpeshkar, \&
  Koch}]{Kramer-CMOS-VLSI-EMD}
Kramer, J., Sarpeshkar, R., \& Koch, C. (1997).
\newblock Pulse-based analog {VLSI} velocity sensors.
\newblock {\em IEEE Transactions on Circuits and Systems-II: Analog and Digital
  Signal Processing\/}, {\em 44\/}(2), 86--101.

\bibitem[{Krapp et~al.(1998)Krapp, Hengstenberg, \&
  Hengstenberg}]{Fly-1998(dendritic-optic-flow)}
Krapp, H.~G., Hengstenberg, B., \& Hengstenberg, R. (1998).
\newblock Dendritic structure and receptive-field organization of optic flow
  processing interneurons in the fly.
\newblock {\em Journal of Neurophysiology\/}, {\em 79\/}(4), 1902--17.

\bibitem[{Krapp \&
  Hengstenberg(1996)}]{Opticflow-1996(self-motion-interneurons)}
Krapp, H.~G., \& Hengstenberg, R. (1996).
\newblock Estimation of self-motion by optic flow processing in single visual
  interneurons.
\newblock {\em Nature\/}, {\em 384\/}, 463--466.

\bibitem[{Krapp et~al.(2001)Krapp, Hengstenberg, \&
  Egelhaaf}]{krapp2001binocular}
Krapp, H.~G., Hengstenberg, R., \& Egelhaaf, M. (2001).
\newblock Binocular contributions to optic flow processing in the fly visual
  system.
\newblock {\em Journal of Neurophysiology\/}, {\em 85\/}(2), 724--734.

\bibitem[{Krejan \& Trost(2011)}]{LGMD1-car-2011(risk-collision-road)}
Krejan, A., \& Trost, A. (2011).
\newblock {LGMD}-based bio-inspired algorithm for detecting risk of collision
  of a road vehicle.
\newblock In {\em Proceedings of the 2011 IEEE 7th international symposium on
  image and signal processing and analysis\/}, (pp. 319--324). IEEE.

\bibitem[{Leong et~al.(2016)Leong, Esch, Poole, Ganguli, \&
  Clandinin}]{Fly-DS-2016(preferred-null)}
Leong, J. C.~S., Esch, J.~J., Poole, B., Ganguli, S., \& Clandinin, T.~R.
  (2016).
\newblock Direction selectivity in drosophila emerges from preferred-direction
  enhancement and null-direction suppression.
\newblock {\em Journal of Neuroscience\/}, {\em 36\/}(31), 8078--8092.

\bibitem[{Liu et~al.(2018{\natexlab{a}})Liu, Neumann, Fu, Pearson, \&
  Yu}]{Pengcheng-IROS18}
Liu, P., Neumann, G., Fu, Q., Pearson, S., \& Yu, H. (2018{\natexlab{a}}).
\newblock Energy-efficient design and control of a vibro-driven robot.
\newblock In {\em Proceedings of the 2018 IEEE/RSJ international conference on
  intelligent robots and systems (IROS)\/}, (pp. 1464--1469). IEEE.

\bibitem[{Liu et~al.(2018{\natexlab{b}})Liu, Yu, \&
  Cang}]{Pengcheng-NonlinearDynamics}
Liu, P., Yu, H., \& Cang, S. (2018{\natexlab{b}}).
\newblock Optimized adaptive tracking control for an underactuated vibro-driven
  capsule system.
\newblock {\em Nonlinear Dynamics\/}, {\em 94\/}(3), 1803--1817.

\bibitem[{Mafrica et~al.(2016)Mafrica, Servel, \&
  Ruffier}]{Mafrica-OF-minisensor-robot}
Mafrica, S., Servel, A., \& Ruffier, F. (2016).
\newblock Minimalistic optic flow sensors applied to indoor and outdoor visual
  guidance and odometry on a car-like robot.
\newblock {\em Bioinspiration {\&} Biomimetics\/}, {\em 11\/}(6), 066007.

\bibitem[{Maisak et~al.(2013)Maisak, Haag, Ammer, Serbe, Meier, Leonhardt,
  Schilling, Bahl, Rubin, Nern, Dickson, Reiff, Hopp, \&
  Borst}]{Maisak_2013(T4-T5-fly)}
Maisak, M.~S., Haag, J., Ammer, G., Serbe, E., Meier, M., Leonhardt, A.,
  Schilling, T., Bahl, A., Rubin, G.~M., Nern, A., Dickson, B.~J., Reiff,
  D.~F., Hopp, E., \& Borst, A. (2013).
\newblock A directional tuning map of drosophila elementary motion detectors.
\newblock {\em Nature\/}, {\em 500\/}(7461), 212--216.

\bibitem[{Martin \&
  Franceschini(1994)}]{Martin-compound-eye-obstacle-avoidance-mobile-vehicle}
Martin, N., \& Franceschini, N. (1994).
\newblock Obstacle avoidance and speed control in a mobile vehicle equipped
  with a compound eye.
\newblock In {\em Proceedings of the intelligent vehicles' 94 symposium\/},
  (pp. 381--386). IEEE.

\bibitem[{Medan et~al.(2007)Medan, Oliva, \&
  Tomsic}]{Crab-2007(LGN-visual-escape)}
Medan, V., Oliva, D., \& Tomsic, D. (2007).
\newblock Characterization of lobula giant neurons responsive to visual stimuli
  that elicit escape behaviors in the crab chasmagnathus.
\newblock {\em Journal of Neurophysiology\/}, {\em 98\/}(4), 2414--28.

\bibitem[{Meng et~al.(2010)Meng, Appiah, Yue, Hunter, Hobden, Priestley,
  Hobden, \& Pettit}]{Meng-2010(LGMD1-FPGA)}
Meng, H., Appiah, K., Yue, S., Hunter, A., Hobden, M., Priestley, N., Hobden,
  P., \& Pettit, C. (2010).
\newblock A modified model for the lobula giant movement detector and its
  {FPGA} implementation.
\newblock {\em Computer Vision and Image Understanding\/}, {\em 114\/},
  1238--1247.

\bibitem[{Meng et~al.(2009)Meng, Yue, Hunter, Appiah, Hobden, Priestley,
  Hobden, \& Pettit}]{Meng-2009(IJCNN-LGMD1)}
Meng, H., Yue, S., Hunter, A., Appiah, K., Hobden, M., Priestley, N., Hobden,
  P., \& Pettit, C. (2009).
\newblock A modified neural network model for lobula giant movement detector
  with additional depth movement feature.
\newblock In {\em Proceedings of the 2009 IEEE international joint conference
  on neural networks (IJCNN)\/}, (pp. 2078--2083). IEEE.

\bibitem[{Milde et~al.(2015)Milde, Bertrand, Benosmanz, Egelhaaf, \&
  Chicca}]{Milde2015}
Milde, M.~B., Bertrand, O.~J., Benosmanz, R., Egelhaaf, M., \& Chicca, E.
  (2015).
\newblock Bioinspired event-driven collision avoidance algorithm based on optic
  flow.
\newblock In {\em Event-based control, communication, and signal processing\/},
  (pp. 1--7). IEEE.

\bibitem[{Milde et~al.(2017)Milde, Blum, Dietm{\"{u}}ller, Sumislawska,
  Conradt, Indiveri, \&
  Sandamirskaya}]{Neurorobotics2017(avoidance-acquisition-neuromorphic)}
Milde, M.~B., Blum, H., Dietm{\"{u}}ller, A., Sumislawska, D., Conradt, J.,
  Indiveri, G., \& Sandamirskaya, Y. (2017).
\newblock Obstacle avoidance and target acquisition for robot navigation using
  a mixed signal analog/digital neuromorphic processing system.
\newblock {\em Frontiers in Neurorobotics\/}, {\em 11\/}, 1--17.

\bibitem[{Missler \& Kamangar(1995)}]{Fly1995(network-tracking-fly)}
Missler, J.~M., \& Kamangar, F.~A. (1995).
\newblock A neural network for pursuit tracking inspired by the fly visual
  system.
\newblock {\em Neural Networks\/}, {\em 8\/}(3), 463--480.

\bibitem[{Moeckel \& Liu(2007)}]{Moeckel-Liu-time-to-travel-model}
Moeckel, R., \& Liu, S.-C. (2007).
\newblock Motion detection circuits for a time-to-travel algorithm.
\newblock In {\em Proceedings of the 2007 IEEE international symposium on
  circuits and systems\/}, (pp. 3079--3082). IEEE.

\bibitem[{Muijres et~al.(2014)Muijres, Elzinga, Melis, \&
  Dickinson}]{Science-2014(fly-banked-turn)}
Muijres, F.~T., Elzinga, M.~J., Melis, J.~M., \& Dickinson, M.~H. (2014).
\newblock Flies evade looming targets by executing rapid visually directed
  banked turns.
\newblock {\em Science\/}, {\em 344\/}(6180), 172--177.

\bibitem[{Mura \& Franceschini(1996)}]{Mura-scanning-retina-ground-robot}
Mura, F., \& Franceschini, N. (1996).
\newblock Obstacle avoidance in a terrestrial mobile robot provided with a
  scanning retina.
\newblock In {\em Proceedings of the 1996 IEEE intelligent vehicles
  symposium\/}, (pp. 47--52). IEEE.

\bibitem[{Nakamura et~al.(2002)Nakamura, Ichimura, \&
  Sawada}]{EMD-2002(fast-global-motion)}
Nakamura, E., Ichimura, M., \& Sawada, K. (2002).
\newblock Fast global motion estimation algorithm based on elementary motion
  detectors.
\newblock In {\em Proceedings of the 2002 IEEE international conference on
  image processing\/}, vol.~2, (pp. 297--300). IEEE.

\bibitem[{Netter \&
  Franceschini(2002)}]{Netter-robotic-aircraft-terrain-neuromorphic}
Netter, T., \& Franceschini, N. (2002).
\newblock A robotic aircraft that follows terrain using a neuromorphic eye.
\newblock In {\em Proceedings of the 2002 IEEE/RSJ international conference on
  intelligent robots and systems (IROS)\/}, (pp. 129--134). IEEE.

\bibitem[{Nordstr{\"o}m(2012)}]{nordstrom2012neural}
Nordstr{\"o}m, K. (2012).
\newblock Neural specializations for small target detection in insects.
\newblock {\em Current Opinion in Neurobiology\/}, {\em 22\/}(2), 272--278.

\bibitem[{Nordstr{\"o}m et~al.(2006)Nordstr{\"o}m, Barnett, \&
  O'Carroll}]{nordstrom2006insect}
Nordstr{\"o}m, K., Barnett, P.~D., \& O'Carroll, D.~C. (2006).
\newblock Insect detection of small targets moving in visual clutter.
\newblock {\em PLoS Biology\/}, {\em 4\/}(3), e54.

\bibitem[{Nordstr{\"o}m et~al.(2011)Nordstr{\"o}m, Bolzon, \&
  O'Carroll}]{nordstrom2011spatial}
Nordstr{\"o}m, K., Bolzon, D.~M., \& O'Carroll, D.~C. (2011).
\newblock Spatial facilitation by a high-performance dragonfly target-detecting
  neuron.
\newblock {\em Biology Letters\/}, {\em 7\/}(4), 588--592.

\bibitem[{Nordstr{\"o}m \& O'Carroll(2006)}]{nordstrom2006small}
Nordstr{\"o}m, K., \& O'Carroll, D.~C. (2006).
\newblock Small object detection neurons in female hoverflies.
\newblock {\em Proceedings of the Royal Society of London B: Biological
  Sciences\/}, {\em 273\/}(1591), 1211--1216.

\bibitem[{O'Carroll(1993)}]{o1993feature}
O'Carroll, D. (1993).
\newblock Feature-detecting neurons in dragonflies.
\newblock {\em Nature\/}, {\em 362\/}(6420), 541.

\bibitem[{Olberg(1981)}]{olberg1981object}
Olberg, R.~M. (1981).
\newblock Object-and self-movement detectors in the ventral nerve cord of the
  dragonfly.
\newblock {\em Journal of Comparative Physiology\/}, {\em 141\/}(3), 327--334.

\bibitem[{Olberg(1986)}]{olberg1986identified}
Olberg, R.~M. (1986).
\newblock Identified target-selective visual interneurons descending from the
  dragonfly brain.
\newblock {\em Journal of Comparative Physiology A: Neuroethology, Sensory,
  Neural, and Behavioral Physiology\/}, {\em 159\/}(6), 827--840.

\bibitem[{Oliva et~al.(2007)Oliva, Medan, \&
  Tomsic}]{Crab-2007(escape-from-looming)}
Oliva, D., Medan, V., \& Tomsic, D. (2007).
\newblock Escape behavior and neuronal responses to looming stimuli in the crab
  chasmagnathus granulatus (decapoda: Grapsidae).
\newblock {\em Journal of Experimental Biology\/}, {\em 210\/}(5), 865--880.

\bibitem[{Oliva \& Tomsic(2014)}]{Crab2014(computation-approach-neurons)}
Oliva, D., \& Tomsic, D. (2014).
\newblock Computation of object approach by a system of visual motion-sensitive
  neurons in the crab neohelice.
\newblock {\em Journal of Neurophysiology\/}, {\em 112\/}(6), 1477--1490.

\bibitem[{O'Shea \& Rowell(1976)}]{LGMD-1976(ON-OFF)}
O'Shea, M., \& Rowell, C. H.~F. (1976).
\newblock The neuronal basis of a sensory analyser, the acridid movement
  detector system.
\newblock {\em Journal of Experimental Biology\/}, {\em 68\/}(2), 289--308.

\bibitem[{O'Shea \& Williams(1974)}]{LGMD-1974}
O'Shea, M., \& Williams, J.~L. (1974).
\newblock The anatomy and output connection of a locust visual interneurone;
  the lobular giant movement detector ({LGMD}) neurone.
\newblock {\em Journal of Comparative Physiology\/}, {\em 91\/}(3), 257--266.

\bibitem[{Pallus \& Fleishman(2014)}]{EMD-MainArticle}
Pallus, A., \& Fleishman, L.~J. (2014).
\newblock A two-dimensional visual motion detector based on biological
  principles.
\newblock Available at \url{https://muse.union.edu/visualmotion/main-article/}.
\newblock (accessed: July 2014).

\bibitem[{Paulk et~al.(2009)Paulk, Dacks, Phillips-Portillo, Fellous, \&
  Gronenberg}]{Bee-2009(visual-processing-brain)}
Paulk, A.~C., Dacks, A.~M., Phillips-Portillo, J., Fellous, J.-M., \&
  Gronenberg, W. (2009).
\newblock Visual processing in the central bee brain.
\newblock {\em Journal of Neuroscience\/}, {\em 29\/}(32), 9987--9999.

\bibitem[{Peron \& Gabbiani(2009{\natexlab{a}})}]{SFA-2009}
Peron, S., \& Gabbiani, F. (2009{\natexlab{a}}).
\newblock Spike frequency adaptation mediates looming stimulus selectivity in a
  collision-detecting neuron.
\newblock {\em Nature Neuroscience\/}, {\em 12\/}(3), 318--326.

\bibitem[{Peron \& Gabbiani(2009{\natexlab{b}})}]{SFA-2009Role}
Peron, S.~P., \& Gabbiani, F. (2009{\natexlab{b}}).
\newblock Role of spike-frequency adaptation in shaping neuronal response to
  dynamic stimuli.
\newblock {\em Biological Cybernetics\/}, {\em 100\/}(6), 505--520.

\bibitem[{Peron et~al.(2009)Peron, Jones, \&
  Gabbiani}]{LGMD-2009(presynaptic-looming-selectivity)}
Peron, S.~P., Jones, P.~W., \& Gabbiani, F. (2009).
\newblock Precise subcellular input retinotopy and its computational
  consequences in an identified visual interneuron.
\newblock {\em Neuron\/}, {\em 63\/}(6), 830--842.

\bibitem[{Pichon et~al.(1990)Pichon, Blanes, \&
  Franceschini}]{Pichon-mobile-robot-self-motion-sensor}
Pichon, J.-M., Blanes, C., \& Franceschini, N. (1990).
\newblock Visual guidance of a mobile robot equipped with a network of
  self-motion sensors.
\newblock {\em Mobile Robots IV\/}, {\em 1195\/}, 44--56.

\bibitem[{Portelli et~al.(2010)Portelli, Ruffier, \&
  Franceschini}]{Portelli-honeybees-height-restore}
Portelli, G., Ruffier, F., \& Franceschini, N. (2010).
\newblock Honeybees change their height to restore their optic flow.
\newblock {\em Journal of Comparative Physiology A\/}, {\em 196\/}(4),
  307--313.

\bibitem[{Posch et~al.(2011)Posch, Matolin, \& Wohlgenannt}]{Posch2011}
Posch, C., Matolin, D., \& Wohlgenannt, R. (2011).
\newblock A {QVGA} 143 d{B} dynamic range frame-free {PWM} image sensor with
  lossless pixel-level video compression and time-domain {CDS}.
\newblock {\em IEEE Journal of Solid-State Circuits\/}, {\em 46\/}(1),
  259--275.

\bibitem[{Raharijaona et~al.(2017)Raharijaona, Serres, Vanhoutte, \&
  Ruffier}]{Rahar-event-based-autopilot-OF-control}
Raharijaona, T., Serres, J., Vanhoutte, E., \& Ruffier, F. (2017).
\newblock Toward an insect-inspired event-based autopilot combining both visual
  and control events.
\newblock In {\em Proceedings of the 2017 3rd international conference on
  event-based control, communication and signal processing\/}, (pp. 1--7).
  IEEE.

\bibitem[{Rajesh et~al.(2005)Rajesh, O'Carroll, \&
  Abbott}]{Rajesh2005(velocity-estimators-insect)}
Rajesh, S., O'Carroll, D., \& Abbott, D. (2005).
\newblock Man-made velocity estimators based on insect vision.
\newblock {\em Smart Materials and Structures\/}, {\em 14\/}(2), 413--424.

\bibitem[{Reichardt(1987)}]{Reichardt-1987(evaluation-optical-motion)}
Reichardt, W. (1987).
\newblock Evaluation of optical motion information by movement detector.
\newblock {\em International Journal of Computer Vision\/}, {\em 161\/}(2),
  533--547.

\bibitem[{Reichardt et~al.(1989)Reichardt, Egelhaaf, \&
  Guo}]{reichardt1989processing}
Reichardt, W., Egelhaaf, M., \& Guo, A.-k. (1989).
\newblock Processing of figure and background motion in the visual system of
  the fly.
\newblock {\em Biological Cybernetics\/}, {\em 61\/}(5), 327--345.

\bibitem[{Reichardt et~al.(1983)Reichardt, Poggio, \&
  Hausen}]{reichardt1983figure}
Reichardt, W., Poggio, T., \& Hausen, K. (1983).
\newblock Figure-ground discrimination by relative movement in the visual
  system of the fly. part ii: Towards the neural circuitry.
\newblock {\em Biological Cybernetics\/}, {\em 46\/}(1), 1--30.

\bibitem[{Rind(1990{\natexlab{a}})}]{DSNs-1990(Rind-locust-DSNs)}
Rind, F. (1990{\natexlab{a}}).
\newblock Identification of directionally selective motion-detecting neurones
  in the locust lobula and their synaptic connections with an identified
  descending neurone.
\newblock {\em Journal of Experimental Biology\/}, {\em 149\/}, 21--43.

\bibitem[{Rind(1990{\natexlab{b}})}]{Rind-1990(locusts-DSNs-morphology)}
Rind, F.~C. (1990{\natexlab{b}}).
\newblock A directionally selective motion-detecting neurone in the brain of
  the locust: Physiological and morphological characterization.
\newblock {\em Journal of Experimental Biology\/}, {\em 149\/}, 1--19.

\bibitem[{Rind(1996)}]{LGMD1-1996(Rind-intracellular-neurons)}
Rind, F.~C. (1996).
\newblock Intracellular characterization of neurons in the locust brain
  signaling impending collision.
\newblock {\em Journal of Neurophysiology\/}, {\em 75\/}(3), 986--995.

\bibitem[{Rind(2002)}]{Rind2002(locust-biology-robot)}
Rind, F.~C. (2002).
\newblock Motion detectors in the locust visual system: From biology to robot
  sensors.
\newblock {\em Microscopy Research and Technique\/}, {\em 56\/}(4), 256--269.

\bibitem[{Rind \& Bramwell(1996)}]{LGMD1-1996(Rind-neural-network)}
Rind, F.~C., \& Bramwell, D.~I. (1996).
\newblock Neural network based on the input organization of an identified
  neuron signaling impending collision.
\newblock {\em Journal of Neurophysiology\/}, {\em 75\/}(3), 967--985.

\bibitem[{Rind \& Leitinger(2000)}]{Rind2000(bio-evidence)}
Rind, F.~C., \& Leitinger, G. (2000).
\newblock Immunocytochemical evidence that collision sensing neurons in the
  locust visual system contain acetylcholine.
\newblock {\em Journal of Comparative Neurology\/}, {\em 423\/}(3), 389--401.

\bibitem[{Rind et~al.(2008)Rind, Santer, \&
  Wright}]{Rind_2008(avoidance-flying-locust)}
Rind, F.~C., Santer, R.~D., \& Wright, G.~A. (2008).
\newblock Arousal facilitates collision avoidance mediated by a looming
  sensitive visual neuron in a flying locust.
\newblock {\em Journal of Neurophysiology\/}, {\em 100\/}, 670--680.

\bibitem[{Rind \& Simmons(1992)}]{DCMD-1992(selective-response-approaching)}
Rind, F.~C., \& Simmons, P.~J. (1992).
\newblock Orthopteran {DCMD} neuron : a reevaluation of responses to moving
  objects . {I} . selective responses to approaching objects.
\newblock {\em Journal of Neurophysiology\/}, {\em 68\/}(5), 1654--1666.

\bibitem[{Rind \& Simmons(1998)}]{Rind1998(local-circuit-locust)}
Rind, F.~C., \& Simmons, P.~J. (1998).
\newblock Local circuit for the computation of object approach by an identified
  visual neuron in the locust.
\newblock {\em Journal of Comparative Neurology\/}, {\em 395\/}(3), 405--415.

\bibitem[{Rind \& Simmons(1999)}]{LGMD1-1999(Rind-seeing-collision)}
Rind, F.~C., \& Simmons, P.~J. (1999).
\newblock Seeing what is coming: Building collision-sensitive neurones.
\newblock {\em Trends in Neurosciences\/}, {\em 22\/}(5), 215--220.

\bibitem[{Rind et~al.(2016)Rind, Wernitznig, Polt, Zankel, Gutl, Sztarker, \&
  Leitinger}]{LGMDs-2016}
Rind, F.~C., Wernitznig, S., Polt, P., Zankel, A., Gutl, D., Sztarker, J., \&
  Leitinger, G. (2016).
\newblock Two identified looming detectors in the locust: Ubiquitous lateral
  connections among their inputs contribute to selective responses to looming
  objects.
\newblock {\em Scientific Reports\/}, {\em 6\/}, 35525.

\bibitem[{Rister et~al.(2007)Rister, Pauls, Schnell, Ting, Lee, Sinakevitch,
  Morante, Strausfeld, Ito, \&
  Heisenberg}]{Rister-2007(dissection-motion-channel)}
Rister, J., Pauls, D., Schnell, B., Ting, C.-Y., Lee, C.-H., Sinakevitch, I.,
  Morante, J., Strausfeld, N.~J., Ito, K., \& Heisenberg, M. (2007).
\newblock Dissection of the peripheral motion channel in the visual system of
  drosophila melanogaster.
\newblock {\em Neuron\/}, {\em 56\/}(1), 155--170.

\bibitem[{Rivera-Alvidrez \& Higgins(2005)}]{EMD2005(contrast-saturation-EMD)}
Rivera-Alvidrez, Z., \& Higgins, C.~M. (2005).
\newblock Contrast saturation in a neuronally-based model of elementary motion
  detection.
\newblock {\em Neurocomputing\/}, {\em 65-66\/}(SPEC. ISS.), 173--179.

\bibitem[{Roubieu et~al.(2013)Roubieu, Expert, Sabiron, \&
  Ruffier}]{Roubieu-1-gram-sensor-EMD}
Roubieu, F., Expert, F., Sabiron, G., \& Ruffier, F. (2013).
\newblock A two-directional 1-gram visual motion sensor inspired by the fly’s
  eye.
\newblock {\em IEEE Sensors Journal\/}, {\em 13\/}(3), 1025--1035.

\bibitem[{Roubieu et~al.(2012)Roubieu, Serres, Franceschini, Ruffier, \&
  Viollet}]{hovercraft2012(autonomous-bees-control)}
Roubieu, F.~L., Serres, J., Franceschini, N., Ruffier, F., \& Viollet, S.
  (2012).
\newblock A fully-autonomous hovercraft inspired by bees: Wall following and
  speed control in straight and tapered corridors.
\newblock In {\em Proceedings of the 2012 IEEE international conference on
  robotics and biomimetics (ROBIO)\/}, (pp. 1311--1318). IEEE.

\bibitem[{Roubieu et~al.(2014)Roubieu, Serres, Colonnier, Franceschini,
  Viollet, \& Ruffier}]{hovercraft2014(vision-bee-corridor)}
Roubieu, F.~L., Serres, J.~R., Colonnier, F., Franceschini, N., Viollet, S., \&
  Ruffier, F. (2014).
\newblock A biomimetic vision-based hovercraft accounts for bees' complex
  behaviour in various corridors.
\newblock {\em Bioinspiration {\&} Biomimetics\/}, {\em 9\/}(3), 036003.

\bibitem[{Ruffier \& Franceschini(2004)}]{Ruffier-MAV-OF-ICRA}
Ruffier, F., \& Franceschini, N. (2004).
\newblock Visually guided micro-aerial vehicle: automatic take off, terrain
  following, landing and wind reaction.
\newblock In {\em Proceedings of the 2004 IEEE international conference on
  robotics and automation (ICRA)\/}, (pp. 2339--2346). IEEE.

\bibitem[{Ruffier \& Franceschini(2005)}]{Ruffier-OF-regulation-autopilot}
Ruffier, F., \& Franceschini, N. (2005).
\newblock Optic flow regulation: The key to aircraft automatic guidance.
\newblock {\em Robotics and Autonomous Systems\/}, {\em 50\/}(4), 177--194.

\bibitem[{Ruffier \& Franceschini(2015)}]{Ruffier-OF-MAV-unsteady-environments}
Ruffier, F., \& Franceschini, N. (2015).
\newblock Optic flow regulation in unsteady environments: A tethered {MAV}
  achieves terrain following and targeted landing over a moving platform.
\newblock {\em Journal of Intelligent {\&} Robotic Systems\/}, {\em 79\/}(2),
  275--293.

\bibitem[{Ruffier et~al.(2003{\natexlab{a}})Ruffier, Viollet, Amic, \&
  Franceschini}]{Ruffier-MAV-OF-circuits}
Ruffier, F., Viollet, S., Amic, S., \& Franceschini, N. (2003{\natexlab{a}}).
\newblock Bio-inspired optical flow circuits for the visual guidance of micro
  air vehicles.
\newblock In {\em Proceedings of the 2003 IEEE international symposium on
  circuits and systems\/}, vol.~3. IEEE.

\bibitem[{Ruffier et~al.(2003{\natexlab{b}})Ruffier, Viollet, \&
  Franceschini}]{Ruffier-two-aerial-micro-robots}
Ruffier, F., Viollet, S., \& Franceschini, N. (2003{\natexlab{b}}).
\newblock {OSCAR} and {OCTAVE}: Two bio-inspired visually guided aerial
  micro-robots.
\newblock In {\em Proceedings of the 11th international conference on advanced
  robotics\/}, (pp. 726--732). IEEE.

\bibitem[{Sabiron et~al.(2013)Sabiron, Chavent, Raharijaona, Fabiani, \&
  Ruffier}]{Sabiron-lightweight-sensor-OF-helicopter-outdoor}
Sabiron, G., Chavent, P., Raharijaona, T., Fabiani, P., \& Ruffier, F. (2013).
\newblock Low-speed optic-flow sensor onboard an unmanned helicopter flying
  outside over fields.
\newblock In {\em Proceedings of the 2013 IEEE international conference on
  robotics and automation\/}, (pp. 1742--1749). IEEE.

\bibitem[{Salt et~al.(2017)Salt, Indiveri, \&
  Sandamirskaya}]{UAV2017(LGMD1-spiking)}
Salt, L., Indiveri, G., \& Sandamirskaya, Y. (2017).
\newblock Obstacle avoidance with lgmd neuron: Towards a neuromorphic uav
  implementation.
\newblock In {\em Proceedings of the 2017 IEEE international symposium on
  circuits and systems (ISCAS)\/}, (pp. 1--4). IEEE.

\bibitem[{Santer et~al.(2012)Santer, Rind, \&
  Simmons}]{Locusts-2012(predator-prey-behaviours)}
Santer, R.~D., Rind, F.~C., \& Simmons, P.~J. (2012).
\newblock Predator versus prey: Locust looming-detector neuron and behavioural
  responses to stimuli representing attacking bird predators.
\newblock {\em PLoS One\/}, {\em 7\/}(11), 1--11.

\bibitem[{Sarkar et~al.(2013)Sarkar, Bello, van Hoof, \&
  Theuwissen}]{Sarkar2013}
Sarkar, M., Bello, D. S.~S., van Hoof, C., \& Theuwissen, A.~J. (2013).
\newblock Biologically inspired {CMOS} image sensor for fast motion and
  polarization detection.
\newblock {\em IEEE Sensors Journal\/}, {\em 13\/}(3), 1065--1073.

\bibitem[{Schnell et~al.(2012)Schnell, Raghu, Nern, \&
  Borst}]{ColumnarCells-2012(fly-wide-field)}
Schnell, B., Raghu, S.~V., Nern, A., \& Borst, A. (2012).
\newblock Columnar cells necessary for motion responses of wide-field visual
  interneurons in drosophila.
\newblock {\em Journal of Comparative Physiology\/}, {\em 198\/}, 389--395.

\bibitem[{Serres et~al.(2008)Serres, Masson, Ruffier, \&
  Franceschini}]{Serres-bee-centering-wall-following}
Serres, J.~R., Masson, G.~P., Ruffier, F., \& Franceschini, N. (2008).
\newblock A bee in the corridor: Centering and wall-following.
\newblock {\em Naturwissenschaften\/}, {\em 95\/}(12), 1181--1187.

\bibitem[{Serres \& Ruffier(2017)}]{Serres2017(review-optic-flow)}
Serres, J.~R., \& Ruffier, F. (2017).
\newblock Optic flow-based collision-free strategies: From insects to robots.
\newblock {\em Arthropod Structure and Development\/}, {\em 46\/}(5), 703--717.

\bibitem[{Shigang \& Rind(2005)}]{Yue-2005(LGMD1-ICRA)}
Shigang, Y., \& Rind, F.~C. (2005).
\newblock A collision detection system for a mobile robot inspired by the
  locust visual system.
\newblock In {\em Proceedings of the 2005 IEEE international conference on
  robotics and automation (ICRA)\/}, (pp. 3832--3837). IEEE.

\bibitem[{Shinomiya et~al.(2014)Shinomiya, Karuppudurai, Lin, Lu, Lee, \&
  Meinertzhagen}]{Shinomiya2014(candidate-off-fly)}
Shinomiya, K., Karuppudurai, T., Lin, T.~Y., Lu, Z., Lee, C.~H., \&
  Meinertzhagen, I.~A. (2014).
\newblock Candidate neural substrates for off-edge motion detection in
  drosophila.
\newblock {\em Current Biology\/}, {\em 24\/}(10), 1062--1070.

\bibitem[{Shinomiya et~al.(2015)Shinomiya, Takemura, Rivlin, Plaza, Scheffer,
  \& Meinertzhagen}]{CommonEvo-2015(ON-OFF-edge)}
Shinomiya, K., Takemura, S.-y., Rivlin, P.~K., Plaza, S.~M., Scheffer, L., \&
  Meinertzhagen, I.~A. (2015).
\newblock A common evolutionary origin for the on- and off-edge motion
  detection pathways of the drosophila visual system.
\newblock {\em Frontiers in Neural Circuits\/}, {\em 9\/}, 33.

\bibitem[{Silva \& Santos(2013{\natexlab{a}})}]{LGMD1-Silva1}
Silva, A., \& Santos, C. (2013{\natexlab{a}}).
\newblock Computational model of the {LGMD} neuron for automatic collision
  detection.
\newblock In {\em Proceedings of the 2013 IEEE 3rd portuguese meeting in
  bioengineering\/}, (pp. 1--4). IEEE.

\bibitem[{Silva \& Santos(2013{\natexlab{b}})}]{LGMD1-2013(Silva-IJCNN)}
Silva, A., \& Santos, C.~P. (2013{\natexlab{b}}).
\newblock Modeling disinhibition within a layered structure of the {LGMD}
  neuron.
\newblock In {\em Proceedings of the 2013 IEEE international joint conference
  on neural networks (IJCNN)\/}, (pp. 1--7). IEEE.

\bibitem[{Simmons \& Rind(1997)}]{Simmons-1997(LGMD2-neuron-locusts)}
Simmons, P.~J., \& Rind, F.~C. (1997).
\newblock Responses to object approach by a wide field visual neurone, the
  {LGMD2} of the locust: Characterization and image cues.
\newblock {\em Journal of Comparative Physiology - A Sensory, Neural, and
  Behavioral Physiology\/}, {\em 180\/}(3), 203--214.

\bibitem[{Simmons et~al.(2010)Simmons, Rind, \&
  Santer}]{LGMD1-Escapes2010(visual-startle-locusts)}
Simmons, P.~J., Rind, F.~C., \& Santer, R.~D. (2010).
\newblock Escapes with and without preparation: The neuroethology of visual
  startle in locusts.
\newblock {\em Journal of Insect Physiology\/}, {\em 56\/}(8), 876--883.

\bibitem[{Simmons et~al.(2013)Simmons, Sztarker, \& Rind}]{DCMD-2013}
Simmons, P.~J., Sztarker, J., \& Rind, F.~C. (2013).
\newblock Looming detection by identified visual interneurons during larval
  development of the locust locusta migratoria.
\newblock {\em The Journal of Experimental Biology\/}, {\em 216\/}(Pt 12),
  2266--2275.

\bibitem[{Snippe \& Koenderink(1994)}]{Snippe_1994(optical-velocity-EMD)}
Snippe, H.~P., \& Koenderink, J.~J. (1994).
\newblock Extraction of optical velocity by use of multi-input reichardt
  detectors.
\newblock {\em Journal of the Optical Society of America\/}, {\em 11\/}(4),
  1222--1236.

\bibitem[{Spalthoff et~al.(2010)Spalthoff, Egelhaaf, Tinnefeld, \&
  Kurtz}]{spalthoff2010localized}
Spalthoff, C., Egelhaaf, M., Tinnefeld, P., \& Kurtz, R. (2010).
\newblock Localized direction selective responses in the dendrites of visual
  interneurons of the fly.
\newblock {\em BMC Biology\/}, {\em 8\/}(1), 36.

\bibitem[{Srinivasan et~al.(1999)Srinivasan, Chahl, Weber, Venkatesh, Nagle, \&
  Zhang}]{Navigation-1999(robot-insect-vision)}
Srinivasan, M., Chahl, J., Weber, K., Venkatesh, S., Nagle, M., \& Zhang, S.
  (1999).
\newblock Robot navigation inspired by principles of insect vision.
\newblock {\em Robotics and Autonomous Systems\/}, {\em 26\/}(2-3), 203--216.

\bibitem[{Srinivasan(2011{\natexlab{a}})}]{Honeybee-2011(flight-navigation-robotics)}
Srinivasan, M.~V. (2011{\natexlab{a}}).
\newblock Honeybees as a model for the study of visually guided flight,
  navigation, and biologically inspired robotics.
\newblock {\em Physiological Reviews\/}, {\em 91\/}, 413--460.

\bibitem[{Srinivasan(2011{\natexlab{b}})}]{Srinivasan2011(visual-insects-robotics)}
Srinivasan, M.~V. (2011{\natexlab{b}}).
\newblock Visual control of navigation in insects and its relevance for
  robotics.
\newblock {\em Current Opinion in Neurobiology\/}, {\em 21\/}(4), 535--543.

\bibitem[{Stafford et~al.(2007)Stafford, Santer, \&
  Rind}]{LGMD1-car2007(collision-detection-cars)}
Stafford, R., Santer, R.~D., \& Rind, F.~C. (2007).
\newblock A bio-inspired visual collision detection mechanism for cars:
  Combining insect inspired neurons to create a robust system.
\newblock {\em Biosystems\/}, {\em 87\/}(2-3), 164--171.

\bibitem[{Stephane~Viollet et~al.(2010)Stephane~Viollet, Menouni, Kerhuel, \&
  Franceschini}]{Viollet-OF-VLSI-sensors}
Stephane~Viollet, T.~R., Franck~Ruffier, Menouni, M., Kerhuel, F. A.~L., \&
  Franceschini, N. (2010).
\newblock Characteristics of three miniature bio-inspired optic flow sensors in
  natural environments.
\newblock In {\em Proceedings of the 2010 IEEE international conference on
  sensor technologies and applications\/}, (pp. 51--55). IEEE.

\bibitem[{Strother et~al.(2014)Strother, Nern, \&
  Reiser}]{Strother2014(direct-observation-on-off)}
Strother, J.~A., Nern, A., \& Reiser, M.~B. (2014).
\newblock Direct observation of on and off pathways in the drosophila visual
  system.
\newblock {\em Current Biology\/}, {\em 24\/}(9), 976--983.

\bibitem[{Strother et~al.(2017)Strother, Wu, Wong, Nern, Rogers, Le, Rubin, \&
  Reiser}]{Fly-DS-2017(emergence-direction-selectivity)}
Strother, J.~A., Wu, S.~T., Wong, A.~M., Nern, A., Rogers, E.~M., Le, J.~Q.,
  Rubin, G.~M., \& Reiser, M.~B. (2017).
\newblock The emergence of directional selectivity in the visual motion pathway
  of drosophila.
\newblock {\em Neuron\/}, {\em 94\/}(1), 168--182.e10.

\bibitem[{Sztarker \& Rind(2014)}]{Sztarker2014(LGMD2-development)}
Sztarker, J., \& Rind, F.~C. (2014).
\newblock A look into the cockpit of the developing locust: Looming detectors
  and predator avoidance.
\newblock {\em Developmental Neurobiology\/}, {\em 74\/}(11), 1078--1095.

\bibitem[{Sztarker \& Tomsic(2008)}]{Crab-2008(neuronal-correlates-escape)}
Sztarker, J., \& Tomsic, D. (2008).
\newblock Neuronal correlates of the visually elicited escape response of the
  crab chasmagnathus upon seasonal variations, stimuli changes and perceptual
  alterations.
\newblock {\em Journal of Comparative Physiology A: Neuroethology, Sensory,
  Neural, and Behavioral Physiology\/}, {\em 194\/}(6), 587--596.

\bibitem[{Takemura et~al.(2013)Takemura, Bharioke, Lu, Nern, Vitaladevuni,
  Rivlin, Katz, Olbris, Plaza, Winston, Zhao, Horne, Fetter, Takemura, Blazek,
  Chang, Ogundeyi, Saunders, Shapiro, Sigmund, Rubin, Scheffer, Meinertzhagen,
  \& Chklovskii}]{Fly-2013(motion-circuit-connectomics)}
Takemura, S.-y., Bharioke, A., Lu, Z., Nern, A., Vitaladevuni, S., Rivlin,
  P.~K., Katz, W.~T., Olbris, D.~J., Plaza, S.~M., Winston, P., Zhao, T.,
  Horne, J.~A., Fetter, R.~D., Takemura, S., Blazek, K., Chang, L.-A.,
  Ogundeyi, O., Saunders, M.~a., Shapiro, V., Sigmund, C., Rubin, G.~M.,
  Scheffer, L.~K., Meinertzhagen, I.~a., \& Chklovskii, D.~B. (2013).
\newblock A visual motion detection circuit suggested by drosophila
  connectomics.
\newblock {\em Nature\/}, {\em 500\/}(7461), 175--181.

\bibitem[{Tammero(2004)}]{Fly-2004(Tammero-spatial-visuomotor)}
Tammero, L.~F. (2004).
\newblock Spatial organization of visuomotor reflexes in drosophila.
\newblock {\em Journal of Experimental Biology\/}, {\em 207\/}(1), 113--122.

\bibitem[{Tammero \& Dickinson(2002)}]{landing-2002}
Tammero, L.~F., \& Dickinson, M.~H. (2002).
\newblock Collision-avoidance and landing responses are mediated by separate
  pathways in the fruit fly, drosophila melanogaster.
\newblock {\em The Journal of Experimental Biology\/}, {\em 205\/}, 2785--2798.

\bibitem[{Vanhoutte et~al.(2017)Vanhoutte, Mafrica, Ruffier, Bootsma, \&
  Serres}]{Vanhoutte-time-of-travel-OF-micro-flying-robot}
Vanhoutte, E., Mafrica, S., Ruffier, F., Bootsma, R.~J., \& Serres, J. (2017).
\newblock Time-of-travel methods for measuring optical flow on board a micro
  flying robot.
\newblock {\em Sensors\/}, {\em 17\/}(3), 571.

\bibitem[{Vogt \& Desplan(2007)}]{Vogt-2007(first-steps-fly)}
Vogt, N., \& Desplan, C. (2007).
\newblock The first steps in drosophila motion detection.
\newblock {\em Neuron\/}, {\em 56\/}(1), 5--7.

\bibitem[{Wang et~al.(2018{\natexlab{a}})Wang, Peng, Baxter, Zhang, Wang, \&
  Yue}]{Huatian-ICANN-angular}
Wang, H., Peng, J., Baxter, P., Zhang, C., Wang, Z., \& Yue, S.
  (2018{\natexlab{a}}).
\newblock A model for detection of angular velocity of image motion based on
  the temporal tuning of the drosophila.
\newblock In V.~K{\r{u}}rkov{\'a}, Y.~Manolopoulos, B.~Hammer, L.~Iliadis, \&
  I.~Maglogiannis (Eds.) {\em Artificial neural networks and machine learning
  -- ICANN 2018\/}, (pp. 37--46). Springer International Publishing.

\bibitem[{Wang et~al.(2016)Wang, Peng, \& Yue}]{Wang-2016(IJCNN-STMD)}
Wang, H., Peng, J., \& Yue, S. (2016).
\newblock Bio-inspired small target motion detector with a new lateral
  inhibition mechanism.
\newblock In {\em Proceedings of the 2016 IEEE international joint conference
  on neural networks (IJCNN)\/}, (pp. 4751--4758). IEEE.

\bibitem[{Wang et~al.(2017)Wang, Peng, \& Yue}]{HWang-ICDL(LPTC-model)}
Wang, H., Peng, J., \& Yue, S. (2017).
\newblock An improved {LPTC} neural model for background motion direction
  estimation.
\newblock In {\em Proceedings of the 7th joint IEEE international conference on
  development and learning and on epigenetic robotics\/}, (pp. 47--52). IEEE.

\bibitem[{Wang et~al.(2018{\natexlab{b}})Wang, Peng, \&
  Yue}]{Hongxin-feedback-stmd}
Wang, H., Peng, J., \& Yue, S. (2018{\natexlab{b}}).
\newblock A feedback neural network for small target motion detection in
  cluttered backgrounds.
\newblock In V.~K{\r{u}}rkov{\'a}, Y.~Manolopoulos, B.~Hammer, L.~Iliadis, \&
  I.~Maglogiannis (Eds.) {\em Artificial neural networks and machine learning
  -- ICANN 2018\/}, (pp. 728--737). Springer International Publishing.

\bibitem[{Wang et~al.(in press at [IEEE Transactions on Cybernetics])Wang,
  Peng, \& Yue}]{wang2018directionally}
Wang, H., Peng, J., \& Yue, S. (in press at [IEEE Transactions on
  Cybernetics]).
\newblock A directionally selective small target motion detecting visual neural
  network in cluttered backgrounds.

\bibitem[{Warzecha et~al.(1993)Warzecha, Egelhaaf, \&
  Borst}]{warzecha1993neural}
Warzecha, A.-K., Egelhaaf, M., \& Borst, A. (1993).
\newblock Neural circuit tuning fly visual interneurons to motion of small
  objects. {I}. dissection of the circuit by pharmacological and
  photoinactivation techniques.
\newblock {\em Journal of Neurophysiology\/}, {\em 69\/}(2), 329--339.

\bibitem[{Warzecha et~al.(2013)Warzecha, Rosner, \&
  Grewe}]{Fly-2013(neuronal-variability)}
Warzecha, A.~K., Rosner, R., \& Grewe, J. (2013).
\newblock Impact and sources of neuronal variability in the fly's motion vision
  pathway.
\newblock {\em Journal of Physiology Paris\/}, {\em 107\/}(1-2), 26--40.

\bibitem[{Webb(2000)}]{Webb2000(robotics-animal-behaviour)}
Webb, B. (2000).
\newblock What does robotics offer animal behaviour?
\newblock {\em Animal Behaviour\/}, {\em 60\/}(5), 545--558.

\bibitem[{Webb(2001)}]{Webb2001(robots-models-behaviour)}
Webb, B. (2001).
\newblock Can robots make good models of biological behaviour?
\newblock {\em Behavioral and Brain Sciences\/}, {\em 24\/}(06), 1033--1050.

\bibitem[{Wernitznig et~al.(2015)Wernitznig, Rind, Polt, Zankel, Pritz, Kolb,
  Bock, \& Leitinger}]{LGMD1-synaptic2015(first-stage-locust)}
Wernitznig, S., Rind, F.~C., Polt, P., Zankel, A., Pritz, E., Kolb, D., Bock,
  E., \& Leitinger, G. (2015).
\newblock Synaptic connections of first-stage visual neurons in the locust
  schistocerca gregaria extend evolution of tetrad synapses back 200 million
  years.
\newblock {\em Journal of Comparative Neurology\/}, {\em 523\/}(2), 298--312.

\bibitem[{Wiederman \& O’Carroll(2013)}]{wiederman2013biologically}
Wiederman, S.~D., \& O’Carroll, D.~C. (2013).
\newblock Biologically inspired feature detection using cascaded correlations
  of off and on channels.
\newblock {\em Journal of Artificial Intelligence and Soft Computing
  Research\/}, {\em 3\/}(1), 5--14.

\bibitem[{Wiederman et~al.(2008)Wiederman, Shoemaker, \&
  O'Carroll}]{Wiederman2008(STMD-clutter)}
Wiederman, S.~D., Shoemaker, P.~A., \& O'Carroll, D.~C. (2008).
\newblock A model for the detection of moving targets in visual clutter
  inspired by insect physiology.
\newblock {\em PLoS One\/}, {\em 3\/}(7), 1--11.

\bibitem[{Wiederman et~al.(2013)Wiederman, Shoemaker, \&
  O'Carroll}]{Wiederman2013(ON-OFF-correlation)}
Wiederman, S.~D., Shoemaker, P.~A., \& O'Carroll, D.~C. (2013).
\newblock Correlation between {OFF} and {ON} channels underlies dark target
  selectivity in an insect visual system.
\newblock {\em Journal of Neuroscience\/}, {\em 33\/}(32), 13225--13232.

\bibitem[{Yakubowski et~al.(2016)Yakubowski, Mcmillan, \&
  Gray}]{LGMD-2016(Gray-background-motion)}
Yakubowski, J.~M., Mcmillan, G.~A., \& Gray, J.~R. (2016).
\newblock Background visual motion affects responses of an insect
  motion-sensitive neuron to objects deviating from a collision course.
\newblock {\em Physiological Reports\/}, {\em 4\/}(10), e12801.

\bibitem[{Yamawaki(2011)}]{Yamawaki2011(mantis-looming-defence)}
Yamawaki, Y. (2011).
\newblock Defence behaviours of the praying mantis tenodera aridifolia in
  response to looming objects.
\newblock {\em Journal of Insect Physiology\/}, {\em 57\/}(11), 1510--1517.

\bibitem[{Yang et~al.(2012)Yang, Wu, \&
  Guo}]{LGMD1-2012(multi-object-detection)}
Yang, T., Wu, S., \& Guo, D. (2012).
\newblock Dynamic range enhance of visual sensor circuits and application for
  multi-object motion detection.
\newblock In {\em Proceedings of the 2012 IEEE international conference on
  intelligent computation technology and automation\/}, (pp. 151--154). IEEE.

\bibitem[{Yue \& {Claire Rind}(2006)}]{Yue-2006(LGMD1-TSNN)}
Yue, S., \& {Claire Rind}, F. (2006).
\newblock Visual motion pattern extraction and fusion for collision detection
  in complex dynamic scenes.
\newblock {\em Computer Vision and Image Understanding\/}, {\em 104\/}(1),
  48--60.

\bibitem[{Yue \& Rind(2006)}]{LGMD1-Glayer(feature-enhancement)}
Yue, S., \& Rind, F.~C. (2006).
\newblock Collision detection in complex dynamic scenes using a lgmd based
  visual neural network with feature enhancement.
\newblock {\em IEEE Transactions on Neural Networks\/}, {\em 17\/}(3),
  705--716.

\bibitem[{Yue \& Rind(2007)}]{Yue-2007(locust-DSNs)}
Yue, S., \& Rind, F.~C. (2007).
\newblock A synthetic vision system using directionally selective motion
  detectors to recognize collision.
\newblock {\em Artificial Life\/}, {\em 13\/}(2), 93--122.

\bibitem[{Yue \& Rind(2009)}]{LGMD1-Yue2009(near-range-navigation)}
Yue, S., \& Rind, F.~C. (2009).
\newblock Near range path navigation using lgmd visual neural networks.
\newblock In {\em Proceedings of the 2009 2nd IEEE international conference on
  computer science and information technology\/}, (pp. 105--109). IEEE.

\bibitem[{Yue \& Rind(2013{\natexlab{a}})}]{Yue-2013(locust-DSNs)}
Yue, S., \& Rind, F.~C. (2013{\natexlab{a}}).
\newblock Postsynaptic organizations of directional selective visual neural
  networks for collision detection.
\newblock {\em Neurocomputing\/}, {\em 103\/}, 50--62.

\bibitem[{Yue \&
  Rind(2013{\natexlab{b}})}]{LGMD1-DSN-competing(LGMD1-DSNs-Hybrid)}
Yue, S., \& Rind, F.~C. (2013{\natexlab{b}}).
\newblock Redundant neural vision systems—competing for collision recognition
  roles.
\newblock {\em IEEE Transactions on Autonomous Mental Development\/}, {\em
  5\/}(2), 173--186.

\bibitem[{Yue et~al.(2006)Yue, Rind, Keil, Cuadri, \&
  Stafford}]{Yue-2006(LGMD1-car)}
Yue, S., Rind, F.~C., Keil, M.~S., Cuadri, J., \& Stafford, R. (2006).
\newblock A bio-inspired visual collision detection mechanism for cars:
  Optimisation of a model of a locust neuron to a novel environment.
\newblock {\em Neurocomputing\/}, {\em 69\/}(13-15), 1591--1598.

\bibitem[{Yue et~al.(2010)Yue, Santer, Yamawaki, \&
  Rind}]{Yue-2010(LGMD1-robot-bilateral)}
Yue, S., Santer, R.~D., Yamawaki, Y., \& Rind, F.~C. (2010).
\newblock Reactive direction control for a mobile robot: A locust-like control
  of escape direction emerges when a bilateral pair of model locust visual
  neurons are integrated.
\newblock {\em Autonomous Robots\/}, {\em 28\/}(2), 151--167.

\bibitem[{Zanker \& Braddick(1999)}]{Zanker1999(noise-motion-speed)}
Zanker, J.~M., \& Braddick, O.~J. (1999).
\newblock How does noise influence the estimation of speed?
\newblock {\em Vision Research\/}, {\em 39\/}(14), 2411--2420.

\bibitem[{Zanker et~al.(1999)Zanker, Srinivasan, \&
  Egelhaaf}]{Zanker-1999(EMD-speed-tuning)}
Zanker, J.~M., Srinivasan, M.~V., \& Egelhaaf, M. (1999).
\newblock Speed tuning in elementary motion detectors of the correlation type.
\newblock {\em Biological Cybernetics\/}, {\em 80\/}(2), 109--116.

\bibitem[{Zanker \& Zeil(2005)}]{Zanker_2005(motion-signal-outdoor)}
Zanker, J.~M., \& Zeil, J. (2005).
\newblock Movement-induced motion signal distributions in outdoor scenes.
\newblock {\em Network: Computation in Neural Systems\/}, {\em 16\/}(4),
  357--376.

\bibitem[{Zeil(2012)}]{Zeil2012(visual-homing-insect)}
Zeil, J. (2012).
\newblock Visual homing: An insect perspective.
\newblock {\em Current Opinion in Neurobiology\/}, {\em 22\/}(2), 285--293.

\bibitem[{Zhang et~al.(2016)Zhang, Zhang, \& Yue}]{Zhang-2016(IJCNN-hybrid)}
Zhang, G., Zhang, C., \& Yue, S. (2016).
\newblock {LGMD} and {DSNs} neural networks integration for collision
  predication.
\newblock In {\em Proceedings of the 2016 IEEE international joint conference
  on neural networks (IJCNN)\/}, (pp. 1174--1179). IEEE.

\bibitem[{Zhang et~al.(2008)Zhang, Wu, Borst, Kuhnlenz, \& Buss}]{Zhang2008}
Zhang, T., Wu, H., Borst, A., Kuhnlenz, K., \& Buss, M. (2008).
\newblock An {FPGA} implementation of insect-inspired motion detector for
  high-speed vision systems.
\newblock In {\em Proceedings of the 2008 IEEE international conference on
  robotics and automation (ICRA)\/}, (pp. 335--340). IEEE.

\bibitem[{Zhang et~al.(2015)Zhang, Yue, \&
  Zhang}]{Fly-2015(model-collision-reichardt)}
Zhang, Z., Yue, S., \& Zhang, G. (2015).
\newblock Fly visual system inspired artificial neural network for collision
  detection.
\newblock {\em Neurocomputing\/}, {\em 153\/}, 221--234.

\bibitem[{Zhao et~al.(2018)Zhao, Hu, Zhang, Wang, \&
  Yue}]{UAV-2018(LGMD1-Opticflow-PID)}
Zhao, J., Hu, C., Zhang, C., Wang, Z., \& Yue, S. (2018).
\newblock A bio-inspired collision detector for small quadcopter.
\newblock In {\em Proceedings of the 2018 IEEE International joint conference
  on neural networks (IJCNN)\/}, (pp. 1--7). IEEE.

\bibitem[{Zheng et~al.(2006)Zheng, de~Polavieja, Wolfram, Asyali, Hardie, \&
  Juusola}]{Fly-2006(feedback-control-photoreceptor)}
Zheng, L., de~Polavieja, G.~G., Wolfram, V., Asyali, M.~H., Hardie, R.~C., \&
  Juusola, M. (2006).
\newblock Feedback network controls photoreceptor output at the layer of first
  visual synapses in drosophila.
\newblock {\em The Journal of General Physiology\/}, {\em 127\/}(5), 495--510.

\end{thebibliography}
\bibliographystyle{apa-good}

\end{document}